\documentclass[preprint,10pt]{elsarticle}
\usepackage{natbib}
\usepackage{geometry}
\usepackage{import}
\usepackage{enumitem}
\usepackage{amsfonts}
\usepackage{algorithmic}
\usepackage{graphicx}
\usepackage{adjustbox}
\usepackage{changepage}

\usepackage{array}
\usepackage[caption=false,font=normalsize,labelfont=sf,textfont=sf]{subfig}
\usepackage{graphicx}
\usepackage{textcomp}
\usepackage{amsthm}
\usepackage{stfloats}
\usepackage{subfig}
\usepackage{multirow}
\usepackage{subcaption}
\usepackage{float}
\usepackage{verbatim}
\usepackage{graphicx}
\usepackage{balance}
\usepackage[scr=rsfs]{mathalfa}
\usepackage{amsmath, amssymb, bm}
 \usepackage{hyperref}
\usepackage{cleveref}
\usepackage{algorithm}
 \usepackage{setspace}
 \usepackage{xcolor}
\usepackage{tikz}
\usepackage{bbm}
\usepackage{lipsum}
\usepackage[acronym]{glossaries}
\glsdisablehyper
\newacronym{name}{2DSig-Norm}{{}}
\newacronym{tag_name}{2DSig-Detect}{{}}
  
\newtheorem{theorem}{Theorem}[section]
\newtheorem{definition}[theorem]{Definition}

\newtheorem{prop}[theorem]{Proposition}

\usepackage{cleveref}
\usepackage{mathtools}
\usepackage{subfig}

\Crefformat{figure}{#2Fig.~#1#3}
\Crefmultiformat{figure}{Figs.~#2#1#3}{ and~#2#1#3}{, #2#1#3}{ and~#2#1#3}
\Crefformat{section}{#2Sec.~#1#3}
\Crefformat{algorithm}{#2Alg.~#1#3}
\Crefformat{theorem}{#2Thm.~#1#3}
\Crefformat{example}{#2Example~#1#3}
\Crefformat{definition}{#2Def.~#1#3}
\Crefformat{proposition}{#2Prop.~#1#3}
\Crefformat{corollary}{#2Cor.~#1#3}
\let\Algorithm\algorithm
\renewcommand\algorithm[1][]{\Algorithm[#1]\setstretch{1.5}}

\newglossaryentry{corpus}{name={corpus},
description={},
symbol={\ensuremath{\mathcal{D}_n}}
}
\newglossaryentry{ES}{name={Expected signature},
description={},
symbol={\ensuremath{\varphi}}
}
\newglossaryentry{cov}{name={same covariance matrix},
description={},
symbol={\ensuremath{\Sigma}}
}
\newglossaryentry{covn}{name={same covariance matrix},
description={},
symbol={\ensuremath{\Sigma_n}}
}

\usepackage[]{footmisc} %no dots in the footnotes!
\journal{Pattern Recognition}
\renewcommand{\footnoterule}{%
   
    \kern -3pt
    \hspace{1.5cm}\rule{0.25\textwidth}{.05pt}
    \kern -3pt
}

\begin{document}

\title{2DSig-Detect: a semi-supervised framework for anomaly detection on image data using 2D-signatures}

 % \begin{itemize}[leftmargin=0.3cm]
 %     \item[\IEEEauthorrefmark{1}] Department of Mathematics and Statistics at University of Strathclyde, Glasgow, G1 1XH, UK.
 %     \item[\IEEEauthorrefmark{2}] Defence AI Research (DARe), The Alan Turing Institute, London NW1 2DB, UK.
 %     \item[\IEEEauthorrefmark{4}] equal contribution, and VN is the PI of The Alan Turing Institute DARe team. 
 %     \item[\IEEEauthorrefmark{3}] corresponding author, e-mail: yue.wu@strath.ac.uk.
 % \end{itemize}

 \affiliation[clyde]{Department of Mathematics and Statistics at University of Strathclyde, Glasgow, G1 1XH, UK.}

 \affiliation[ATI]{Defence AI Research (DARe), The Alan Turing Institute, London NW1 2DB, UK.}    
 \affiliation[ATIeq]{equal contribution, and VN is the PI of The Alan Turing Institute DARe team.}        
\affiliation[*]{corresponding author, e-mail: yue.wu@strath.ac.uk}

\author[clyde]{Xinheng Xie}
\author[ATI,ATIeq]{Kureha Yamaguchi}
\author[ATI,ATIeq]{Margaux Leblanc}
\author[ATI,ATIeq]{Simon Malzard}
\author[ATI,ATIeq]{Varun Chhabra}
\author[ATI]{Victoria Nockles}
\author[clyde,*]{Yue Wu}

% \author{Authors.\thanks{}}
% \author{%
%   \IEEEauthorblockN{%
%     Xinheng Xie\IEEEauthorrefmark{1}, %\textsuperscript{\textsection},
%        Kureha Yamaguchi\IEEEauthorrefmark{2}\IEEEauthorrefmark{4},
%         % Kureha Yamaguchi 
%         % (Alphabetical Ordering until decision made)
%         Margaux Leblanc\IEEEauthorrefmark{2}\IEEEauthorrefmark{4}, 
%         % Margaux Leblanc
%         Simon Malzard\IEEEauthorrefmark{2}\IEEEauthorrefmark{4}, 
%         % Simon Malzard
%         Varun Chhabra\IEEEauthorrefmark{2}\IEEEauthorrefmark{4}, 
%         % Varun Chhabra
%         %\IEEEauthorrefmark{4}
%             Victoria Nockles\IEEEauthorrefmark{2},
%             % Victoria Nockles
%             %\IEEEauthorrefmark{4}
%     and Yue  Wu \IEEEauthorrefmark{1}\IEEEauthorrefmark{3}%
%   }%
 % \IEEEauthorblockA{\IEEEauthorrefmark{1} Affiliation 1}%
 % \IEEEauthorblockA{\IEEEauthorrefmark{2} The Alan Turing Institute, London, UK}%
 % \IEEEauthorblockA{\IEEEauthorrefmark{3} Affiliation 3}%
% }

%
\begin{abstract}

{The rapid and widespread deployment of machine learning within critical systems raises serious questions about their security against adversarial attacks. Models performing image-related tasks, are vulnerable to integrity violations causing misclassifications that do not compromise normal system operation, but rather, produce undesirable outcomes for targeted inputs. This paper introduces a novel technique for anomaly detection in images called \emph{\acrshort{tag_name}}, which is a 2D-signature-embedded semi-supervised framework rooted in rough path theory.  We demonstrate that 2D-signatures can be applied to detect both adversarial examples and backdoored data, mitigating against test-time and training-time integrity attacks. Our results demonstrate both the efficacy and superior computational  efficiency of our 2D-signature method for detecting adversarial manipulations to the training and test data.} 
  
\end{abstract}


\maketitle

% \IEEEdisplaynontitleabstractindextext
% \IEEEpeerreviewmaketitle

\begingroup\renewcommand\thefootnote{}

\footnotetext{\textit{
 % \begin{itemize}[leftmargin=0.3cm]
 %     \item[\IEEEauthorrefmark{1}] Department of Mathematics and Statistics at University of Strathclyde, Glasgow, G1 1XH, UK.
 %     \item[\IEEEauthorrefmark{2}] Defence AI Research (DARe), The Alan Turing Institute, London NW1 2DB, UK.
 %     \item[\IEEEauthorrefmark{4}] equal contribution, and VN is the PI of The Alan Turing Institute DARe team. 
 %     \item[\IEEEauthorrefmark{3}] corresponding author, e-mail: yue.wu@strath.ac.uk.
 % \end{itemize}
%}}
%\footnotetext{\textit{
%\hspace{0pt}\\
This work was supported by the Turing's Defence and Security programme through a partnership with the UK government in accordance to the framework agreement between GCHQ \& The Alan Turing  Institute.}}

\endgroup

%%%%% Main Section

\section{Introduction}\label{sec:introduction}
In recent years, AI has advanced significantly, especially in tasks such as image and object recognition as well as image generation \cite{goodfellow2014generative}\cite{he2016deep}\cite{ho2020denoising}. However, these models are quite vulnerable; even tiny modifications can disrupt their functionality. At the data level, the two major threats are data poisoning  \cite{fung2020limitations}\cite{gu2019badnets}  and evasion attacks  \cite{carlini2017towards}\cite{madry2017towards}. Data poisoning attacks corrupt the training data, leading AI models to learn incorrect behaviors \cite{biggio2012poisoning}. Examples of this include label flipping  \cite{fung2020limitations} and backdoor attacks  \cite{gu2019badnets}. Evasion attacks, on the other hand, deceive the AI model at inference time by presenting specially crafted input samples that cause misclassifications  \cite{goodfellow2014explaining}. For instance, traditional adversarial techniques such as the projected gradient descent method (PGDM)  \cite{madry2017towards}, the square attack   \cite{andriushchenko2020square}, and the method developed by Carlini \& Wagner (C\&W)  \cite{carlini2017towards}, exemplify approaches that apply minor, often imperceptible, changes to input data, effectively misleading machine learning models into making erroneous classifications or decisions.
Consequently, developing robust defenses to safeguard these AI models is critical to ensure their reliability and security.
Adversarial training has been proposed as a method to improve robustness against specific attacks. However, while effective to some extent, supervised learning techniques, including adversarial training, may still be vulnerable to unknown test-time evasion (TTE) attacks that were not included in the training set   \cite{li2022review}. Semi-supervised or unsupervised anomaly detection provides a practical alternative as it does not require labeling all the data. These methods can be utilised in various scenarios, improving the model's adaptability and generalisation to new and unseen data types, thus lowering the costs associated with data preparation. 
Directly benefiting from conventional anomaly detection methods to address adversarial attacks on images however remains challenging for three primary reasons. First, adversarial attacks are intentionally designed to be imperceptible to the human eye, making the noise they introduce very subtle. In some advanced attacks, even a single pixel may be altered  \cite{Jiawei2019OnePixel}, resulting in only a slight difference between benign and compromised images \cite{Grosse2017StatDectADML}. Consequently, traditional methods struggle to distinguish these two distributions effectively \cite{Liang2021DetectAdImageExamples}. 
{Second, there is a lack of faithful, low-dimensional features for images, which affects the performance of conventional methods such as Gaussian Mixture Models (GMM) or Support Vector Machines (SVMs) with high-dimensional data \cite{Lingyan2011HighDimGMM}.} This high dimensionality further complicates the application of traditional anomaly detection techniques. Third, adversarial attacks do not necessarily follow a particular form of distribution, such as Gaussian, as exemplified by the canonical assumptions of GMMs.

In this paper, we seek to resolve these challenges through embedding 2D-signature features of learned representations (of images) into a semi-supervised anomaly detection framework. We call our algorithm \emph{2DSig-Detect} (Algorithm \ref{alg:2dsig-detect}). Fundamentally, we define an anomaly score associated to an image based on its "distance" in embedding space to a dataset of instances labelled as normal or benign. We investigate two choices of distance, namely the conformance score and the covariance norm. We apply a threshold to this anomaly score to ascertain whether a given test instance is anomalous relative to the benign corpus. Our approach does not make any assumptions on the distribution of the data or adversarial attack, and through our usage of 2D-signatures and learned representations we obtain useful low-dimensional features. Moreover our algorithm has a free choice of distance metric. Combined, these mitigate some of the issues with conventional algorithms.

\noindent {Below we outline our novel contributions in this paper:}

\begin{itemize}
    \item {We introduce 2D-signature method which allows for the computation of both cross-channels and self-channels. Prior 2D-signature calculations only compute cross-channel terms and thus result in not being able to learn "gray-scale" information, i.e. information from correlations from the same R, G or B channel with itself.}
    \item {Compared with non-signature based methods, i.e. CNNs, our method allows for a compact description which separates out the cross-channel and self-channel information. CNNs automatically combine this information over the convolution kernel.}
    \item {As a proof of concept, we demonstrate that the 2D-signature method can be applied to mitigate against evasion and backdoor attacks. Our results are competitive with the spectral signature method \cite{tran2018spectral}
    for mitigating backdoor attacks, whilst being significantly more computationally efficient.}
\end{itemize}

\subsection{2D-signature from learned representations}

The signature transform, based in rough path theory  \cite{lyons1998differential}, can be used in machine learning as a feature transform on the underlying path of multi-modal streamed data \cite{lyons2014rough}. The resulting signature is a graded infinite dimensional sequence composed of the coordinate iterated integrals of the path. The signature is a natural choice for the feature set of streamed data for three reasons: it captures cross-channel interactions, is an efficient summary of the data, and ignoring a pathological set, uniquely characterises the underlying path of the data.

With its efficiency in capturing the total ordering of data and summarising the data over segments, signature-based machine learning models have been proved efficient in several fields of application, from automated recognition of Chinese handwriting  \cite{xie2017learning} to early-warning prediction of sepsis  \cite{morrill2020utilization, morrill2019signature} and the diagnosis of mental health problems  \cite{perez2018signature,moore2019random}.

Recently, the signature transform has been generalised to higher orders \cite{giusti2022topological, diehl2022two, diehl2024signature}, in particular to two dimensions. The 2D-signature transform fulfills a similar role for images as the signature does for time series. It provides a graded summary of image data (which may have an arbitrary number of channels) over their "area" information. The signature is graded in the sense that its sequence is delineated in levels, where each level captures increasing statistical variation of the path and surface for the 1D and 2D case respectively. One major advantage of using 2D-signature features is that the length of the 2D-signature at each level only depends on the number of channels, thus does not grow with resolution. This invariance means that the computational complexity of the 2D-signature transform is manageable, leading to consistent performance across varying resolutions. Moreover, 2D-signature features are more memory and compute efficient compared with the storage and processing costs of the original image. This is particularly important for applications with limited resources, such as mobile devices.

By expanding integrated integrals from 1D line elements to 2D plane elements, a freedom of choice of parameterisation across 2D elements allows for the computation of both cross-channel and self-channel terms. We note that in \cite{giusti2022topological} a restriction to only computing 2D-signatures with cross channel terms results in not being able to learn gray-scale information. We compute both sets of terms in our implementation. Our 2D-signatures therefore have the advantage of compactly being able to separate out the contribution from information coming from the same channel to information coming from different channels, where conventional CNN models automatically combine this information. 

\begin{figure}[!t]
    \centering
    \includegraphics[width=6in]{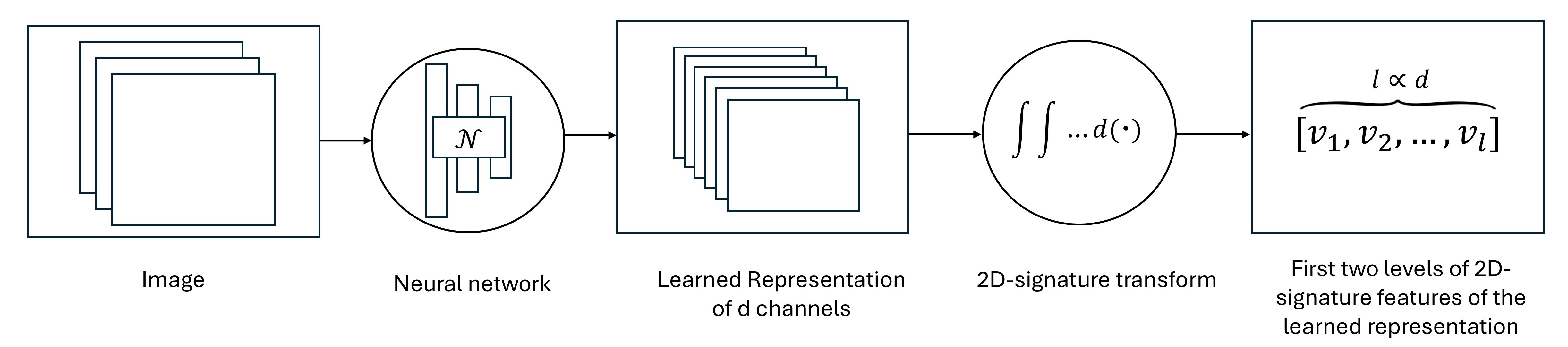}
    \caption{Process of Learned Representations Extraction and 2D-signature Calculation.}
    \label{fig:overall_framework}
\end{figure}

Learned representations refer to high-dimensional projections of data that encode relevant features or patterns inferred by a model from unlabelled data. These representations have proven efficient in helping to separate distributions when they are mixed  \cite{tran2018spectral}. This motivates using learned representations and 2D-signatures together: 
\emph{first we lift the data to a higher-dimensional feature space of representations where the data is better separated, and then we use 2D-signatures to extract faithful and compact features.} {The simplified process flow is shown in Figure \ref{fig:overall_framework}.}

Our algorithm is inspired by \cite{shao2023dimensionless} and  \cite{arrubarrena2024novelty}; the former presents an anomaly detection method for sequential data, in which the authors introduce the Mahalanobis-distance-based \emph{conformance score} defined on the range of the signature map. The conformance score gives the minimum distance between a sample and a set (in our case, this is in this space of signature features). Subsequently,  \cite{arrubarrena2024novelty} applied this method to detect radio-frequency interference in radio-astronomy signals. We propose a modified anomaly detection algorithm, called 2DSig-Detect, that adapts and extends their approach. We choose to use a Mahalanobis-equivalent distance called the covariance norm. We do not follow  \cite{shao2023dimensionless} and use the conformance score because the computational cost of their method increases with the size of the training set, raising issues when applied to tasks like online anomaly detection where swift sensor responses and {low latency} are crucial priorities. To demonstrate the advantage of our choice of distance we compare the performance of 2DSig-Detect under the conformance score and covariance norm, respectively.

We establish conditions under which our algorithm achieves both low false negative and false positive rates, providing theoretical evidence for the effectiveness of 2DSig-Detect. The effectiveness of 2DSig-Detect is evaluated against the same framework but with the conformance score (instead of the covariance norm). In addition, we evaluate our approach by benchmarking against a GMM implementation and the naive-norm. Our results consistently demonstrate superior performance of 2DSig-Detect in terms of both time cost and multiple metrics, such as false negative rate, F1-score, and  area under the receiver operating characteristic curve (AUC-ROC).

\subsection{Paper organisation}
This paper is organized as follows: Section \ref{sec:methods} presents our generalised framework for anomaly detection. In particular we introduce the covariance norm and conformance score in \ref{ssec:scores},  and following this in \ref{ssec:twoframeworks} detail how to integrate them into a semi-supervised framework given in Algorithm \ref{alg:threshold} alongside theoretical error bounds for our chosen distance metrics in Proposition \ref{prop: covnorm} and Proposition \ref{prop: conformancescore}. Section \ref{ssec:2D-signature} and \ref{ssec:lreps} describe the 2D-signature and learned representations, and then Section \ref{ssec:2dsigframeworks} explains how these are integrated together in our proposed algorithm, 2DSig-Detect, shown in Algorithm \ref{alg:2dsig-detect}.

After introducing our algorithm, we present experimental results in Section \ref{sec:experiments}. We describe the datasets used in the experiments, includng CUReT and CIFAR-10 (Section \ref{ssec:datasets}), and perform three experiments:
\begin{itemize}
\item Texture classification (Section \ref{ssec: classification})
\item Testset-level defence i.e. evasion attacks (Section \ref{ssec:testlevel})
\item Trainset-level defence i.e. backdoor attacks (Section \ref{ssec:trainlevel})

\end{itemize}

For each experiment we explain the setup used to evaluate the performance of 2DSig-Detect, including against benchmarks. We highlight the the effectiveness of our model at detecting these different types of attacks. In addition to this, in Section \ref{ssec:role_lr} we explain the role of learned representations, using the context of evasion attacks to exemplify this. 

Finally, Section \ref{sec:discussion}  concludes the paper, summarising our key findings and suggesting directions for future work in this area.
\section{Methods}\label{sec:methods}
In this section, we provide a comprehensive description of the underlying metrics that we will use to develop our 2D-signature based anomaly detection scores, along with a clear definition of how outliers are detected and an explanation of why we view one framework as preferable over the other.

\subsection{Underlying metrics for anomaly scores}\label{ssec:scores}

% [Varun]
We consider two choices of metric for 2DSig-Detect, namely the covariance norm and the conformance score. These result in two variants of 2DSig-Detect which we refer to as \emph{2DSig-Norm} (for the covariance norm) and \emph{2DSig-Conf} (for the conformance score). The choice of metric effects the computational cost of the algorithm, which influences our choice between them. This subsection introduces the covariance norm and conformance score metrics, explains why they are suitable for anomaly detection tasks, and describes how they can be computed. We defer a comparison between their computational costs to the next subsection, where we introduce the general framework in which we use these metrics to find anomalies.

\begin{algorithm}
\caption{function \emph{covariance\_norm}.\label{alg:covariance}}
\begin{algorithmic}[1]
\STATE \textbf{input:} $x\in \mathbb{R}^{d}$, a single input data;
\STATE \hspace{0.9cm} $\mathcal{C}$, a finite corpus of data with $|\mathcal{C}|$ number of elements, and its distribution is $\mathcal{L}_\mathcal{C}$.
\STATE \textbf{initialization}: a $d\times d$ covariance matrix $A=0$.
\STATE \textbf{output:} $\|x\|_{\mathcal{L}_\mathcal{C}}$, the covariance norm of $x$ under $\mathcal{L}_\mathcal{C}$.
\STATE Set $\mu=\frac{1}{|\mathcal{C}|}\sum_{y\in \mathcal{C}}y$.
\STATE
Let $\bar{Y}= \left[y-\mu\right]_{y\in  \mathcal{C}}$, an $|\mathcal{C}|\times d$ matrix.
\STATE Set $A=\frac{1}{|\mathcal{C}|-1}\bar{Y}^T\bar{Y}$.
\STATE \textbf{return}: $\|x\|_{\mathcal{L}_\mathcal{C}}:=\sqrt{ (x-\mu)^T A^{-1}(x-\mu)}$.
\end{algorithmic}
\end{algorithm}

\begin{definition}[Covariance norm] \label{def:covnorm}
Assume that there is a centred distribution $\mathcal{L}$ on a vector space $V$ with a finite second moment.  Denote by $V'$ the dual of $V$. The covariance quadratic form $\text{cov}(f,g):=\mathbb{E}^\mathcal{L}[f(\cdot)g(\cdot)]$,  where $f, g\in V'$,  induces a dual norm defined for $x\in V$ by 
\begin{footnotesize}
\begin{equation}\label{eqn:covnorm}
\|x\|_\mathcal{L}=\sup_{f:\,\text{cov}(f,f)\leq 1}f(x).
\end{equation}
\end{footnotesize}
\end{definition}
As suggested by \cite{shao2023dimensionless}, this norm in theory splits $V$ into two subspaces, based on whether $x$ is on the linear span of the support of $\mathcal{L}$, denoted by $\text{span}\big(\text{supp}(\mathcal{L})\big)$: $\|x\|_\mathcal{L}<\infty$ when $x\in \text{span}\big(\text{supp}(\mathcal{L})\big)$, and $\|x\|_\mathcal{L}=\infty$ when $x\notin \text{span}\big(\text{supp}(\mathcal{L})\big)$.   This can be seen as a natural tool for detecting outliers for distributions, as outliers typically deviate significantly from the origin (or the mean), they tend to have large, if not infinite, norm values. Note that the covariance norm coincides with the \emph{Mahalanobis distance} \cite{ghorbani2019mahalanobis} when $V$ is the Euclidean space and $\mathcal{L}$ with invertible covariance matrix. When $V=\mathbb{R}^d$, $d\in \mathbb{N}$, we are able to derive an explicit formula for Definition \ref{def:covnorm}.
\begin{prop}\label{prop: covnorm_eq} Suppose that there is a centred distribution $\mathcal{L}$ on $\mathbb{R}^d$ with a finite second moment, then for $x\in \mathbb{R}^d/\{\mathbf{0}\}$
$$\|x\|_\mathcal{L}^2=\langle x, A^{-1}x \rangle,$$
where $A_{i,j}=\text{cov}(e_i,e_j)$ and $(e_i)_{1\leq i\leq d}$ is a canonical basis for  $\mathbb{R}^d$.
\end{prop}
In practice, given a corpus of data, say $\mathcal{C}$, the covariance norm of an incoming instance $x$ to $\mathcal{C}$ can be computed through Algorithm \ref{alg:covariance}. For data $X\in \mathbb{R}^{n \times d}$, a matrix with $n$ input data vectors, the $\{\|x\|_{\mathcal{L}_\mathcal{C}}\}_{x\in X} := \sqrt{\text{diag}((X-\mu)A^{-1}(X-\mu)^T)}$, where $\text{diag}(M)$ represents the diagonal elements of $M$, and $(\mu,A)$ is pair of the empirical mean and covariance of $\mathcal{C}$.

Previous literature \cite{shao2023dimensionless} further introduce the following notation for anomaly detection purpose.
\begin{algorithm}
\begin{small}
\caption{function \emph{conformance\_score}.\label{alg:conformance}}
\begin{algorithmic}[1]
\STATE \textbf{input:} $x\in \mathbb{R}^{d}$, a single input data;
\STATE \hspace{1.1cm} $\mathcal{C}$, a finite corpus of data with $|\mathcal{C}|$ number of elements, and its distribution is $\mathcal{L}_\mathcal{C}$.
\STATE \textbf{output:} $\text{dist}(x;\mathcal{L}_\mathcal{C})$, the conformance score of $x$ to corpus $\mathcal{C}$.
\STATE \textbf{initialization}: a $d\times d$ covariance matrix $A=0$.
\STATE set $\mu=\frac{1}{|\mathcal{C}|}\sum_{y\in \mathcal{C}}y$ ;
\STATE
Let $\bar{Y}= \left[y-\mu\right]_{y\in  \mathcal{C}}$ and $Y= \left[y-x\right]_{y\in  \mathcal{C}}$ two $|\mathcal{C}|\times d$ matrices.
\STATE Set $A=\frac{1}{|\mathcal{C}|-1}\bar{Y}^T\bar{Y}$.
\STATE Set $\text{dist}(x;\mathcal{L}_\mathcal{C})=\sqrt{ \min \big(\operatorname{diag}(YA^{-1}Y^T)\big)}$.
\STATE \textbf{return}: $\text{dist}(x;\mathcal{L}_\mathcal{C})$.
\end{algorithmic}
\end{small}
\end{algorithm}

\begin{definition}[Conformance score]\label{def:conformance} Suppose that there is a distribution $\mathcal{L}$ on a vector space $V$ with a finite second moment.   The the conformance score of $x$ to $\mathcal{L}$ is defined as the distance
$$\text{dist}(x;\mathcal{L}):=\inf_{y\in \operatorname{supp}(\mathcal{L})}\|y-x\|_{\mathcal{L}}.$$
\end{definition}

The conformance score traces the closest point to $x$ in the closed support of $\mathcal{L}$ and computes the corresponding variance distance (norm).  We observe that $\text{dist}(x;\mathcal{L})=0$ if $x\in \text{supp}(\mathcal{L})$ and  $\text{dist}(x;\mathcal{L})=\infty$ if $x\notin \text{span}\big(\text{supp}(\mathcal{L})\big)$.  Indeed,  the conformance score is mainly used to quantify the distance of $x$ to the distribution $\mathcal{L}$ when $x\in \text{span}\big(\text{supp}(\mathcal{L})\big) - \text{supp}(\mathcal{L})$.  An example is an empirical distribution associated to a finite set of observations $\{y_i : y_i\in V, 1\leq i\leq N\}$.  When $N\gg \text{dim}(V)$, then it is possible that the linear span of the support of this empirical distribution is $V$ itself. In practice, given a corpus of data, say $\mathcal{C}$, the covariance norm of an incoming instance $x$ to $\mathcal{C}$ can be computed through Algorithm \ref{alg:conformance}.

\subsection{Frameworks for anomaly scores}\label{ssec:twoframeworks}
We may adopt either Definition \ref{def:covnorm} or Definition \ref{def:conformance} to measure the ``distance" of each input $x$ to the underlying distribution of $\mathcal{C}$: the shorter the distance is, the less likely this instance is an outlier. Reasonable thresholds are needed for making a decision after obtaining the covariance norm of $x$ (resp. the conformance score of $x$ to $\mathcal{L}_\mathcal{C}$): whether the score is high enough so that $x$ can be classified as anomaly. We will refer to them as the \emph{cov-norm framework} and the \emph{conf-score framework} throughout this paper.  Algorithm \ref{alg:threshold} describes one way to determine the threshold within $\mathcal{C}$, where a list of scores is generated and sorted on the validation set, and the threshold is set to be the $r$th percentile of the list, denoted by $\alpha_r$.
\begin{algorithm}
\begin{small}
\caption{function \emph{threshold}.\label{alg:threshold}}
\begin{algorithmic}[1]
\STATE \textbf{input:} $\mathcal{C}$, a finite corpus of data with distribution $\mathcal{L}_\mathcal{C}$;
\STATE \hspace{0.9cm} $r$, rth percentile.
\STATE\textbf{output:} thresholds $\alpha_{r}$ which is rth percentile;
\STATE \hspace{1.1cm} $\mathcal{C}_t$, the reference corpus.
\STATE randomly split $\mathcal{C}$ evenly into two sets, $\mathcal{C}_t$  and $\mathcal{C}_v$.
\STATE \textbf{initialization}: an empty list $l$.
 \FOR {each $x\in \mathcal{C}_v$}
\STATE append $\text{covariance\_norm}(x, \mathcal{C}_t)$ of Algorithm \ref{alg:covariance} (resp.  $\text{conformance\_score}(x, \mathcal{C}_t)$ of Algorithm \ref{alg:conformance})  to $l$;
\ENDFOR 
\STATE  \textbf{sort} $l$ in an ascending order.
\STATE Find the value in the list $l$ that is rth percentile.
\STATE \textbf{return}: $\alpha_{r}$ and $\mathcal{C}_t$.
\end{algorithmic}
\end{small}
\end{algorithm}

Algorithm \ref{alg:threshold} also provides a reference corpus for measuring distance of unseen data point. Note that Algorithm \ref{alg:threshold} can be applied with various distances by replacing Line 8 with the custom distance. Regarding the computational cost, by construction, the conf-score framework is nth-fold of that of the cov-norm framework if $n$ is the number of samples in the training set.

In the following, we present some useful probability bounds of Type-I and Type-II errors (false positive and false negative) for both frameworks. The proofs are postponed to \ref{sec:app_proofs}.
\begin{prop}[Bounds for Type-I and Type-II errors of cov-norm framework]\label{prop: covnorm} 
Suppose $\mathcal{L}_c$ is a distribution with mean and covariance pair $(\mu, A)$. Assume that A is finite and invertible.  Consider the covariance norm $\|x-\mu\| _{\mathcal{L}_c}$, where $x$ may be drawn from $\mathcal{L}_c$, or drawn from an unknown distribution $\mathcal{L}_u$, with mean and covariance pair ($\mu_u$, $A_u$). Assume that $\sqrt{d}\ll \big|A^{-\frac{1}{2}}(\mu-\mu_u)\big|/\|A^{-1/2}A^{1/2}_u\|$, where $\|B\|$ represents the matrix operator norm of a matrix $B$. 

Then for a predetermined $\delta\in \big(\sqrt{d}, |A^{-\frac{1}{2}}\big(\mu-\mu_u)|\big)$ such that
\begin{footnotesize}
$$\sqrt{d}\ll \min\Big(\delta, \frac{\big|A^{-\frac{1}{2}}(\mu-\mu_u)\big|-\delta}{\|A^{-1/2}A^{1/2}_u\|}\Big),$$
\end{footnotesize}
it holds that
\begin{footnotesize}
\begin{equation}\label{eqn:prop_covnorm_normal}
    \mathbb{P}\left(\|x-\mu\| _{\mathcal{L}_c} > \delta \mid x \sim \mathcal{L}_c\right) \leq \frac{d}{\delta^2}
\end{equation}
\end{footnotesize}
 and
 \begin{footnotesize}
  \begin{equation}\label{eqn:prop_covnorm_abnormal}
     \mathbb{P}\left(\|x-\mu\| _{\mathcal{L}_c} < \delta \mid  x\sim \mathcal{L}_u\right) \leq \frac{d\big\|A^{-\frac{1}{2}} A_{u}^{\frac{1}{2}}\big\|^2}{(|A^{-\frac{1}{2}}(\mu_u-\mu)|-\delta)^2}.
 \end{equation} 
\end{footnotesize}

 In particular, when $\delta = \frac{\big|A^{-\frac{1}{2}}(\mu_u - \mu)\big|}{1+\big\|A^{-\frac{1}{2}} A_{u}^{\frac{1}{2}}\big\|} $ , 
 \begin{footnotesize}
 \begin{equation}\label{eqn:prop_covnorm_abnormal2}
     \mathbb{P}\left(\|x-\mu\| _{\mathcal{L}_c} < \delta \mid  x\sim \mathcal{L}_u\right) \leq \frac{d}{\delta^2}.
 \end{equation} 
 \end{footnotesize}
\end{prop}

\begin{prop}[Bounds for Type-I and Type-II errors of conf-score framework]\label{prop: conformancescore} Given a corpus $\mathcal{C}$ ($\mathbb{R}^d$-valued) consisting of $n$ samples drawn independently from a distribution $\mathcal{L}_c$, with ground truth mean and covariance $(\mu, A)$. Assume that $A$ is invertible. Given the unknown instance $x$ and consider the conformance score $\text{dist}(x; \mathcal{L}_c)$.

For a pair $(d,\delta)$ such that
\begin{footnotesize}
$$\sqrt{2d}<\delta<\big|A^{-\frac{1}{2}}(\mu_u - \mu)\big| \text{\ and\ }  
\frac{d\big\|\Delta_A^{\frac{1}{2}}\big\|^2}{\left(\Delta_\mu-\delta\right)^2}<1,$$
\end{footnotesize}
where ($\mu_u$, $A_u$) are the mean and covariance pair for the unknown distribution $\mathcal{L}_u$, and $\Delta_A:=A^{-\frac{1}{2}} A_u A^{-\frac{1}{2}}+I$, then
\begin{footnotesize}
\begin{equation}\label{eqn:prop_conformancescore_normal}
    \mathbb{P}\left(\operatorname{dist}\left(x ; \mathcal{L}_c\right)>\delta \mid x \sim \mathcal{L}_c\right)  \leqslant \left(\frac{ 2d}{\delta^2}\right)^n,
\end{equation}
\end{footnotesize}
and
\begin{footnotesize}
\begin{equation}\label{eqn:prop_conformancescore_abnormal}
    \mathbb{P}\left(\operatorname{dist}\left(x ; \mathcal{L}_c\right)<\delta \mid x \sim \mathcal{L}_u\right)  \leqslant 1-\left(1-\frac{ d\big\|\Delta_A^{\frac{1}{2}}\big\|^2}{\left(\Delta_\mu-\delta\right)^2}\right)^n.
\end{equation}
\end{footnotesize}

    \end{prop}
Proposition \ref{prop: conformancescore} gives an increasing up-bound for Type-II error, which is not expected. Note that 
\begin{footnotesize}
\begin{equation}\label{eqn:meandiff}
    \|\mu_u\|_{\mathcal{L}_c}=|A^{-1/2}(\mu_u-\mu)|,
\end{equation}
\end{footnotesize}
the \emph{mean difference} under the covariance norm of $\mathcal{L}_c$, appears in both propositions. Note that under certain conditions, the mean difference appears in the formula of Kullback-Leibler divergence \cite{bishop2006pattern}. Proposition \ref{prop: covnorm} and Proposition \ref{prop: conformancescore} both suggest that the key features for these two frameworks to work well are the dimensionality as well as how separated the two distributions are (mainly measured by mean difference).

Note that we prefer the covariance norm over the conformance score because under some circumstances, it will give similar performance but the former  saves computational costs. In the following, we will introduce a transform on images to capture their distribution while reducing dimensionality (Section \ref{ssec:2D-signature}), and find ways to leverage the distribution difference (Section \ref{ssec:role_lr}).
\subsection{2D-Signature}\label{ssec:2D-signature}
We will give a brief introduction to 2D-signature features for color images and generalise them for high-dimensional representations. Define $[n]:=\{1,2,\ldots,n\}$.  For an image of size $(N, M,d)$, with $N,M,d\in\mathbb{N}$,  i.e. with $M\times N$ many pixels and $d$ many channels, the image data is defined as follows.
\begin{definition}[Continuous and discrete image data]\label{def:image_data}
    For $N,M, d \in\mathbb{N}$,  the image information can be represented via a map $\mathbf{x} : \{1, \dots, N\} \times \{1, \dots, M\} \to [0,1]^d $, where $d$ represents the number of channels of an image,  $\mathbf{x}^i_{k_1,k_2}$ denotes the value of $ i $-th coordinate of the image, $i\in [d]$, at the location of pixels $ (k_1, k_2) $ with $k_1 \in [N]$ and $k_2\in [M]$. Through interpolation, the (normalised) image data can be well-represented via a piecewise continuous map $x:[0,1]^2 \to [0,1]^d$, where $x^i(s,t)$ represents the value of $ i $-th coordinate of the image at the location $ (s, t)$.  
We denote the space of image data as $\mathcal{Z}_d$.
\end{definition}
Note that $(N,M,\cdot)$ is also known as image resolution. Definition \ref{def:image_data} coincides with color images when $d=3$, where the three coordinates represents red/green/blue (RGB) channels respectively, and reduces to gray images when $d=1$. 
 
Image data is parameterised by two components, one for each dimension of the data, which are referred as $s$ and $t$ in the following context.
 
 \hspace{0pt}\\\textbf{The differentials and the definition} The key components of 2D-signature are the area increments we want to examine for image data. We shall adopt the notation of \cite{zhang2022two} for compact expressions:
 \begin{footnotesize}
 \begin{equation}
     \mathrm{d} x^i_{s,t}:=\frac{\partial^2 x^{i}(s,t)}{\partial s \partial t}\, \mathrm{d}s\mathrm{d}t,\qquad i\in [d], 
 \end{equation}
 \end{footnotesize}
\begin{footnotesize}
 \begin{equation}
     \hat{ \mathrm{d}} x^{ij}_{s,t}:=\frac{\partial x^{i}(s,t)}{\partial s }\frac{\partial x^{j}(s,t)}{\partial t }\, \mathrm{d}s\mathrm{d}t,\qquad i\in [d].
 \end{equation}
 \end{footnotesize}
Both differentials provide different insights into the behavior of the function \( x^i(s,t) \) in the \( s \)-\( t \) plane: the first differential $\mathrm{d} x^i_{s,t}$ measures how the rate of change with respect to one variable varies as the other variable changes and is sensitive to the curvature or bending of the function surface in the \( s \)-\( t \) plane; and the second one measures the interaction of the independent rates of change in \( s \) and \( t \), and captures how changes in one direction affect changes in the other direction multiplicatively. The choice of these two differentials comes from the fact that, for any sufficiently smooth function $f:\mathbb{R}^d\to \mathbb{R}$, if considering the rectangle increment of it over $[s_1,s_2]\times [t_1,t_2]$, then
\begin{footnotesize}
\begin{align}\label{eqn:f_area}
    \begin{split}
        &f(x_{s_2,t_2})-f(x_{s_1,t_2})-f(x_{s_2,t_1})+f(x_{s_1,t_1})\\
        &=\sum_i\int\int_{[s_1,s_2]\times [t_1,t_2]}\frac{\partial f(x_{s,t})}{\partial x^i}\,\mathrm{d}x^i_{s,t}+\sum_{i,j}\int\int_{[s_1,s_2]\times [t_1,t_2]}\frac{\partial^2 f(x_{s,t})}{\partial x^i\partial x^j}\,\hat{\mathrm{d}}x^{ij}_{s,t}.
    \end{split}
\end{align}
\end{footnotesize}
Equivalently, the area feature of $f$ acting on image $x$ can be represented through the linear combination of $\mathrm{d}x^{i}_{s,t}$ and $\mathrm{d}x^{ij}_{s,t}$. As the integrals with respect to $\mathrm{d}x^{i}_{s,t}$ or $\mathrm{d}x^{ij}_{s,t}$ can be computed, it would be expected that through a Taylor expansion, the area effect of an aribitrary $f$ can be approximated as
\begin{footnotesize}
\begin{align}\label{eqn:f_taylor}
    \begin{split}
        &f(x_{s_2,t_2})-f(x_{s_1,t_2})-f(x_{s_2,t_1})+f(x_{s_1,t_1})\\
        &\approx \sum_ic_i\int\int_{[s_1,s_2]\times [t_1,t_2]}\,\mathrm{d}x^i_{s,t}+\sum_{i,j}C_{i,j}\int\int_{[s_1,s_2]\times [t_1,t_2]}\,\hat{\mathrm{d}}x^{ij}_{s,t}\\
        &\quad+\sum_{i,j}c_{i,j}\int_{s_1<v<s<s_2}\int_{ t_1<u<t<t_2}\,\mathrm{d}x^j_{v,u}\,\mathrm{d}x^i_{s,t}\\
        &\quad+\sum_{i,j,\bar{i},\bar{j}}C_{i,j,\bar{i},\bar{j}}\int_{s_1<v<s<s_2}\int_{ t_1<u<t<t_2}\,\hat{\mathrm{d}}x^{ij}_{v,u}\,\hat{\mathrm{d}}x^{ij}_{s,t}\\
        &\quad +\cdots,
    \end{split}
\end{align}
\end{footnotesize}
where the constants are to be determined through learning from the data. However, as commented in \cite{zhang2022two}, the number of terms will explode when we expand the Taylor series, this explains why we are not aware of any extension up to an arbitrary order in literature. We therefore would follow \cite{zhang2022two}, restricting to the differential 
\begin{footnotesize}
 \begin{equation}\label{eqn:dx}
     \mathrm{\mathbf{d}}x_{s,t}:=
 \begin{bmatrix}
    \mathrm{d}x^1_{s,t} \\
    \ldots \\
    \mathrm{d}x^d_{s,t}\\
       \hat{ \mathrm{d}}  x^{11}_{s,t} \\
              \hat{ \mathrm{d}}  x^{22}_{s,t} \\
    \ldots \\
     \hat{ \mathrm{d}} x^{dd}_{s,t}\\ 
\end{bmatrix},
 \end{equation}
 \end{footnotesize}
and define the 2D-signature up to the second level only. 
\begin{definition}[2D-signature for images, up to the second level]\label{def:2d_signature}
Let $x: [0,1]^2 \rightarrow [0,1]^d$ be continuous image data such that the following integration makes sense. For any closed internals $J_1, J_2\subset [0,1]$, define 
\begin{footnotesize}
\begin{align}\label{eqn:2dsig_n}
    \begin{split}
      &\big(S_{J_1\times J_2}(x)\big)^{p} \\
      &=\underset{\underset{t_{1} < \dots < t_{n}, t_i \in J_2}{s_{1} < \dots < s_{n}, s_i \in J_1}} { \int \dots \int} \mathrm{\mathbf{d}}x_{s_1,t_1} \otimes \dots \otimes \mathrm{\mathbf{d}}x_{s_n,t_n}\in (\mathbb{R}^{2d})^{\otimes p},  
    \end{split}
\end{align}
\end{footnotesize}
and set 
\begin{footnotesize}
\begin{equation}\label{eqn:2dsig_full}
    S_{J_1\times J_2}(x)= \big(1, S^1_{J_1\times J_2}(x),  S^2_{J_1\times J_2}(x) \big)
\end{equation}
\end{footnotesize}
while $S^p_{J_1\times J_2}(x)=\big(S_{J_1\times J_2}(x)\big)^{p}$ is the degree $p$ or level $p$ of the signature. We write $S(x)$ short for $S_{J_1\times J_2}(x)$ when the intervals are obvious or when we do not need to specify.  
\end{definition}
For instance, when $J_1=J_2=[0,1]$ and $p=1$, 
\begin{footnotesize}
 $$S^1(x)=\int\int_{[0,1]^2}\mathrm{\mathbf{d}}x_{s,t}=
 \begin{bmatrix}
    \int\int_{[0,1]^2}\mathrm{d}x^1_{s,t} \\
    \ldots \\
   \int\int_{[0,1]^2} \mathrm{d}x^d_{s,t}\\
   \int\int_{[0,1]^2} \hat{ \mathrm{d}}  x^{11}_{s,t} \\
    \ldots \\
   \int\int_{[0,1]^2}\hat{ \mathrm{d}} x^{dd}_{s,t}\\ 
\end{bmatrix}.$$
\end{footnotesize}
The definition coincides with 2D-id-signature in \cite{diehl2024signature} if $\mathrm{\mathbf{d}}x_{s,t}$ in Eqn. \eqref{eqn:dx} only include the first half, ie , $\{\mathrm{d}x_{s,t}^i\}_{i=1}^d$, where the extension can be made up to an arbitrary order and its algebraic structures are further explored. We do not follow \cite{diehl2024signature} because the integrals involving $\hat{ \mathrm{d}}  x^{ii}_{s,t}$ capture nonlinear effect of pixels from the first level (see \textbf{Discretization} below) and play a vital role in learning tasks \cite{zhang2022two}.

 The number of features in Eqn. \eqref{eqn:2dsig_full}  would be $2d$ for level 1, and $(2d)^2$ for level 2. For image-related tasks, 
  especially in later layers, where image dimensions have often been reshaped to have more than their 3 initial layers, interaction between channels is not so crucial for our learning purpose \cite{szegedy2016rethinking}. We may consider the components that reinforces the effect on the same channel in the second level, i.e., we consider $\mathrm{d}x^{i}\mathrm{d}x^i$ and $\hat{\mathrm{d}}x^{ii}\hat{\mathrm{d}}x^{ii}$. Thus the number of features would be $2d$ as well for this special collection of level 2.

  \hspace{0pt}\\\textbf{Discretization} As we have two basic notations, we shall give examples on level 1 only for illustration. 
  Note that for any $0\leq s_1<s_2\leq 1$ and $0\leq  t_1<t_2\leq 1$, it holds that
  \begin{footnotesize}
  \begin{align}\label{eqn:forwarddiff1}
    \begin{split}
        &\int_{[t_1,t_2]}\int_{[s_1,s_2]}\frac{\partial^2 x^{i}(s,t)}{\partial s \partial t}\, \mathrm{d}s\mathrm{d}t=x^{i}(s_2,t_2)-x^{i}(s_2,t_1)-x^{i}(s_1,t_2)+x^{i}(s_1,t_1).
        % &\approx x^{i}(s+h,t+h)-x^{i}(s+h,t)-x^{i}(s,t+h)+x^{i}(s,t).   
    \end{split}
\end{align}
\end{footnotesize}
Therefore
\begin{footnotesize}
\begin{align}\label{eqn:2dlevel11}
\begin{split}
&\int\int_{[0,1]^2} \frac{\partial^2 x^{i}(s,t)}{\partial s \partial t}\mathrm{d}s\mathrm{d}t 
=\mathbf{x}^i_{N, M} - \mathbf{x}^i_{N,1} - \mathbf{x}^i_{1, M} + \mathbf{x}^i_{1, 1 }.
\end{split}
\end{align}
\end{footnotesize}

For a small stepsize $h\ll 1$, using forward difference gives 
\begin{footnotesize}
  \begin{align}\label{eqn:forwarddiff2}
    \begin{split}
        & \frac{\partial x^{i}(s,t)}{\partial s }\frac{\partial x^{i}(s,t)}{\partial t }\, \mathrm{d}s\mathrm{d}t\\
        &\approx (x^{i}(s+h,t)-x^{i}(s,t))\big(x^{i}(s,t+h)-x^{i}(s,t)\big).
    \end{split}
\end{align}
\end{footnotesize}
Thus
\begin{footnotesize}
\begin{align}\label{eqn:2dlevel12}
\begin{split}
&\int\int_{[0,1]^2} \frac{\partial x^{i}(s,t)}{\partial s }\frac{\partial x^{i}(s,t)}{\partial t }\mathrm{d}s\mathrm{d}t\\
& \approx \sum^{N-1}_{k_1 = 1} \sum^{M-1}_{k_2 = 1}\Big(\left(\mathbf{x}^i_{k_1 + 1, k_2} - \mathbf{x}^i_{k_1, k_2}\right) \left(\mathbf{x}^i_{k_1, k_2+1} - \mathbf{x}^i_{k_1, k_2 }\right)\Big).
\end{split}
\end{align}
\end{footnotesize}
Clearly, Level 1 of the 2D-signature contains both linear effects and nonlinear effects of pixel information. Level 2 further reinforces these effects through nonlinear operations (iterated integrals). One may find the discretization for level 2 of the 2D-signature in \ref{sec:app_dis}.

Note that there exists another type of 2D-signature, called \emph{Type II} 2D-signature in this paper, whose expression can be derived by iterating the Jacobian minor as defined in \cite{giusti2022topological}.Type II 2D-signature is not considered in this article because its expression focuses solely on cross-channel interactions, thereby excluding its application to grayscale images. However, when a $d$-channel image is expanded to a $(d+2)$-channel image by padding at positions $(s,t)$, it is claimed that the corresponding Type II 2D-signature feature has the characteristic property: the law of random images can be completely determined by the expected 2D-signature of the expanded images (see \cite[Theorem 6]{giusti2022topological}).

\subsection{Learned representations}
\label{ssec:lreps}
It has been shown that learned representations (e.g. via a neural network) may better separate distributions \cite{tran2018spectral}.  Inspired by this observation, in practice we may 
consider lifting the space where a corpus $\mathcal{C}$ lives in to a higher-dimensional space   $\mathcal{R}(\mathcal{C})$,
where 
$\mathcal{R}$ is a carefully-chosen learned representation and $\mathcal{R}(\mathcal{C}):=\{\mathcal{R}(y), y\in \mathcal{C}\}$.

Later in Section \ref{ssec:role_lr}, we explore the capability of different learned representations in separating distributions. For a well-trained classification model, it's expected that the later layers are adept at distinguishing different image clusters. In contrast, for anomaly detection, the earlier layers of this trained model are more effective at emphasising the differences between benign and polluted images, rather than the later layers.

\subsection{2DSig-Detect}
\label{ssec:2dsigframeworks}
\begin{figure}[!t]
    \centering
    \includegraphics[width=6in]{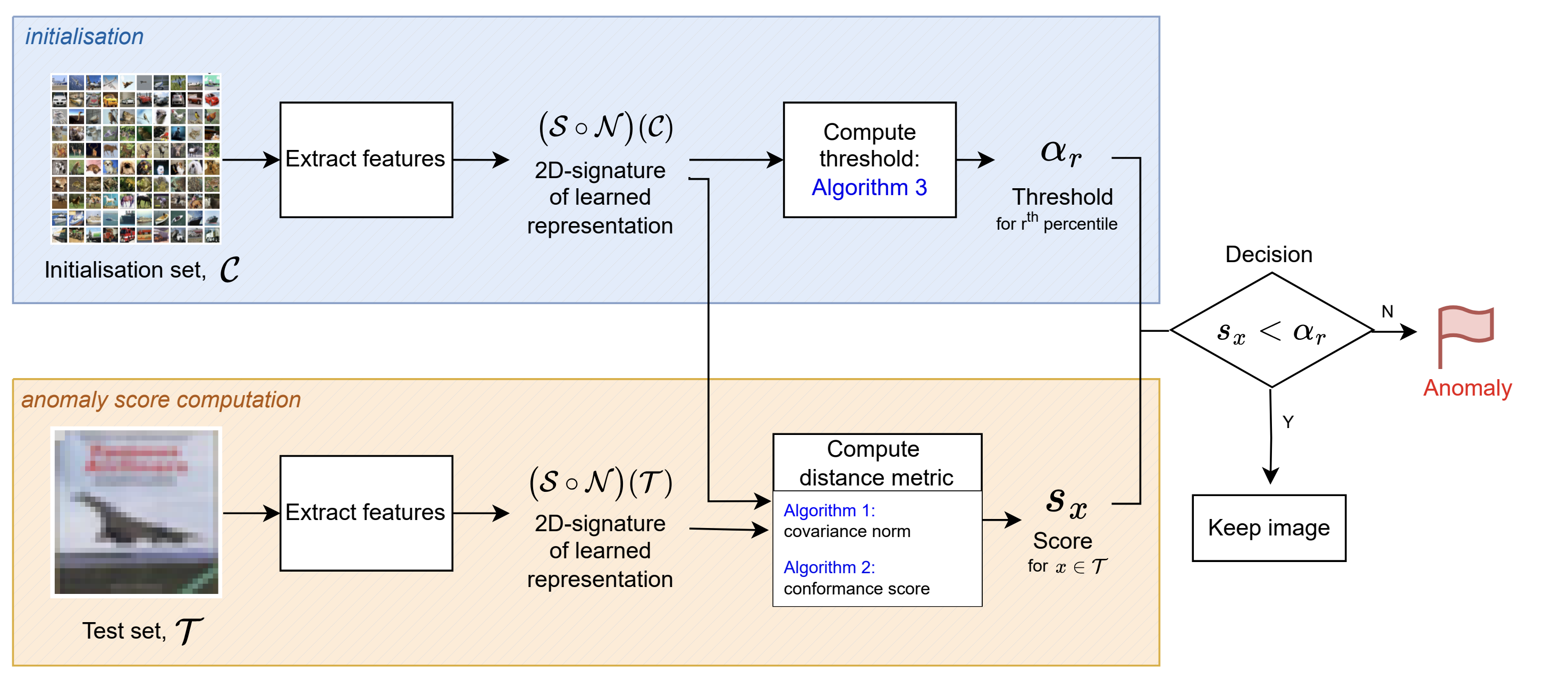}
    \caption{Flowchart of the proposed 2DSig-Detect framework, as outlined in Algorithm \ref{alg:2dsig-detect}.}
    \label{fig:flowchart}
\end{figure}

Considering its ability in capturing effects of image data, the proposed methods are therefore embedding 2D-signature features in the base frameworks: we extract a higher-dimensional representation of an image $x$ through a neural network transform $\mathcal{R}$, then apply the 2D-signature transform $S^i$ ($i=1$ or $2$) to $\mathcal{R}(x)$, and finally feed these 2D-signature features $S^i(\mathcal{R}(x))$ into one of the two base frameworks (Algorithm \ref{alg:covariance} or \ref{alg:conformance}) to get a score. The threshold can be obtained via applying Algorithm \ref{alg:threshold} to the transformed train set  $(S^i\circ \mathcal{R})(\mathcal{C})$. Together, this results in Algorithm \ref{alg:2dsig-detect}, which we call \emph{2DSig-Detect}. The flowchart of this process is summarised and presented in Fig. \ref{fig:flowchart}. 
\begin{algorithm}[!t]
\begin{small}
\caption{function \emph{2DSig-Detect}.\label{alg:2dsig-detect}}
\begin{algorithmic}[1]
\STATE \textbf{input:} $\mathcal{C}$, a finite corpus of data with distribution $\mathcal{L}_\mathcal{C}$;
\STATE  $\mathcal{T}$,  a test set of data;\\
\STATE  $\mathcal{N}$, a neural network model providing feature representation $\mathcal{R}_\phi$;\\
\STATE  $\mathcal{S}$, other transform such as 2D-signature;\\
\STATE  $r$, rth percentile.
\STATE\textbf{output:} $\mathcal{T}_b$, a subset of $\mathcal{T}$.
\STATE extract features $\big(\mathcal{S\circ \mathcal{N}}\big)(\mathcal{C})$ of $\mathcal{C}$.
\STATE apply Algorithm \ref{alg:threshold} to $\big(\mathcal{S\circ \mathcal{N}}\big)(\mathcal{C})$: randomly split $\big(\mathcal{S\circ \mathcal{N}}\big)(\mathcal{C})$ evenly into two sets, $\big(\mathcal{S\circ \mathcal{N}}\big)(\mathcal{C})_t$  and $\big(\mathcal{S\circ \mathcal{N}}\big)(\mathcal{C})_v$, and get a threshold $\alpha_{r}$.
\STATE extract features $\big(\mathcal{S\circ \mathcal{N}}\big)(\mathcal{T})$ of $\mathcal{T}$.
\STATE initialise an empty set $\mathcal{T}_b$.
\FOR{each $x\in \mathcal{T}$}
\STATE assign a score $s_x$ to $x$ by applying either Algorithm \ref{alg:covariance} or Algorithm \ref{alg:conformance} to $\big(\mathcal{S\circ \mathcal{N}}\big)(x)$ and $\big(\mathcal{S\circ \mathcal{N}}\big)(\mathcal{C})_t$;
 \IF{$s_x<\alpha_r$}
\STATE $\mathcal{T}_b\leftarrow\mathcal{T}_b+\{x\}$;
\ENDIF
\ENDFOR
\STATE \textbf{return}: $\mathcal{T}_b$.
\end{algorithmic}
\end{small}
\end{algorithm}

Throughout this paper, we refer to 2DSig-Detect (Algorithm \ref{alg:2dsig-detect}) with the covariance norm and conformance score as \emph{2DSig-Norm} and \emph{2DSig-Conf}, respectively. We emphasize that \emph{2DSig-Norm} and \emph{2DSig-Conf} are versions of the same algorithm: 2DSig-Detect. The only change is the choice of distance metric that governs the resulting anomaly score. We propose, for computational efficiency reasons that the experimental section (Section \ref{sec:experiments}) will make clear, to favour \emph{2DSig-Norm} over \emph{2DSig-Conf}; instead we use the latter as a benchmark. In addition to this, we benchmark against our base framework (Algorithm \ref{alg:threshold}) with flattened pixel features, which we refer to as the \emph{naive-norm} and \emph{naive-conf} approach, respectively. We also compare against the Gaussian mixture method with 2D-signature features, termed \emph{2DSig-GMM}.
\section{Applications}\label{sec:experiments}
In this section, we show how the proposed 2DSig-Norm can be used in applications, including classification and defence against both backdoor and evasion attacks, and present the performance of \acrshort{name} in comparison to the 2DSig-Conf and naive-norm\footnote{The computational cost for naive-conf is sometimes very high so we ignore it in most of the experiments.} frameworks using standard image datasets. All experiments were conducted on a server running Ubuntu 22.04.4 LTS (Jammy Jellyfish) 64-bit. The server features 755.3 GiB of memory and an Intel® Xeon® Gold 6140 CPU @ 2.30GHz with 72 cores, providing a robust computational environment for the applications. 

\subsection{Datasets}
\label{ssec:datasets}
\begin{figure}[!ht]
    \centering
    \includegraphics[width=0.7\textwidth]{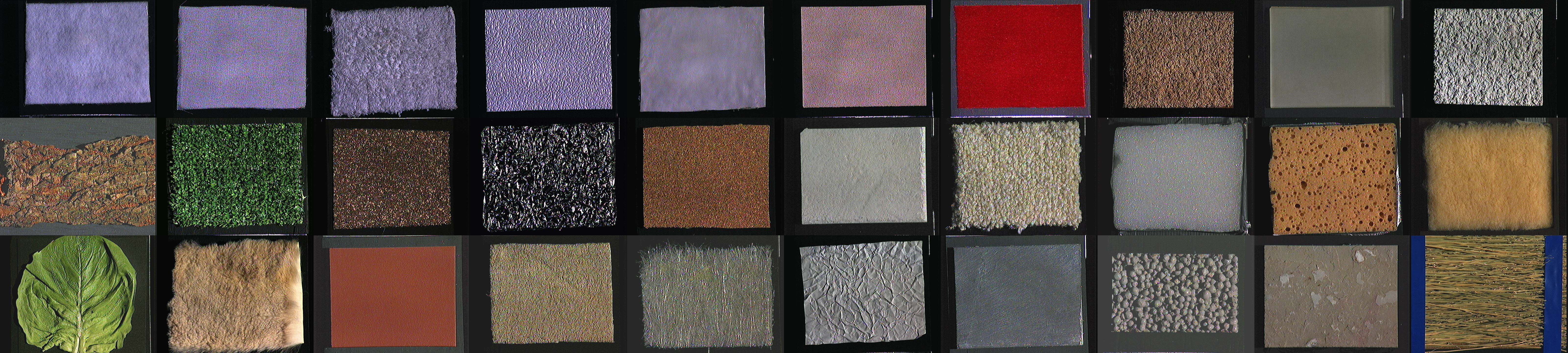}
    \caption{The first 30 images from the CUReT database.}
    \label{fig:curet_examples}
\end{figure}

\textbf{CUReT} (Columbia-Utrecht Reflectance and Texture Database \cite{CUReT}) is a data set of 5612 texture images in 61 classes. As the original images contain black boundaries, they are manually cropped to a size of $240\times 320$. Figure \ref{fig:curet_examples} shows examples of the first 30 images from the CUReT database. Following  \cite{zhang2022two}, 42 textures from this dataset are used for our first experiment.

\begin{figure}[!ht]
\centering
\includegraphics[width=0.5\textwidth]{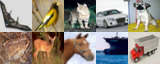}
\caption{Randomly selected examples from each class in the CIFAR-10 dataset.}
\label{fig:cifar10_example}
\end{figure}

\hspace{0pt}\\ \textbf{Cifar10 \cite{Krizhevsky09learningmultiple}} is a dataset consisting of 60,000 (32,32,3) color images in 10 classes, with 6,000 images per class. The dataset is divided into 50,000 training images and 10,000 test images. The 10 classes are airplane, automobile, bird, cat, deer, dog, frog, horse, ship, and truck. CIFAR-10 is a standardized and widely used dataset in machine learning and computer vision research \cite{simonyan2014very}\cite{huang2017densely}. It provides a diverse set of images, making it a valuable resource for evaluating algorithms. Figure \ref{fig:cifar10_example} shows randomly selected examples from each class in the CIFAR-10 dataset.

\subsection{Texture image classification}\label{ssec: classification}
The 2D-signature transform achieves a superior performance in texture image classification on CUReT as a informative feature  \cite{zhang2022two}. In the light of this observation, we identify texture image classification as a special anomaly detection problem: if we choose the nth texture, i.e. class 42 as the target texture, the remaining 41 kinds of textures would be treated as anomalous to the target class. 

\begin{figure}[h!]
    \centering
    \includegraphics[width=0.5\textwidth]{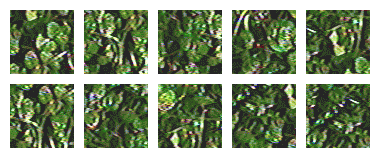}
    \caption{Examples of 64x64 patches from the Soleirolia plant class in the CUReT dataset.}
    \label{fig:Soleirolia plant}
\end{figure}

\hspace{0pt}\\\textbf{Data processing} To prevent overfitting, each of these 42 textures is divided into left and right halves, and one of the halves are randomly drawn, with one portion allocated for generating the training set, while the other half is reserved for testing purposes. We randomly sample 200 (resp. 100) images of size $(m \times m)$ from each train (resp. test) texture, $m=32, 64$. Fig. \ref{fig:Soleirolia plant} presents a selection of 64x64 patches extracted from Soleirolia plant class of the CUReT dataset. Among the 200 samples in each train class, half of them are randomly drawn as the validation set for generating the score distribution and determining the threshold $\alpha_r$, for two different thresholds, $r=80$ and $r=90$th percentiles. The blue curve in the bottom subplot of Fig. \ref{fig:l1_con_cov} gives an example of the cumulative distribution from 2DSig-Norm, where we can identify the 80th  (resp. 90th) percentile as the threshold.

\subsubsection{Experiment set-ups}
For each class, we generate three sets: scores from validation set (referred as \emph{validation scores}), scores from test set of the same class (benign set), referred as \emph{benign scores}, and scores from test sets of the remaining 41 classes (anomaly sets), referred as \emph{anomaly scores}.  The threshold shall be determined from the first set, and apply to the latter two. However, this introduces a highly unbalanced score set.  For a fair comparison, we design two kinds of experiments for different comparison purposes.

\hspace{0pt}\\\textbf{Experiment 1.} To compare performance between the two base frameworks introduced in Section \ref{ssec:twoframeworks}, either with or without the 2D-signature, we will consolidate scores from all 42 validation sets into a unified validation score set. Similarly, we will get a unified benign score set and a unified anomaly score set. The threshold is determined on the unified validation score set, where one may sort the scores in an ascending order, and choose the $r$-th  percentile as the threshold, and then apply it to the unified benign score set and the unified anomaly score set.
Though the distributions of different classes on the training set may be different, Eqn. \eqref{eqn:prop_covnorm_normal} (resp. Eqn. \eqref{eqn:prop_conformancescore_abnormal}) ensures that taking the covariance norm (resp. conformance score) is a way of \emph{standardising} distributions: subtracting the dataset mean from a data point and then multiplied by the inverse of the covariance matrix.

\hspace{0pt}\\\textbf{Experiment 2.} To compare performance between 2DSig-Norm and other unsupervised methods, we 
% randomly drawn 100 anomaly scores per class 
randomly selected 100 images from different classes in the segmented anomaly sets, so that the number of benign scores equals the number of anomaly scores. Anomaly detection is conducted per class and the mean metric plus standard deviation is recorded.

\hspace{0pt}\\\textbf{Metrics.} In Experiment 1, F1-score (for $r=80$ on validation), the area under the receiver operating characteristic curve (AUC)  and time cost are used for model comparison. In Experiment 2, accuracy, F1-score and true positive rate (TPR) are used for model comparison. In this paper, ``positives" refer to anomalies, while ``negatives" refer to benign instances.

\subsubsection{Results.}
\textbf{Experiment 1.} 
For $m=32$, we compare the proposed 2DSig-Norm against 2DSig-Conf as well as the naive-norm, in terms of F1-score (for $r=80$), AUC and time cost. We examine the performance of level 1 and level 2 of the 2D-signature respectively. Each experiment recorded in Table \ref{tab:size32_comparision} is repeated 10 times across different random seeds, where the mean and standard deviation are recorded. While both the 2DSig-Norm and the 2DSig-Conf exhibit significant advantages over naive-norm across all metrics, the 2DSig-Norm demonstrates a slight performance advantage over the 2DSig-Conf in both F1-score and AUC metrics, achieving this with only 1/7 of the time required by 2DSig-Conf. 

\begin{figure}[!t]
     \centering
    \includegraphics[width=6.0in, trim=120pt 65pt 120pt 60pt, clip]{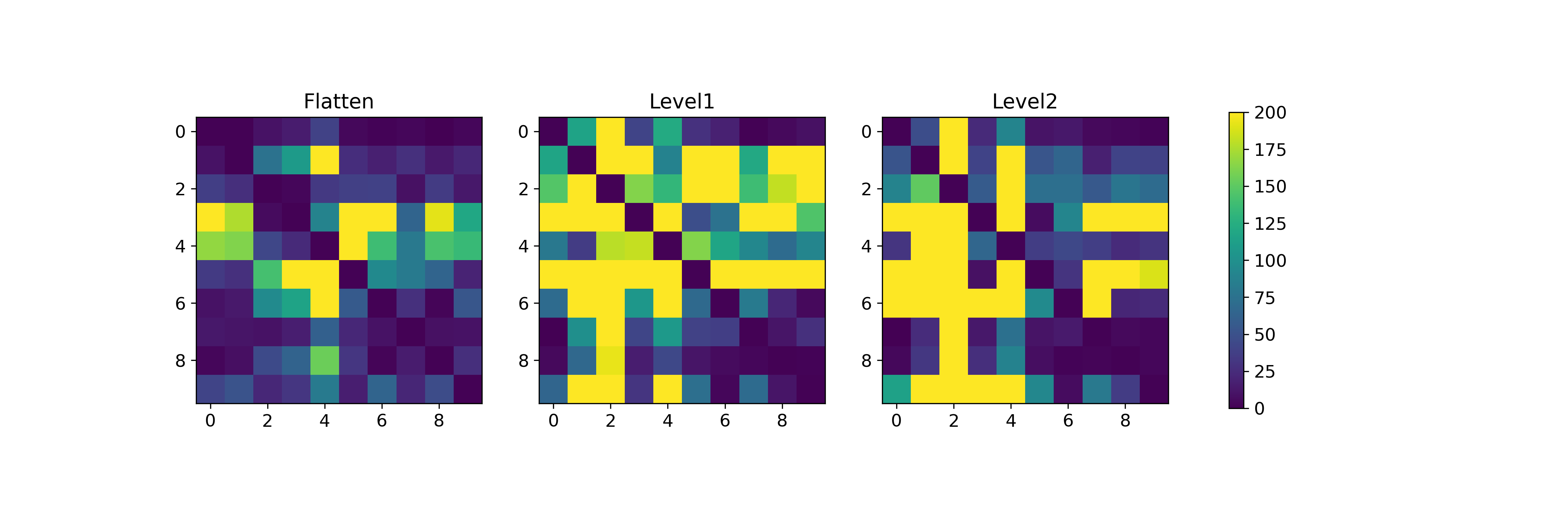}
    \caption{Mean differences calculated via \eqref{eqn:meandiff} between the ith class of the training set and the jth class of the test set, $1\leq i,j\leq 10$, on different features (Left: the flatten representation; Middle: the level 1 feature of 2D-signature; Right: the level 2 feature of 2D-signature).}
    \label{fig:heatmap}
\end{figure}

To understand the performance difference, the mean difference between classes is also examined for different features, namely, the flatten feature, i.e., the flattened pixel information, level 1 of 2D-signature and level 2 of the 2D-signature in Fig. \ref{fig:heatmap}. The heat map of level 1 of the 2D-signature shows the largest discrepancy among classes, which coincides with Table \ref{tab:size32_comparision}, where under the same base framework, the AUC of level 1 is slightly higher than the one of level 2.
\begin{table}[]
   \caption{Performance comparison for different features via either conf-score or cov-norm framework on CUReT dataset with sampling image size $(32, 32, 3)$. Note that F1-score is captured at $\alpha_r$ with $r=80$th percentile.}
    \label{tab:size32_comparision}
        \centering
        \begin{footnotesize}
\begin{tabular}{cc|c|c|c}
\hline
\hline
\multicolumn{2}{l|}{}                                      & F1-score       & AUC            & time (s)             \\ \hline
\multicolumn{1}{l|}{\multirow{2}{*}{\bf 2DSig-Norm}} & Level1 & $0.92\pm 0.00$ & {$\color{red}0.88\pm 0.01$} & {\color{red}$0.62\pm 0.00$}    \\ \cline{2-5} 
\multicolumn{1}{l|}{}                            & Level2 & {\color{red}$0.94\pm 0.01$} & $0.86\pm 0.01$ & $0.73\pm 0.01$    \\ \hline
\multicolumn{1}{l|}{\multirow{2}{*}{2DSig-Conf}} & Level1 & $0.92\pm 0.00$ & $0.87\pm 0.01$ & $4.30\pm 0.00$    \\ \cline{2-5} 
\multicolumn{1}{l|}{}                            & Level2 & $0.93\pm 0.01$ & $0.85\pm 0.01$ & $9.71\pm 0.01$    \\ \hline
\multicolumn{2}{c|}{naive-norm}                            & $0.76\pm 0.01$ & $0.72\pm 0.01$ & $716.20\pm 51.80$ \\ \hline
\hline
\end{tabular}
\end{footnotesize}
\end{table}

\begin{figure}[!t]
     \centering
      \makebox[\textwidth][c]{%
          \includegraphics[width=2.5in]{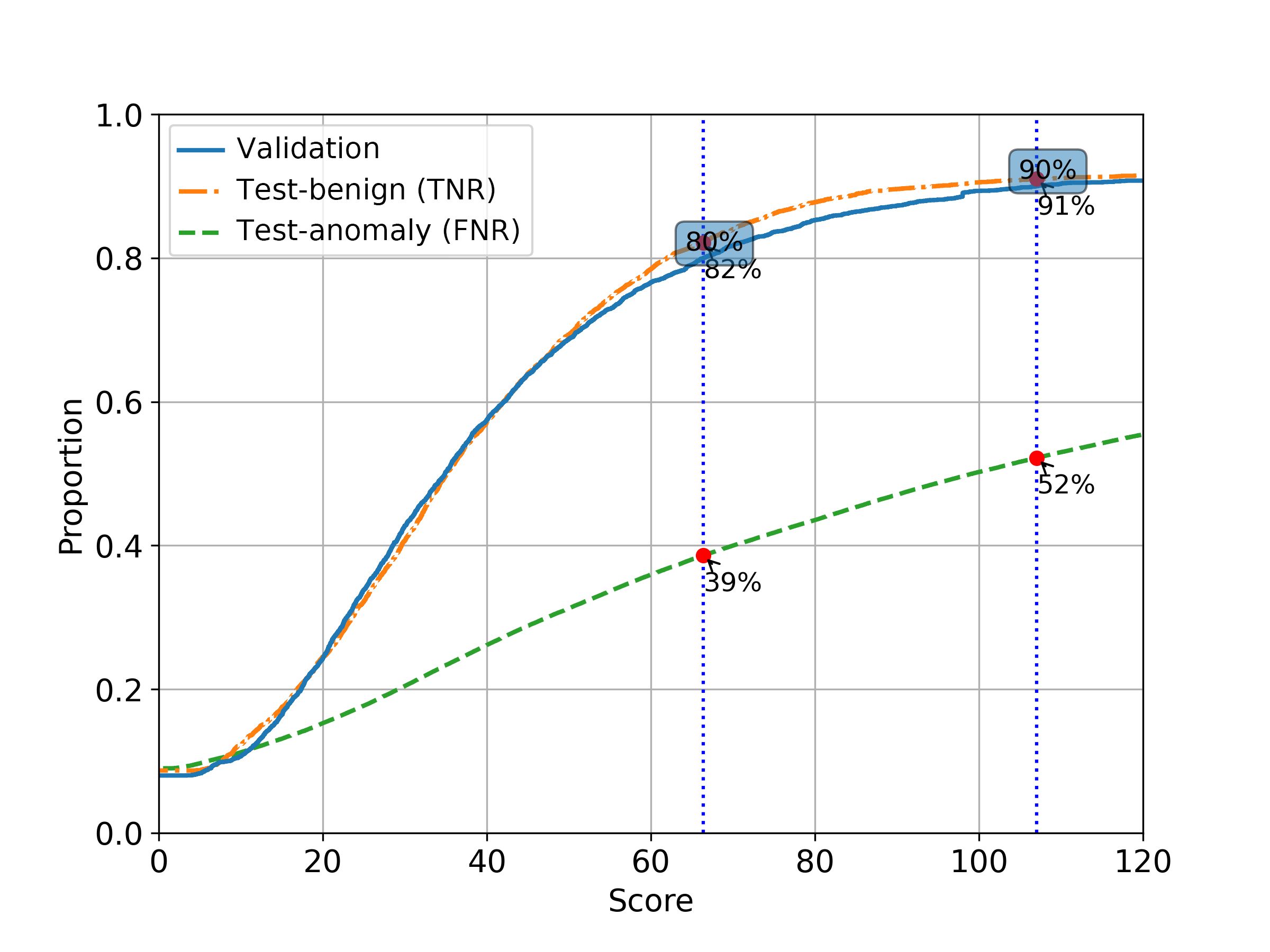}
          \hspace{-0.3in}% 压缩水平间距
        \includegraphics[width=2.5in]{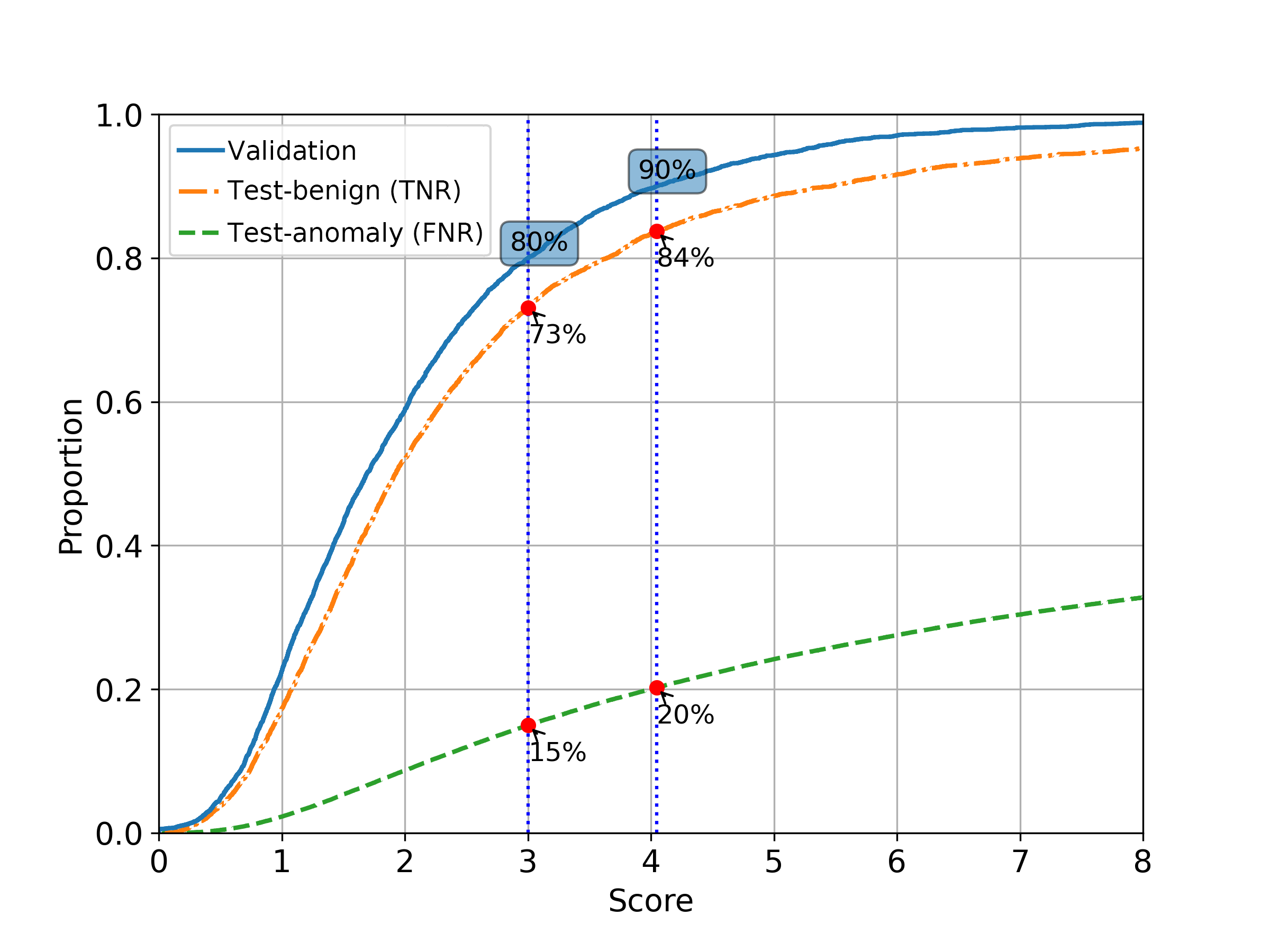}
        \hspace{-0.3in}% 压缩水平间距
     \includegraphics[width=2.5in]{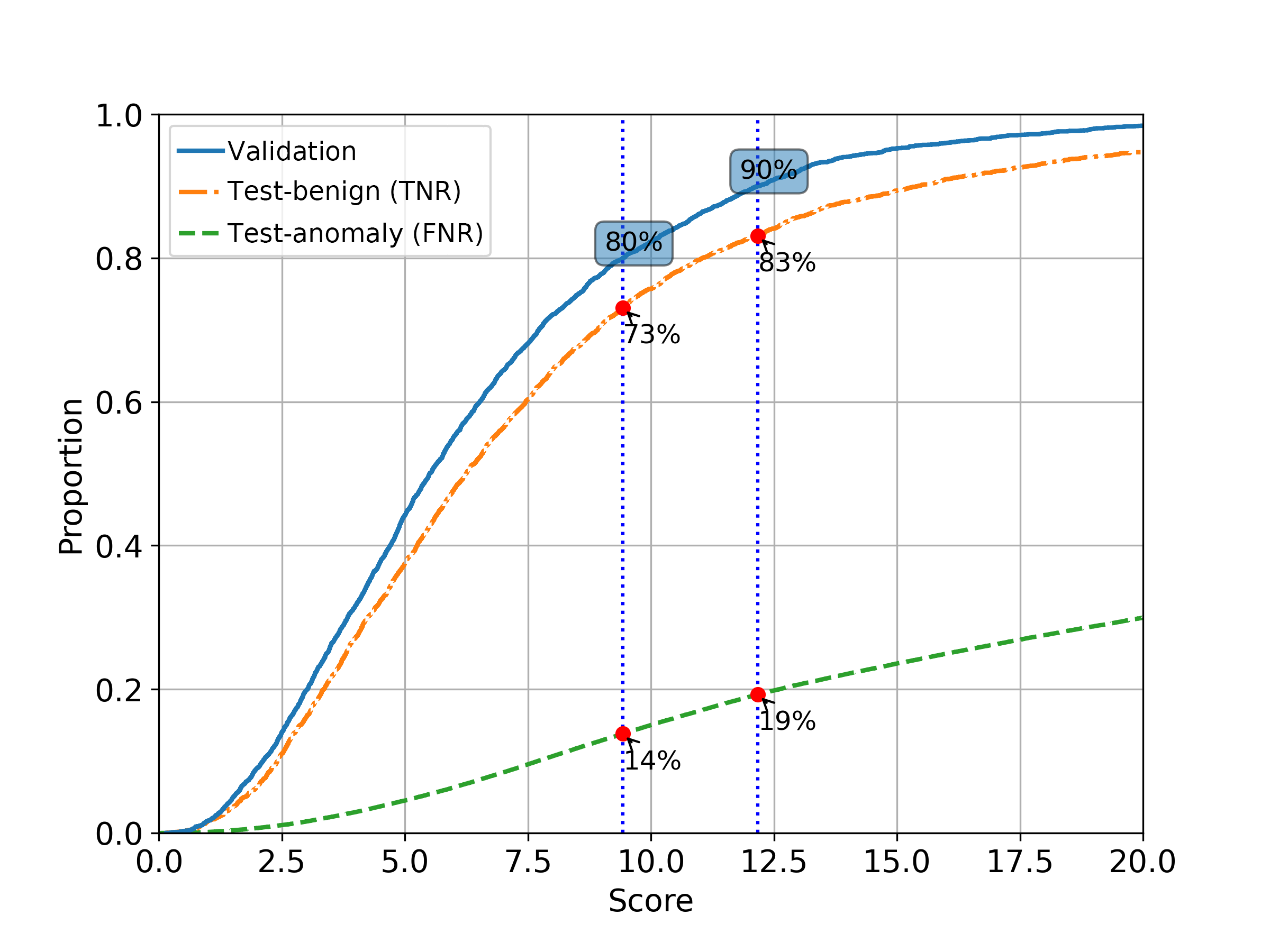} }
    \caption{The impact on TNR and FNR of the test set when choosing different threshold ($\alpha_r$ with $r=80$th percentile and $90$th percentile) of scores on validation set (TNR: percentage shown on the dashed-dotted orange curve; FNR: percentage shown on the dashed green curve) of three methods. From left to right are: naive-norm, 2DSig-Conf (Level1), and 2DSig-Norm (Level1).}
    \label{fig:l1_con_cov}  
\end{figure}

\begin{figure}[!t]
     \centering
        \includegraphics[width=2.5in]{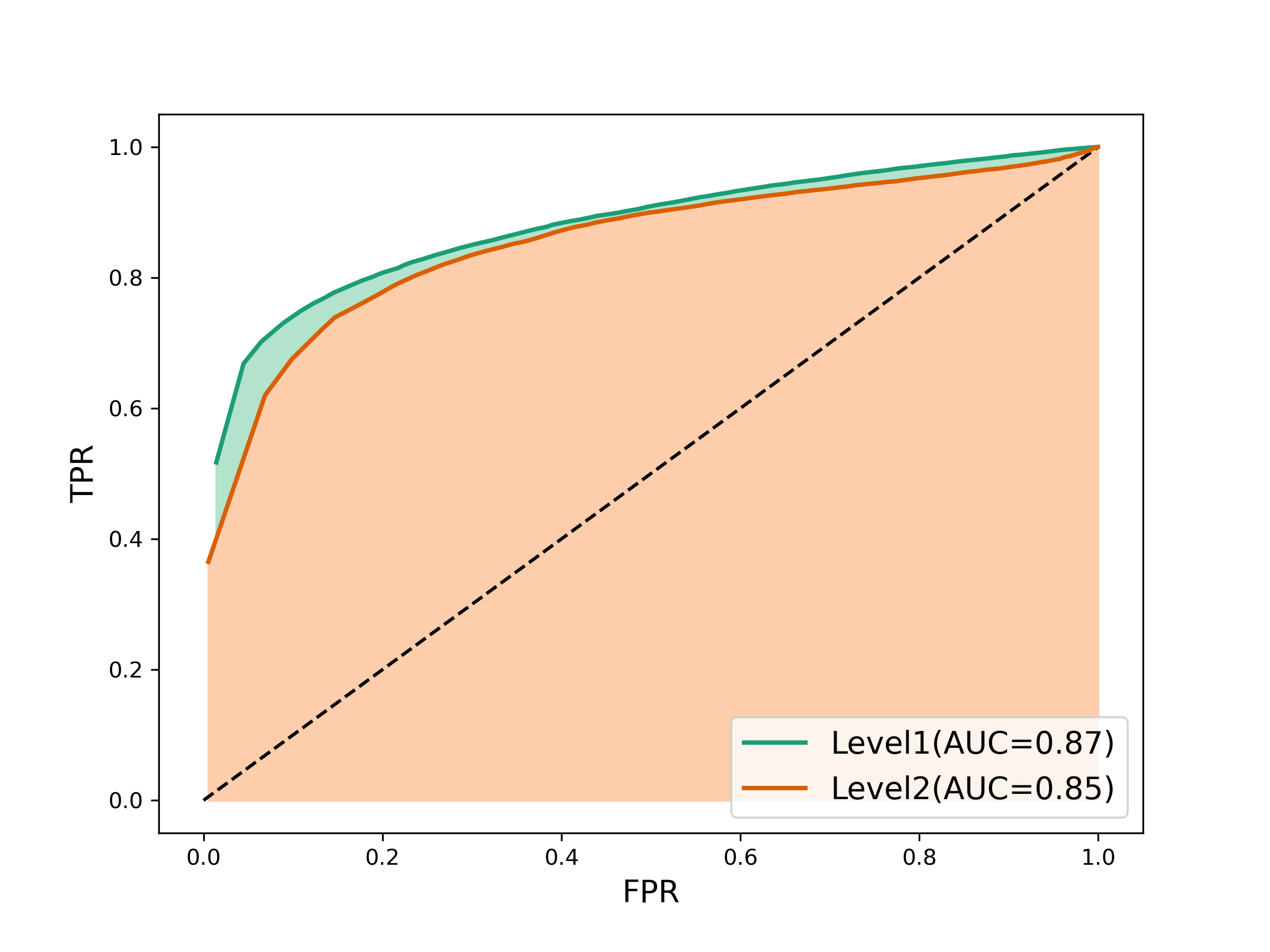} \qquad
     \includegraphics[width=2.5in]{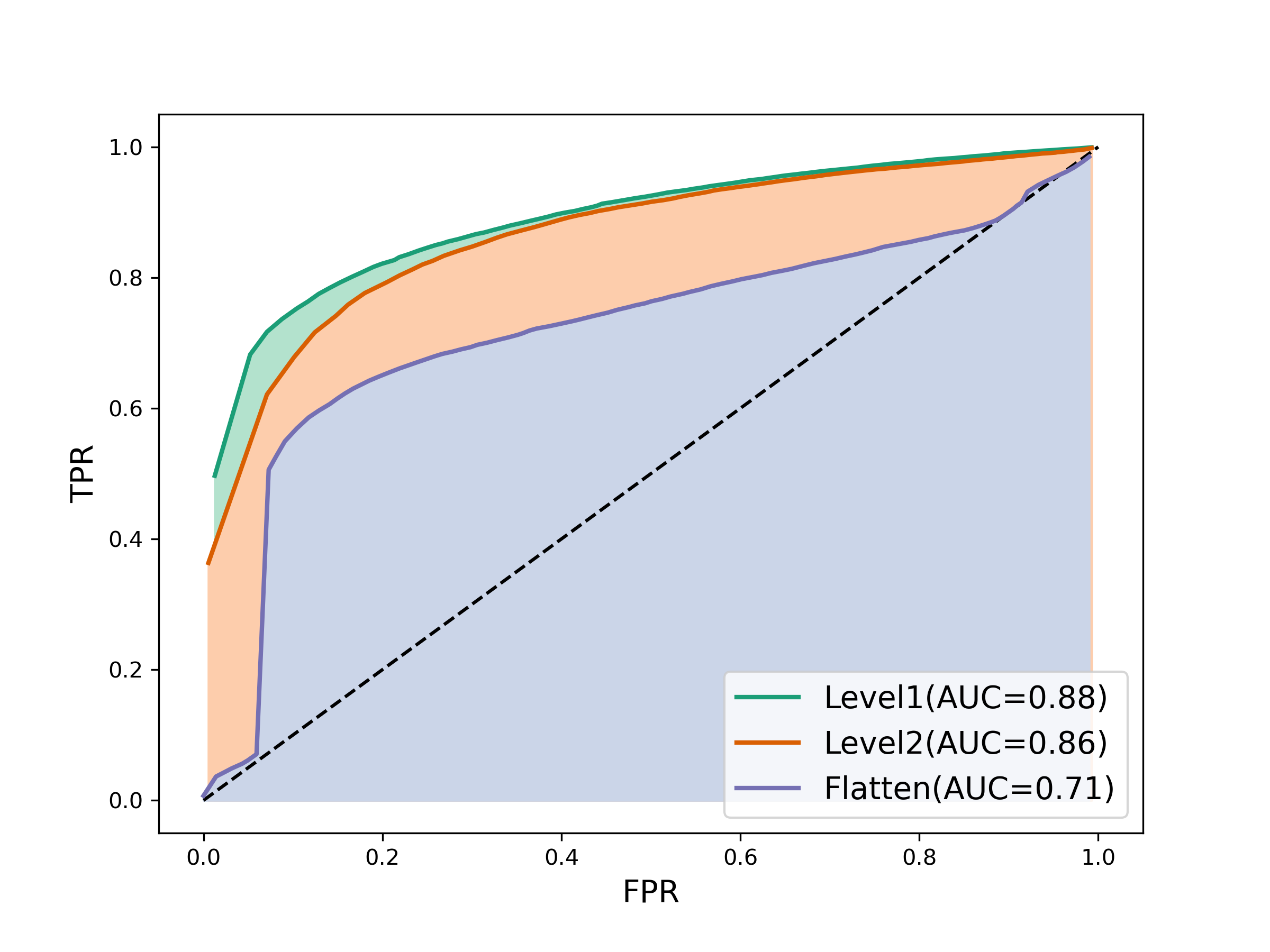}\\
    \caption{The ROC curves of three methods. Left: 2DSig-Conf on level 1 and level 2 2D-signature features; Right: 2DSig-Norm on level 1 and level 2 2D-signature features and naive-norm with the flatten representation feature.}
    \label{fig:roc_con_cov}  
\end{figure}

The ROC curves for Experiment 1 is shown in Fig. \ref{fig:l1_con_cov}, demonstrating how the true negative rate (TNR) and false negative rate (FNR) change with different thresholds $\alpha_r$, $r=80$th or $90$th percentile. The naive-norm gives the worst FNR while achieving the highest TNR. Note that for the curves generated using the naive norm, there is a significant percentage of outliers (approximately ten percent) with scores of zero, leading to a non-zero lower bound for TNR. This indicates poor performance of the naive norm: regardless of how low the threshold is set, at least ten percent of outliers remain undetected.  Compared to 2DSig-Conf, 2DSig-Norm achieves both higher TNR and lower FNR consistently. Fig. \ref{fig:roc_con_cov} gives corresponding ROC curves of Table \ref{tab:size32_comparision}. Note that the sharp jump of the ROC curve of naive-norm is due to the non-zero lower bound for TNR (and therefore FPR).

Compared to the the flatten representation, 
the models with 2D-signature features give better performance while significantly reducing the training time without loss of performance. We will focus more on comparing between 2DSig-Norm and 2DSig-Conf in the remaining part. 

For $m=64$, the results are recorded in Table \ref{tab:roc_time-64}. 

\begin{table}[]
\centering
   \caption{Comparison of the performance between 2DSig-Norm and 2DSig-Conf on the CUReT dataset with sampled image dimensions of $(64, 64, 3)$. The metrics used for comparison are the mean$\pm $std of AUC and the time cost in seconds.}
    \label{tab:roc_time-64}
    \begin{footnotesize}
    \begin{tabular}{cc|c|c}
\hline
\hline
\multicolumn{2}{l|}{}                                        & AUC            & time  (s)            \\ \hline
\multicolumn{1}{l|}{\multirow{2}{*}{\bf 2DSig-Norm}} & Level1 &  {\color{red}$0.91\pm 0.01$}  & {\color{red}$0.65\pm 0.07$}     \\ \cline{2-4} 
\multicolumn{1}{l|}{}                            & Level2 & $0.89\pm 0.01$ & $0.75\pm 0.03$    \\ \hline
\multicolumn{1}{l|}{\multirow{2}{*}{2DSig-Conf}} & Level1 &  $0.91\pm 0.02$  & $4.11\pm 0.07$     \\ \cline{2-4} 
\multicolumn{1}{l|}{}                            & Level2 &  $0.88\pm 0.01$ & $9.56\pm 0.03$     \\ \hline
\hline
\end{tabular}
\end{footnotesize}
\end{table}

\hspace{0pt}\\\textbf{Experiment 2.}  We compared two frameworks with Gaussian mixture model (GMM) \cite{reynolds2009gaussian} on level 1 of 2D-signature. GMM is chosen as a strong baseline for anomaly detection due to its ability to model complex data distributions probabilistically and operate in an unsupervised manner, providing a flexible and adaptable approach to identifying anomalies \cite{reynolds2009gaussian}. The mean accuracy, TPR and F1-score (with standard deviations) are reported in Table \ref{tab:gmm_comparision}. The 2DSig-Norm consistently outperforms the other two models on all three metrics. The TPR of 2DSig-Norm is twice that of the 2DSig-GMM.

\begin{table}[]
\centering
   \caption{Performance comparison among 2DSig-GMM, 2DSig-Conf and 2DSig-Norm on CUReT dataset with sampled image dimensions of $(32, 32, 3)$. The metrics used for comparison are the mean$\pm $std of accuracy, TPR and F1-score.}
    \label{tab:gmm_comparision}
    \begin{footnotesize}
\begin{tabular}{c|c|c|c}
\hline
\hline
           & Acc        & TPR              & F1-score         \\
           \hline
2DSig-GMM  & $0.675\pm 0.100$ & $0.458\pm 0.238$ & $0.637\pm 0.126$ \\
2DSig-Conf & $0.782\pm 0.120$ & $0.978\pm 0.027$ & $0.790\pm 0.138$ \\
{\bf 2DSig-Norm} & {\color{red}$0.794\pm 0.128$} & {\color{red}$0.983\pm 0.027$} & {\color{red}$0.803\pm 0.135$}\\
\hline
\hline
\end{tabular}
\end{footnotesize}
\end{table}

\begin{figure}[!t]
     \centering
     \includegraphics[width=3in]{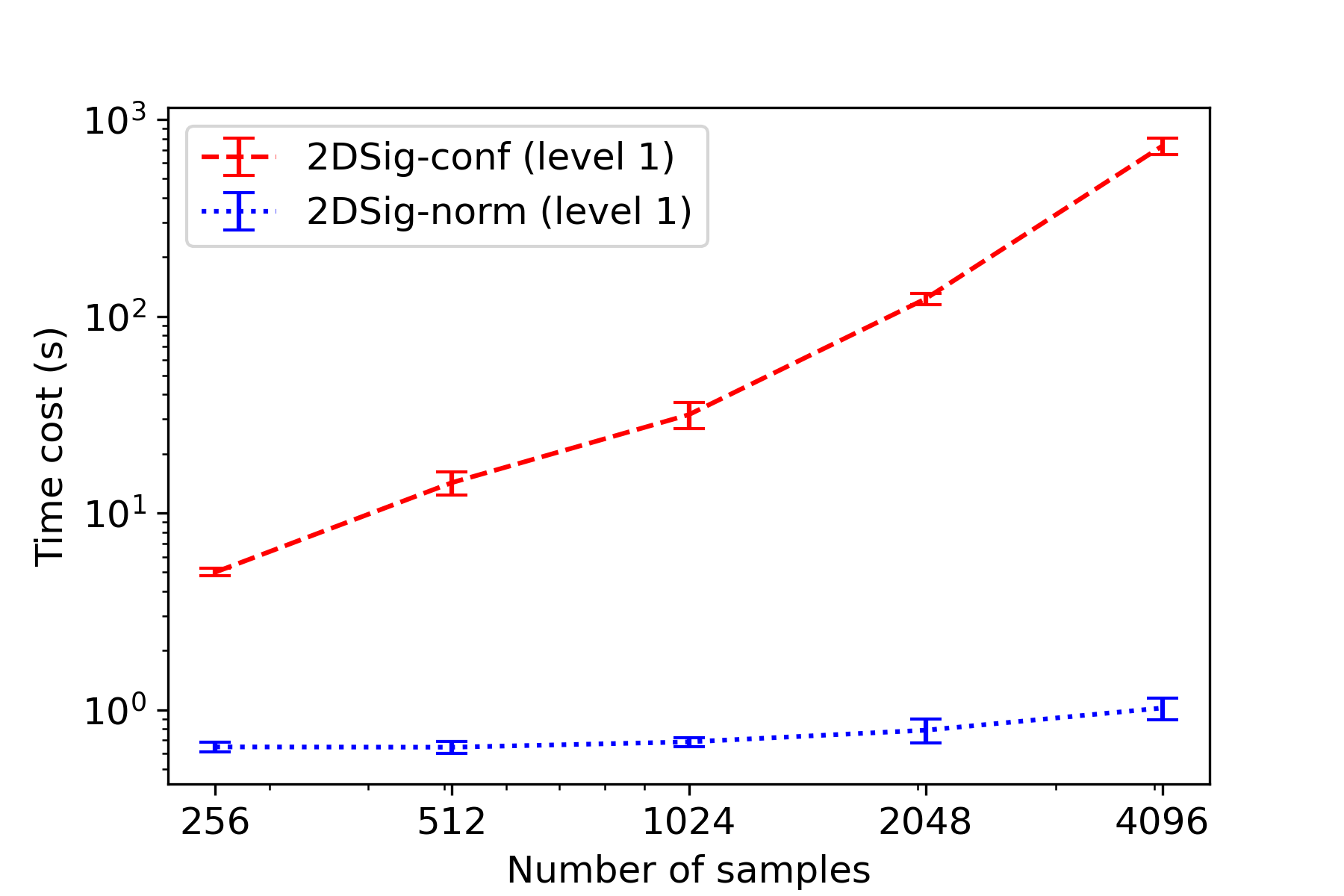}
          \includegraphics[width=3in]{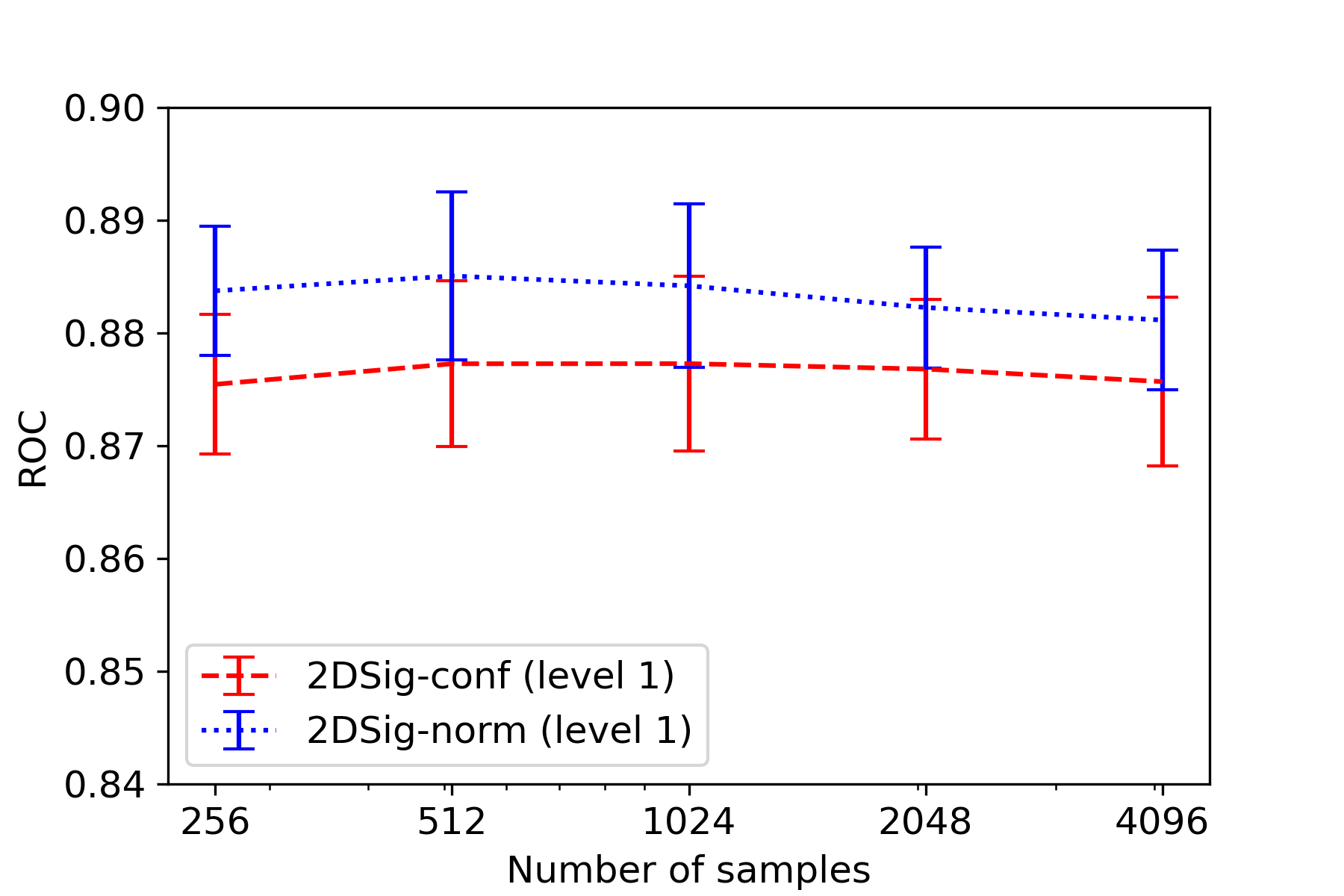}\\
    \caption{Time cost and AUC (mean$\pm $std) against number of samples in training set for  2DSig-Conf and 2DSig-Norm on the level 1 feature.}
    \label{fig:time_n}  
\end{figure}

\hspace{0pt}\\\textbf{Complexity} To examine how the performance changes with the number of samples in the training set, we implement the first experiment via the 2DSig-Norm and the 2DSig-Conf (level 1 feature) on $2^j$ training samples, $j=8, \ldots, 12$ and plot the time cost as well as AUC (with standard deviation) against number of samples in Fig. \ref{fig:time_n}. The time cost of the 2DSig-Conf and the 2DSig-Norm increase as polynomial and sub-linearly with respect to the number of samples respectively, while AUC stays at the same level for both. This again supports our choice: the 2DSig-Norm provides a cost-saving solution compared to the 2DSig-Conf.

\subsection{Testset-level defense} \label{ssec:testlevel}
The following experiments focus on detecting the presence of adversarial perturbations in images at test time. We aim to design a fast and accurate anomaly detector, using both 2DSig-Norm and 2DSig-Conf of the \emph{\acrshort{tag_name}}, to defend the trained model against both untargeted and targeted evasion attacks. The perturbed images are crafted using the iterative fast gradient sign method (IFGSM) \cite{kurakin2017adversarial} and the projected gradient descent method (PGDM) \cite{madry2018towards}, two common evasion attack algorithms. The threat model considered assumes the adversary's knowledge of the trained model and capability to manipulate data input at inference time, with the goal of producing an integrity violation \cite{Biggio2018}. We consider the image classification task on the CIFAR-10 dataset, using the pre-trained models RepVGG-A2 and ResNet-20, both achieving $92$\%+ classification accuracy. Table \ref{tab:sar} shows that both IFGSM and PGDM give significantly high untargeted successful attack rate (SAR) and reasonable targeted SAR on ResVGG-A2\footnote{The results are similar for ResNet-20 so we do not report them}.

\subsubsection{Evasion attacks}
Evasion attacks involve manipulating input data at test/inference time to deceive machine learning models through integrity violations. Adversarial examples are commonly used in these attacks, using methods such as IFGSM and PGSM, which are both rooted in gradient-based optimization techniques \cite{akhtar2018threat}.

\hspace{0pt}\\\textbf{IFGSM} is an attack method that perturbs input data by iteratively adding noise in the direction of the gradient of the loss function with respect to the input, scaled by a small step size. This small perturbation can significantly alter the model's output, causing it to misclassify the input. Starting from the original image $\mathbf{x}_0 = \mathbf{x}$, the image is updated in the direction of the gradient of the loss function $\mathcal{L}(\theta, \mathbf{x}, l)$ with respect to the input image $\mathbf{x}$ over multiple iterations $N$:
\begin{footnotesize}
\[
\mathbf{x}_{i+1} = \mathbf{x}_i + h \cdot \text{sign}(\nabla_{\mathbf{x}_i} \mathcal{L}(\theta, \mathbf{x}_i, l)), i\in [N],
\]
\end{footnotesize}
where $h$ is a small step size, $\theta$ represents the model parameters, $l$ is the true label of the image, and $\nabla_{\mathbf{x}_i} \mathcal{L}(\theta, \mathbf{x}_i, l)$ is the gradient of the loss function with respect to the input image $\mathbf{x}_i$ at iteration $i$. The noise level is defined as $\varepsilon:=hN$, the maximum perturbation that can be achieved.

\begin{figure}[ht]
    \centering
    % First row of images
    \subfloat[Airplane - IFGSM]{
       % \begin{subfigure}[b]{\linewidth}
            \centering
            \includegraphics[width=0.9\textwidth]{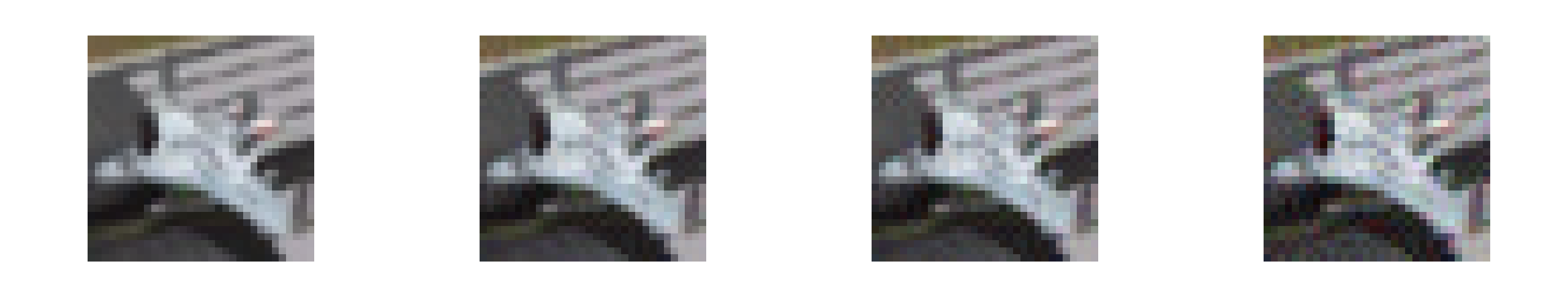}
                % \caption{Airplane - IFGSM}
            %\hspace{0.05\textwidth}
     } 
   % \end{subfigure}
   \\
    % \hspace{0.05\textwidth}
     \subfloat[Cat - PGDM]{
           % \begin{subfigure}[b]{\linewidth}
            \centering
            \includegraphics[width=0.9\textwidth]{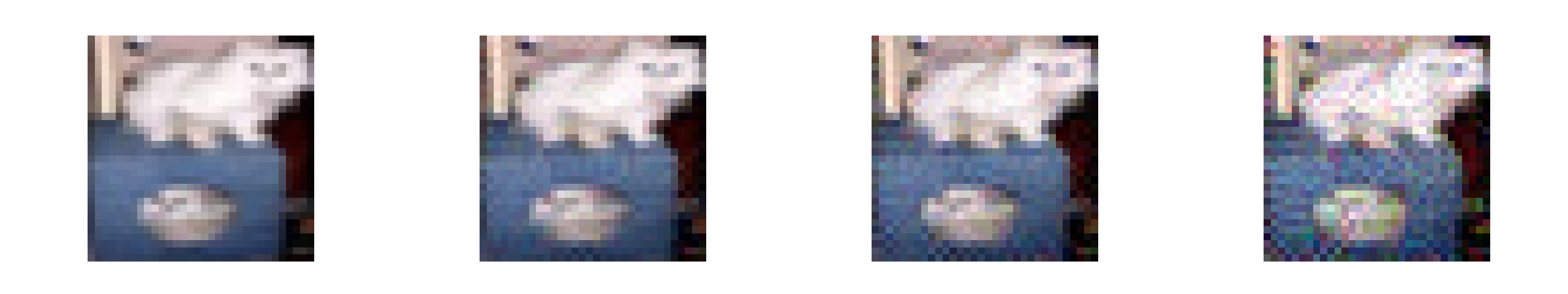}
                % \caption{Automobile - PGDM}
            % \hspace{0.05\textwidth}
     } 
   % \end{subfigure}
    \caption{Examples of adversarial images generated by IFGSM and PGDM with different noise levels. From left to right: original image, noise levels 0.03, 0.05, and 0.10.}
    \label{fig:pgd_examples}
\end{figure}

\hspace{0pt}\\\textbf{PGDM} is an attack method that iteratively perturbs input data similar to IFGSM, while ensuring that the perturbed input remains within a specified boundary. This boundary, often defined by an \(\varepsilon\)-ball, ensures that the perturbation does not deviate too much from the original input, maintaining the integrity of the original data. Starting from the original image \(\mathbf{x}_0 = \mathbf{x}\), the image is updated in the direction of the gradient of the loss function \(\mathcal{L}(\theta, \mathbf{x}, l)\) with respect to the input image \(\mathbf{x}\) over multiple iterations \(N\):
\begin{footnotesize}
\[
\mathbf{x}_{i+1} = \text{clip}_{\mathbf{x}, \varepsilon}(\mathbf{x}_i + h \cdot \text{sign}(\nabla_{\mathbf{x}_i} \mathcal{L}(\theta, \mathbf{x}_i, l))), \quad i \in [N],
\]
\end{footnotesize}
where the function \(\text{clip}_{\mathbf{x}, \varepsilon}(\cdot)\) projects the updated image back onto the \(\varepsilon\)-ball around the original image \(\mathbf{x}\), ensuring that the perturbation remains within the maximum allowed perturbation \(\varepsilon\). Examples of adversarial images generated using different noise levels (0.03, 0.05, 0.10) for both IFGSM and PGD methods are shown in Figure \ref{fig:pgd_examples}. These images were generated using the Adversarial Robustness Toolbox (ART) \cite{art2023}.

\hspace{0pt}\\\textbf{Untargeted and targeted evasions.} Untargeted evasions aim to cause a machine learning model to misclassify an input without specifying a particular incorrect class. The goal is simply to make the model's prediction incorrect, without regard to what the incorrect prediction is. Targeted evasions, on the other hand, aim to manipulate the input so that the model incorrectly classifies it as a specific, desired class. In this scenario, the attacker has a particular target label in mind and modifies the input to ensure the model predicts this target label \cite{Biggio2018}. 
\subsubsection{The experiment setup}
 The experiment is performed and tested ten times for both IFGSM and PGDM on CIFAR-10, each selecting 20\% of the data at random. Depending on the type of attack, images generated from untargeted and targeted attacks with noise level $\varepsilon\in \{0.02,0.03\}$ on some pretrained trained model are collected for each  of the ten classes, and named as \emph{untarget} and \emph{target}. We collect three test sets: a benign set, an untargeted set, and a targeted set. The benign test set is the original test set of CIFAR-10. Each image within the benign test set will be attacked untargetly and and be included in the untargeted test set. For simplicity, we will refer to \emph{untarget class $i$} within the untargeted test set as set consisting of images that were originally from class $i$ but have undergone an untargeted attack. The targeted test set is generated differently: for each \emph{target class $j$} within the targeted test set, benign images are randomly and evenly selected from the other nine classes, and a targeted attack is carried out to try to mislabel them as class $j$.

The trained models considered in our example are ResNet20 and RepVGG-A2 (Table \ref{tab:repvgg_a2}) pre-trained by Chen \cite{chen2024pytorch}, with the corresponding classification accuracy  92.66\% and 95.00\%. Attack rates are reported in Table \ref{tab:sar}.
The learned representation is chosen to be the output of \emph{Residual Block 1} of ResNet-20 defined in Table \ref{tab:resnet20}, with a short discussion on the choice later. Indeed, the appropriate learned representation should emphasise the distributional differences between benign images and those with attacks, while minimizing the differences between various image classes. Thus we exclude later blocks and prefer earlier blocks that may try to capture the edge information, where the attack information may be misidentified as edge information. 
The 2D-signature features extracted are level 1 and level 2. As the length of fully flattened representation is long, that is $16,384$ from output size $(32,32,16)$, we choose to take the channel-mean as the flattened representation for naive-norm for comparison. 

For each class, a threshold is generated on the corresponding validation set, this determines whether the incoming instance is an anomaly for that class or not, as shown in Fig. \ref{fig:flowchart}. In total there are therefore ten thresholds, one for each class. We will mark an instance for deletion only if it is identified as anomaly for all ten classes. One incoming instance will be sorted as benign if there is one class of the ten such that it is identified as benign for that class. 

Note that the goal of this experiment is anomaly detection, not classification. This restrict criteria may in general require a much lower percentile $r$ as the threshold than $r=$80th or 90th percentile as in Section \ref{ssec: classification}. We choose in this experiment $r=20$th percentile for both frameworks, regardless of the features added on.

\begin{table}[]
\centering
   \caption{The successful attack rate (SAR) for various attacks generated from RepVGG-A2 at noise levels \(\varepsilon \in \{0.02, 0.03\}\) on both RepVGG-A2 (self) and ResNet-20 (transfer attack).}
    \label{tab:sar}
    \begin{footnotesize}
\begin{tabular}{cc|cc|cc}
\hline
\hline
\multicolumn{2}{c|}{\multirow{2}{*}{}}                           & \multicolumn{2}{c|}{RepVGG-A2 (self)}          & \multicolumn{2}{c}{ResNet-20 (transfer attack)}                                       \\ \cline{3-6} 
\multicolumn{2}{c|}{}                                            & \multicolumn{1}{c|}{Untarget} & Target  & \multicolumn{1}{c|}{Untarget } & Target  \\ \hline\hline
\multicolumn{1}{c|}{\multirow{2}{*}{IFGSM}} & $\varepsilon=0.03$ & \multicolumn{1}{c|}{94.20\%}  & 66.40\% & \multicolumn{1}{c|}{64.64\%}                 & 25.92\%               \\ \cline{2-6} 
\multicolumn{1}{c|}{}                       & $\varepsilon=0.02$ & \multicolumn{1}{c|}{87.75\%}  & 53.92\% & \multicolumn{1}{c|}{58.72\%}                 & 25.92\%               \\ \hline
\multicolumn{1}{c|}{\multirow{2}{*}{PGDM}}  & $\varepsilon=0.03$ & \multicolumn{1}{c|}{96.00\%}  & 77.04\% & \multicolumn{1}{c|}{76.00\%}                 & 36.00\%               \\ \cline{2-6} 
\multicolumn{1}{c|}{}                       & $\varepsilon=0.02$ & \multicolumn{1}{c|}{94.40\%}  & 67.32\% & \multicolumn{1}{c|}{65.08\%}                 & 26.96\%               \\ \hline \hline
\end{tabular}
\end{footnotesize}
\end{table}

\hspace{0pt}\\\textbf{Metric.} We compare 2DSig-Norm against 2DSig-Conf in terms of their time cost and AUC, on level 1 and level 2 features respectively. The naive-norm is implemented as another benchmark. For performance stability, we perform the experiment ten times using different random seeds and record the mean performance together with the corresponding standard deviation. 
\subsubsection{Results}

\hspace{0pt}\\\textbf{IFGSM}
Fig. \ref{fig:repvgga2_resnet20_auc_ifgsm_untarget} 
and Fig.\ref{fig:repvgga2_resnet20_auc_ifgsm_target} in \ref{sec:app_plots} present the model performance to defense against both untargeted and targeted IFGSM attack of RepVgg-A2 at different noise level. Fig. \ref{fig:resnet20_resnet20_auc_ifgsm_untarget} 
and Fig.\ref{fig:resnet20_resnet20_auc_ifgsm_target} in \ref{sec:app_plots} present the model performance to defense both untargeted and targeted IFGSM attack of ResNet-20 at different noise level. Both 2DSig-Norm and 2DSig-Conf (with the same choice of 2D-signature features) give similar AUCs, much higher than naive-norm. For the same model, the performance to defense attack at noise $\varepsilon=0.03$ is higher than the one to defense attack at noise $\varepsilon=0.02$, because the noise signal is stronger at the higher noise level.  Table \ref{tab:ifgsm_time_003} reports the time costs for different models to detect IFGSM attacks in 3,000 images. Only the time cost at noise level $\varepsilon=0.03$ is recorded as the cost difference is subtle between different noise levels.  2DSig-Norm achieves the same performance as 2DSig-Conf at only 1/200 of the time cost.
As the performance of 2DSig-Norm and 2DSig-Conf are similar, we report the F1-score and TPR for 2DSig-Norm (level 2) only in Table \ref{tab:ifgsm_f1}.

\begin{table}[]
\centering
\caption{The time cost (s) for different models to defend against IFGSM attack with noise level $\varepsilon=0.03$ per 3,000 images.}
\label{tab:ifgsm_time_003}
\begin{footnotesize}
\begin{tabular}{cc|c|c}
\hline
\hline
\multicolumn{2}{c|}{}                                          & RepVGG-A2         & ResNet-20         \\ \hline
 \hline
\multicolumn{1}{c|}{\multirow{2}{*}{\bf 2DSig-Norm}} & Level1      & {\color{red}0.0056$\pm$0.0004} & {\color{red}0.0065$\pm$0.0004} \\ \cline{2-4} 
\multicolumn{1}{c|}{}                            & Level2      & 0.0068$\pm$0.0010 & 0.0103$\pm$0.0031 \\ \hline
\multicolumn{1}{c|}{\multirow{2}{*}{2DSig-Conf}} & Level1      & 0.9872$\pm$0.0566 & 0.9394$\pm$0.0523 \\ \cline{2-4} 
\multicolumn{1}{c|}{}                            & Level2      & 1.1059$\pm$0.0539 & 1.0460$\pm$0.0426 \\ \hline
\hline
\end{tabular}
\end{footnotesize}
\end{table}

\begin{table}[]
\centering
\caption{The F1-score and TPR for 2DSig-Norm to defend against IFGSM attack with different noise levels $\varepsilon$.}
\label{tab:ifgsm_f1}
\begin{footnotesize}
\begin{tabular}{cc|cc|cc}
\hline
\hline
\multicolumn{2}{c|}{\multirow{2}{*}{}}                              & \multicolumn{2}{c|}{RepVGG-A2}                         & \multicolumn{2}{c}{ResNet-20}                         \\ \cline{3-6} 
\multicolumn{2}{c|}{}                                               & \multicolumn{1}{c|}{Untargeted}        & Targeted          & \multicolumn{1}{c|}{Untargeted}        & Targeted          \\ \hline\hline
\multicolumn{1}{c|}{\multirow{2}{*}{F1-score}} & $\varepsilon=0.03$ & \multicolumn{1}{c|}{0.86 $\pm$ 0.01} & 0.85 $\pm$ 0.01 & \multicolumn{1}{c|}{0.84 $\pm$ 0.01} & 0.85 $\pm$ 0.01 \\ \cline{2-6} 
\multicolumn{1}{c|}{}                          & $\varepsilon=0.02$ & \multicolumn{1}{c|}{0.79 $\pm$ 0.01} & 0.77 $\pm$ 0.01 & \multicolumn{1}{c|}{0.77 $\pm$ 0.02} & 0.76 $\pm$ 0.02 \\ \hline
\multicolumn{1}{c|}{\multirow{2}{*}{TPR}}      & $\varepsilon=0.03$ & \multicolumn{1}{c|}{0.94 $\pm$ 0.01} & 0.94 $\pm$ 0.02 & \multicolumn{1}{c|}{0.92 $\pm$ 0.02} & 0.94 $\pm$ 0.01 \\ \cline{2-6} 
\multicolumn{1}{c|}{}                          & $\varepsilon=0.02$ & \multicolumn{1}{c|}{0.83 $\pm$ 0.01} & 0.79 $\pm$ 0.01 & \multicolumn{1}{c|}{0.78 $\pm$ 0.04} & 0.77 $\pm$ 0.03 \\ \hline
\hline
\end{tabular}
\end{footnotesize}
\end{table}

\hspace{0pt}\\\textbf{PGDM} Fig. \ref{fig:repvgga2_resnet20_auc_pgdm_untarget} 
and Fig. \ref{fig:repvgga2_resnet20_auc_pgdm_target} in \ref{sec:app_plots} present the model performance to defend against both untargeted and targeted PGDM attack of RepVgg-A2 at different noise level. Fig. \ref{fig:resnet20_resnet20_auc_pgdm_untarget} 
and Fig.\ref{fig:resnet20_resnet20_auc_pgdm_target} in \ref{sec:app_plots} present the model performance to defend against both untargeted and targeted IFGSM attacks on ResNet-20 at different noise levels.  The results for PGDM are consistent with the one for IFGSM, where the performance of the 2DSig-Norm and the 2DSig-Conf show significantly better (and similar) performance to detect polluted images than naive-norm. Note that their performance to defend against PGDM attacks are slightly better in terms of AUC than the defence to attacks generated by IFGSM. Table \ref{tab:pgdm_time_003} reports the time costs for different models to detect PGDM attacks (at noise level $\varepsilon=0.03$) in 3,000 images. Similar to IFGSM, the 2DSig-Norm achieves the same performance as the 2DSig-Conf using only 1/100 of the time. The F1-score and TPR for the 2DSig-Norm (level 2) to defend against PGSM attacks are presented in Table \ref{tab:pgdm_f1}. Compared with Table \ref{tab:ifgsm_f1}, all metric results are higher here.

\begin{table}[]
\centering
\caption{The time cost (s) for different models to defend against PGDM attacks with noise level $\varepsilon=0.03$ for 3,000 images.}
\label{tab:pgdm_time_003}
\begin{footnotesize}
\begin{tabular}{cc|c|c}
\hline
\hline
\multicolumn{2}{c|}{}                                          & RepVGG-A2         & ResNet-20         \\ \hline
 \hline
\multicolumn{1}{c|}{\multirow{2}{*}{\bf 2DSig-Norm}} & Level1      & {\color{red}0.0065$\pm$0.0003} & {\color{red}0.0068$\pm$0.0003} \\ \cline{2-4} 
\multicolumn{1}{c|}{}                            & Level2      & 0.0080$\pm$0.0006 & 0.0086$\pm$0.0019 \\ \hline
\multicolumn{1}{c|}{\multirow{2}{*}{2DSig-Conf}} & Level1      & 0.9919$\pm$0.0284 & 0.9394$\pm$0.0706 \\ \cline{2-4} 
\multicolumn{1}{c|}{}                            & Level2      & 1.0767$\pm$0.0448 & 1.0066$\pm$0.0645 \\ \hline
\hline
\end{tabular}
\end{footnotesize}
\end{table}

\subsubsection{The role of learned representations}\label{ssec:role_lr}
\begin{figure}[!t]
    \centering
    \begin{minipage}{0.45\textwidth}
        \centering
        \subfloat[Normalisation {(ResNet-20)}]{%
            \includegraphics[width=\linewidth, trim=120pt 1020pt 120pt 160pt, clip]{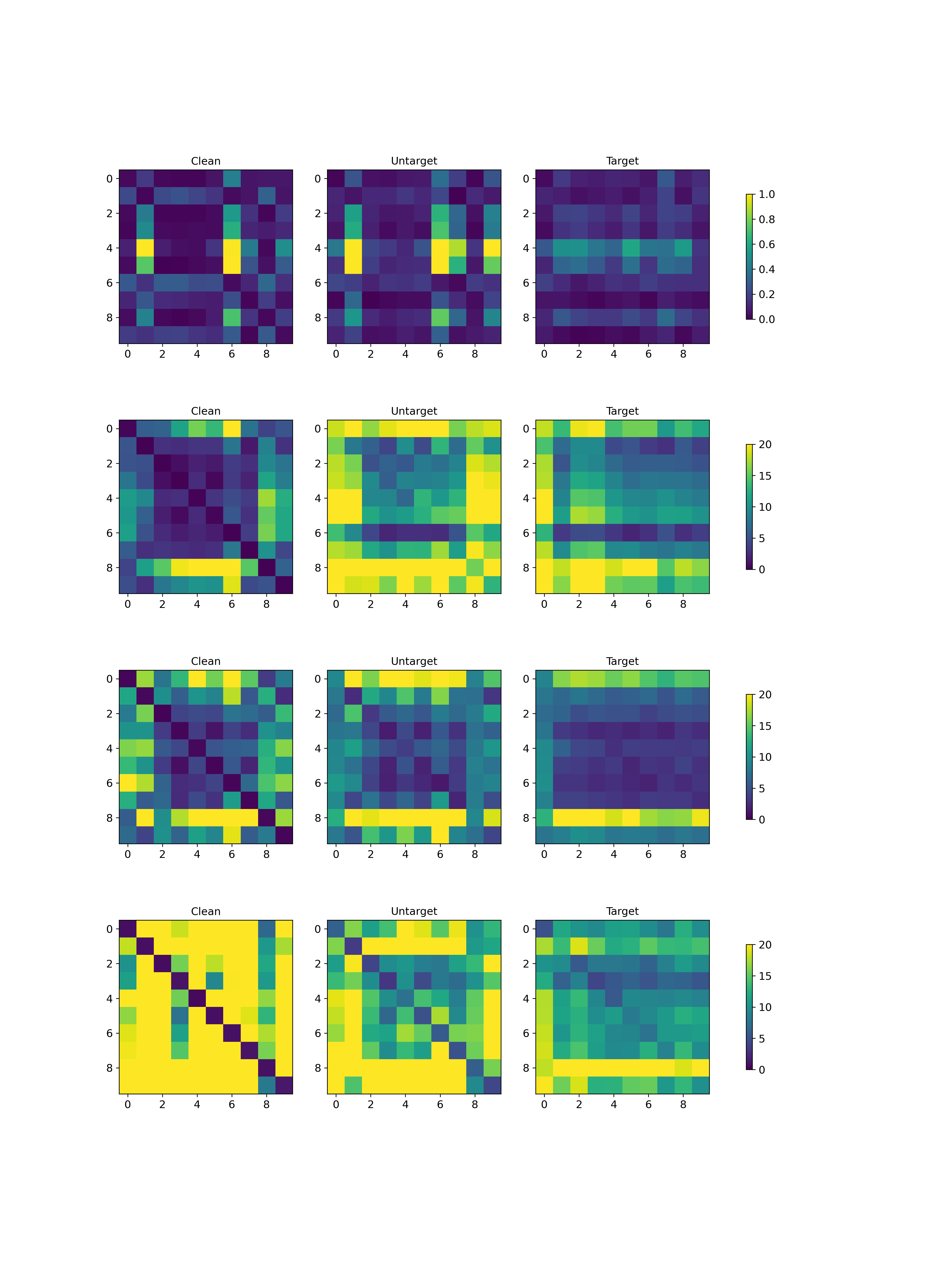}
        }\\
        \subfloat[Residual Block 1 {(ResNet-20)}]{%
            \includegraphics[width=\linewidth, trim=120pt 740pt 120pt 460pt, clip]{evasion/meandiff_ifgsm_003_10_repvgga2_resnet20_l1.png}
        }\\
        \subfloat[Residual Block 2 {(ResNet-20)}]{%
            \includegraphics[width=\linewidth, trim=120pt 460pt 120pt 740pt, clip]{evasion/meandiff_ifgsm_003_10_repvgga2_resnet20_l1.png}
        }\\
        \subfloat[Residual Block 3 {(ResNet-20)}]{%
            \includegraphics[width=\linewidth, trim=120pt 180pt 120pt 1020pt, clip]{evasion/meandiff_ifgsm_003_10_repvgga2_resnet20_l1.png}
        }
    \end{minipage}
    \hspace{0.05\textwidth}  % 控制两列之间的间距
    \begin{minipage}{0.45\textwidth}
        \centering
        \subfloat[RepVGG Block 1 { (RepVGG-A2)}]{%
            \includegraphics[width=\linewidth, trim=120pt 1020pt 120pt 160pt, clip]{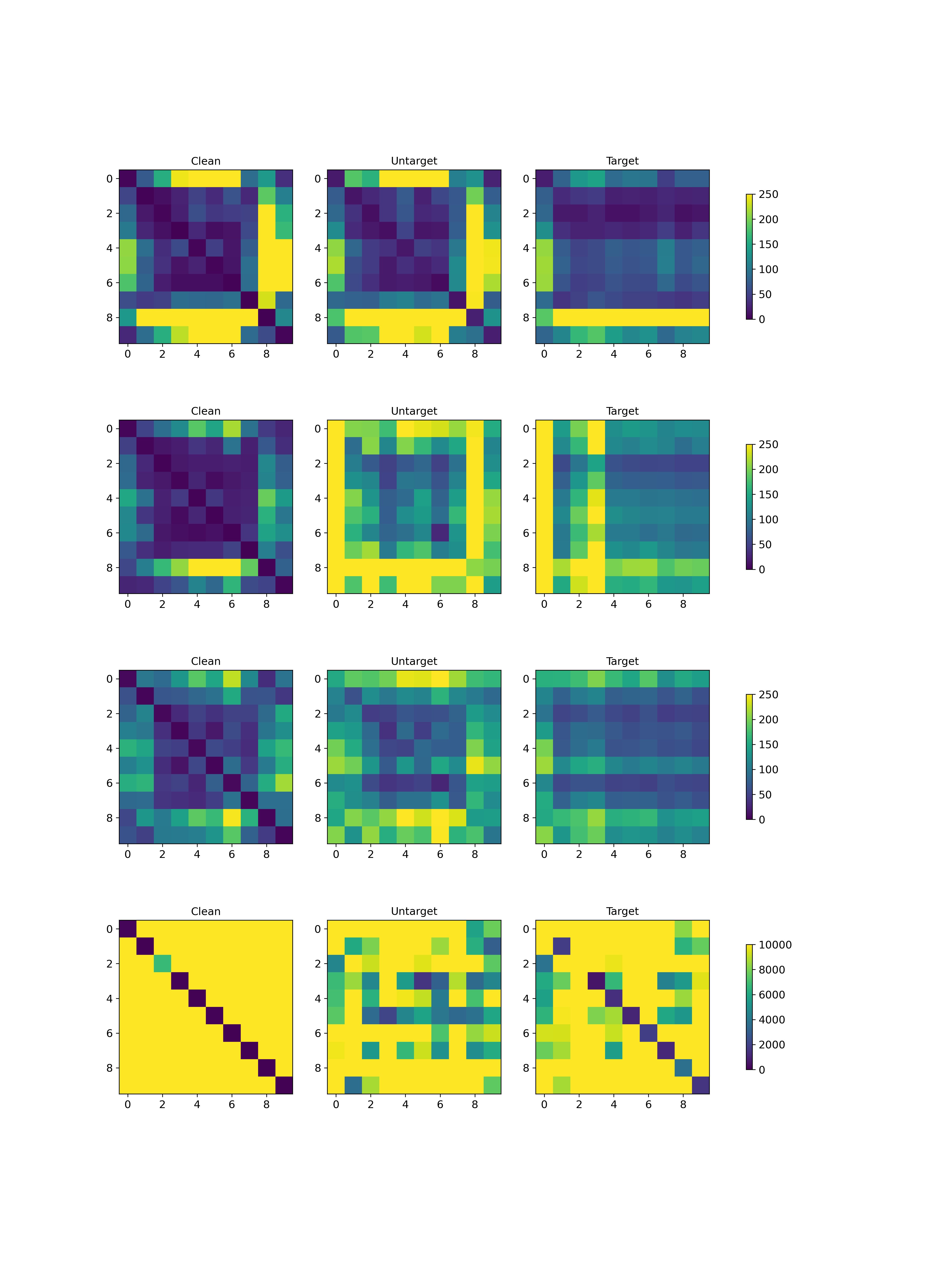}
        }\\
        \subfloat[RepVGG Block 2 { (RepVGG-A2)}]{%
            \includegraphics[width=\linewidth, trim=120pt 740pt 120pt 460pt, clip]{evasion/meandiff_ifgsm_003_10_repvgga2_repvgga2_l1.png}
        }\\
        \subfloat[RepVGG Block 3 { (RepVGG-A2)}]{%
            \includegraphics[width=\linewidth, trim=120pt 460pt 120pt 740pt, clip]{evasion/meandiff_ifgsm_003_10_repvgga2_repvgga2_l1.png}
        }\\
        \subfloat[RepVGG Block 4 { (RepVGG-A2)}]{%
            \includegraphics[width=\linewidth, trim=120pt 180pt 120pt 1020pt, clip]{evasion/meandiff_ifgsm_003_10_repvgga2_repvgga2_l1.png}                
        }
    \end{minipage}
    \caption{Mean difference heat plot between the $i$th class of the training set (the rows per heat map represent the classes of the training set) and the $j$th class of the test set (the rows per heat map represent the classes of the corresponding test set); three different test sets considered (Left: benign test images; Middle: test images with untargeted attacks; Right: test images with targeted attacks); {the left panel consists of four subplots comparing different learned representations from ResNet-20, while the right panel consists of four subplots comparing learned representations from RepVGG-A2.}}
    \label{fig:repvgga2_repvgga2} 
    \label{fig:repvgga2_resnet20} 
\end{figure}

To investigate the role of learned representations, we adopt a standard settting from literature \cite{kurakin2018adversarial} and use $10$-step IFGSM attack with noise level $\varepsilon=0.03$ on RepVGG-A2 as an example. Recall that we have a benign test set, an untarget test set and a target test set. Per test set and per learned representation, we will generate a $10\times 10$ mean difference matrix, where we examine the mean difference \eqref{eqn:meandiff} between the 2D-signature features extracted from test instances of the $i$-th class and and the features extracted from train instances, including only benign images, of the $j$-th class, $i,j\in [10]$. The heat maps of the learned representations from ResNet20 and RepVGG-A2 are presented in Fig. \ref{fig:repvgga2_resnet20}. Recall that the appropriate learned representation we seek for should emphasise the distributional differences between benign images and those with attacks (between the first and the rest matrices in the same row), while minimizing the differences between various image classes (within the same matrix). Thus we would opt for the learned representation that demonstrates slight color variation within the same matrix, while between different matrices in the same row.

{The top row of the left panel in } Fig. \ref{fig:repvgga2_resnet20} shows the mean differences when there is no learned representation extracted. The top-left figure of {the left panel in} Fig. \ref{fig:repvgga2_resnet20} shows the similarity between different classes of Cifar-10. Distributions of ten classes of Cifar10 are similar without being lifted to higher-dimensional manifold. 
The similarity between classes while detrimental to image classification performance when treated as an anomaly detection problem (as outlined in Section \ref{ssec: classification}), is precisely what we desire for identifying evasion attacks.
With only a small perturbation added, the difference between the heat map of the clean images (the top-left figure) and the one of the polluted images with the untargeted attack (the top-middle figure) is therefore subtle. Same observation applies to a comparison between the the top-left and the top-right figures.
The similarity among the top three figures in {the left panel of} Fig. \ref{fig:repvgga2_resnet20} indicates that without utilising a higher-dimensional representation of a trained model, it would be challenging to distinguish between noisy and clean (or benign) images, as reflected on visual inspection of individual samples.

The second to the last rows examine the mean differences when the learned representation are extracted from Residual Block 1 to Residual Block 3 of ResNet-20 respectively. The left column (from top to bottom) illustrates how the distribution difference between different classes are gradually reinforced (non-diagonals) when model learns to classify. The clear diagonal pattern of the bottom-left figure in {the left panel of} Fig. \ref{fig:repvgga2_resnet20} indicates that the later layer of the trained model (ResNet-20) has successfully categorized ten classes into ten distinct distributions. The middle and right columns (from top to bottom) illustrate how the distributions of the noisy images are gradually altered when the model learns to classify. The distribution difference between the benign images and the noisy images is visible at Residual Block 1 of ResNet-20 (the second row), and later at Residual Block 2-3 is invisible again when the model focuses on classification and sorts noisy images into one of the ten distributions it has been learning through high-level features. With a high (untargeted) ASR, the images being attacked untargetedly tend to be misclassified, thus it is anticipated that the distribution of the benign class $i$ will be much different from that the one of the untarget class $i$ at later layers. Similarly, with a high (targeted) ASR, the images being attacked and mislabelled as class $i$ tend to be misclassified as class $i$, thus it is anticipated that the distribution of the benign class $i$ will be similar to that of the target class $i$ at later layers. However, as the attack targets on RepVGG-A2, its attack ability is reduced when applied to a different trained model like ResNet-20 (see Table \ref{tab:sar}). This explains why we can still observe the diagonal line in the bottom middle panel, though the color differences between diagonal and non-diagonal terms are not as distinct in the bottom-left one. Similarly, one cannot observe a clear pattern within the bottom-right one as the transferable targeted SAR is low in Table \ref{tab:sar}.

\begin{figure}[!t]
     \centering
     \makebox[\textwidth][c]{
     \subfloat[Normalization only]{%
          \includegraphics[width=2in]{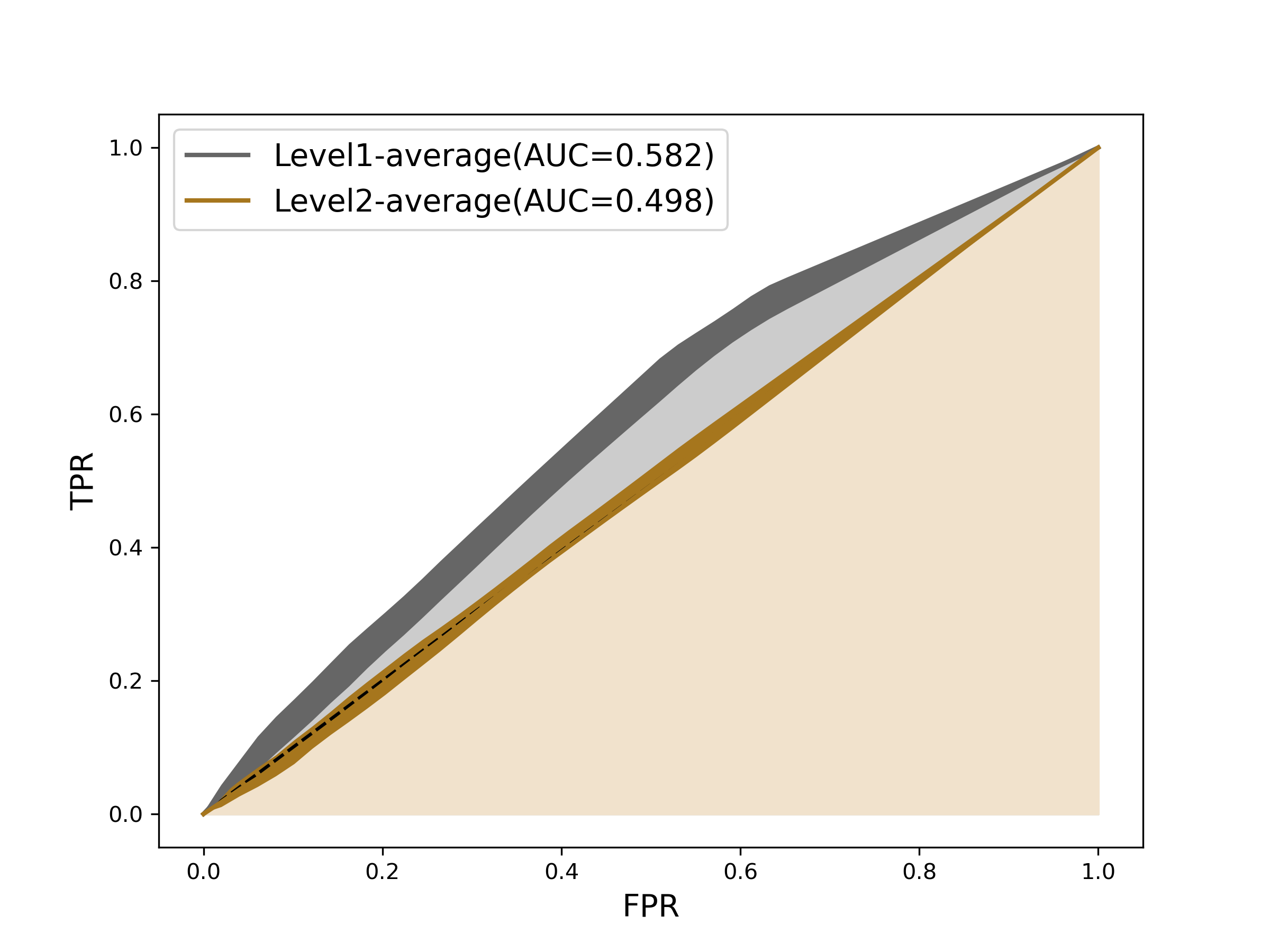}
          \hspace{-0.25in}% 
          \includegraphics[width=2in]{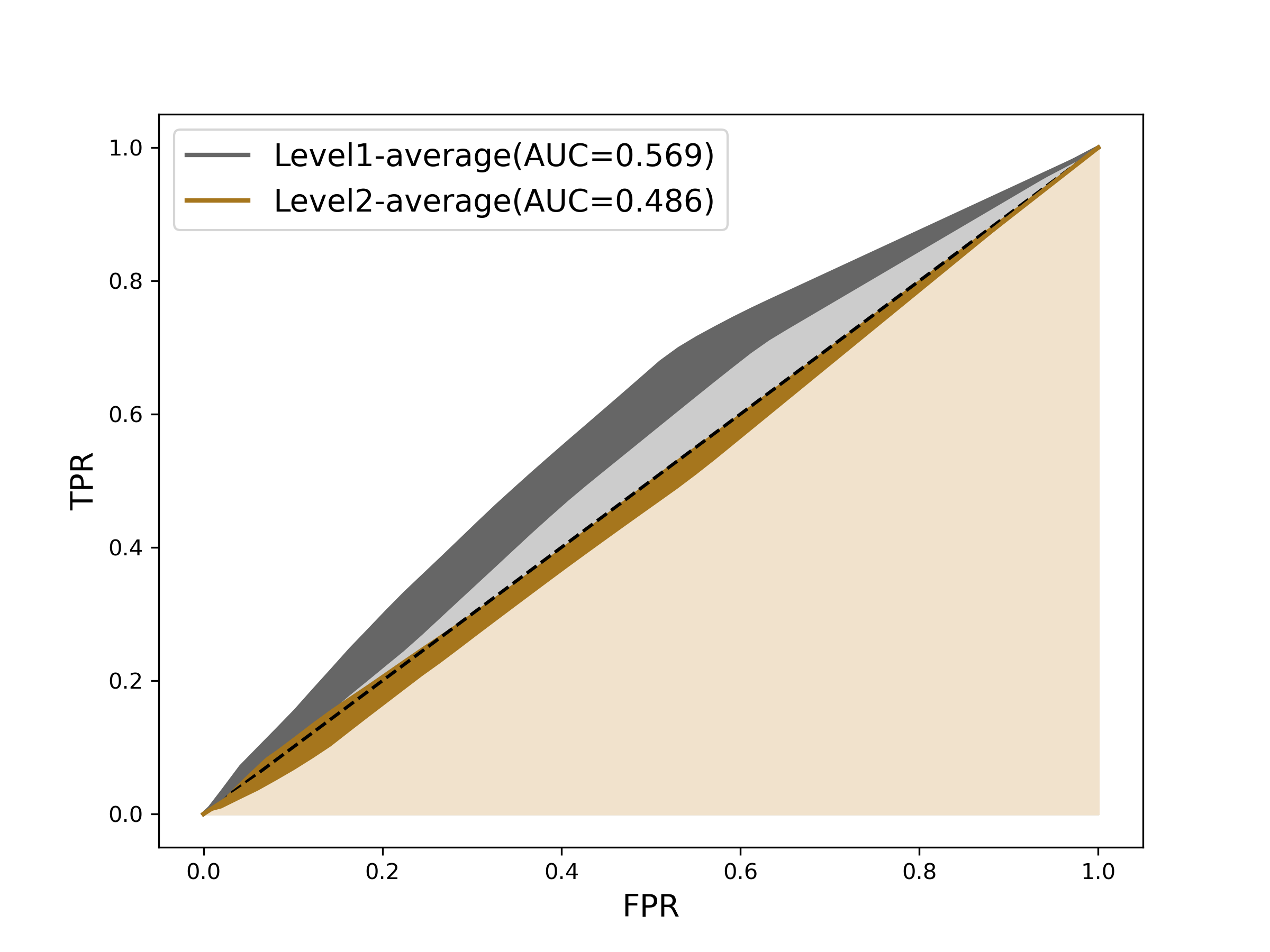} }\hspace{-0.2in}
    \subfloat[Residual Block 1]{%
          \includegraphics[width=2in]{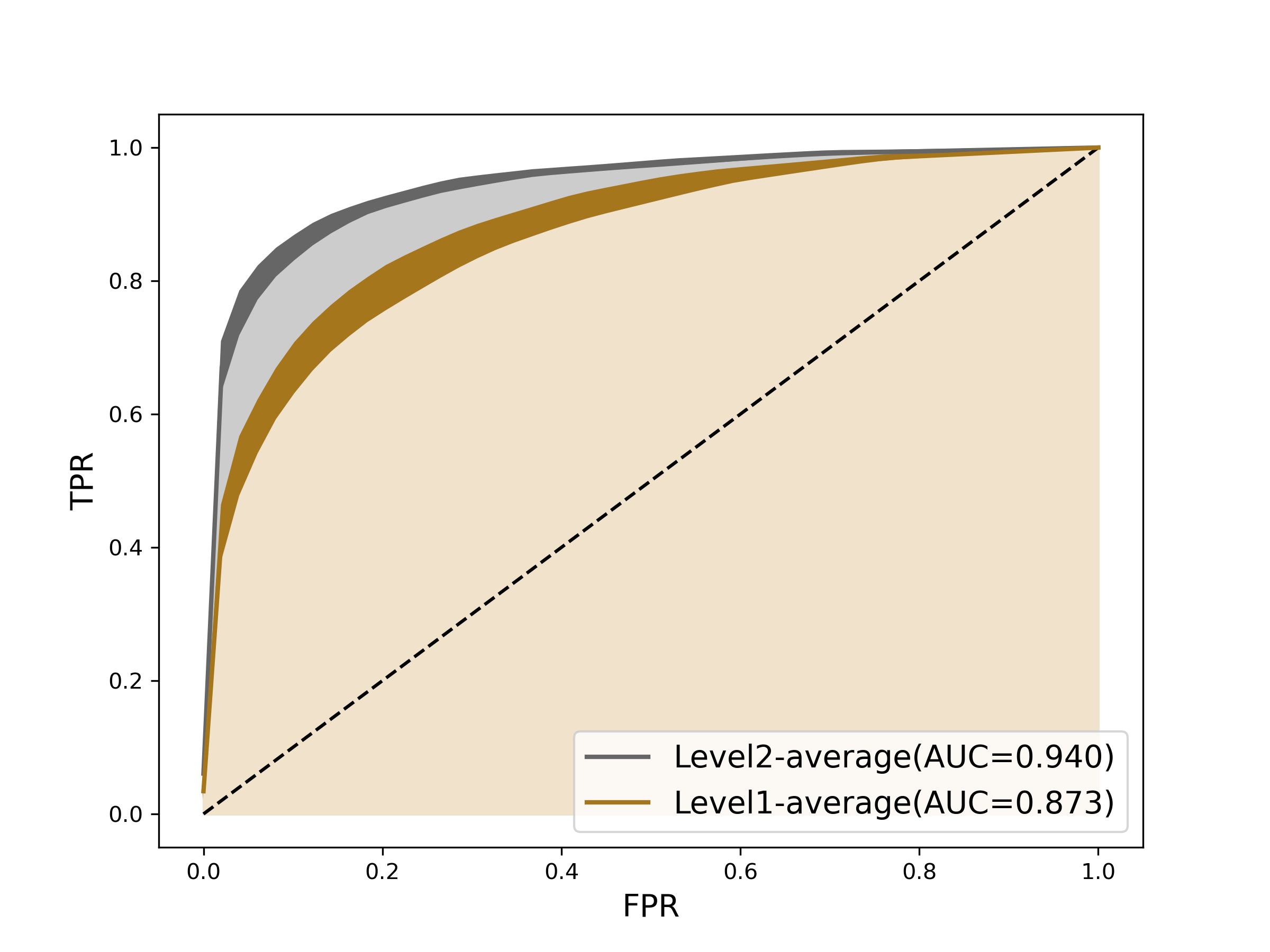} 
          \hspace{-0.25in}% 
          \includegraphics[width=2in]{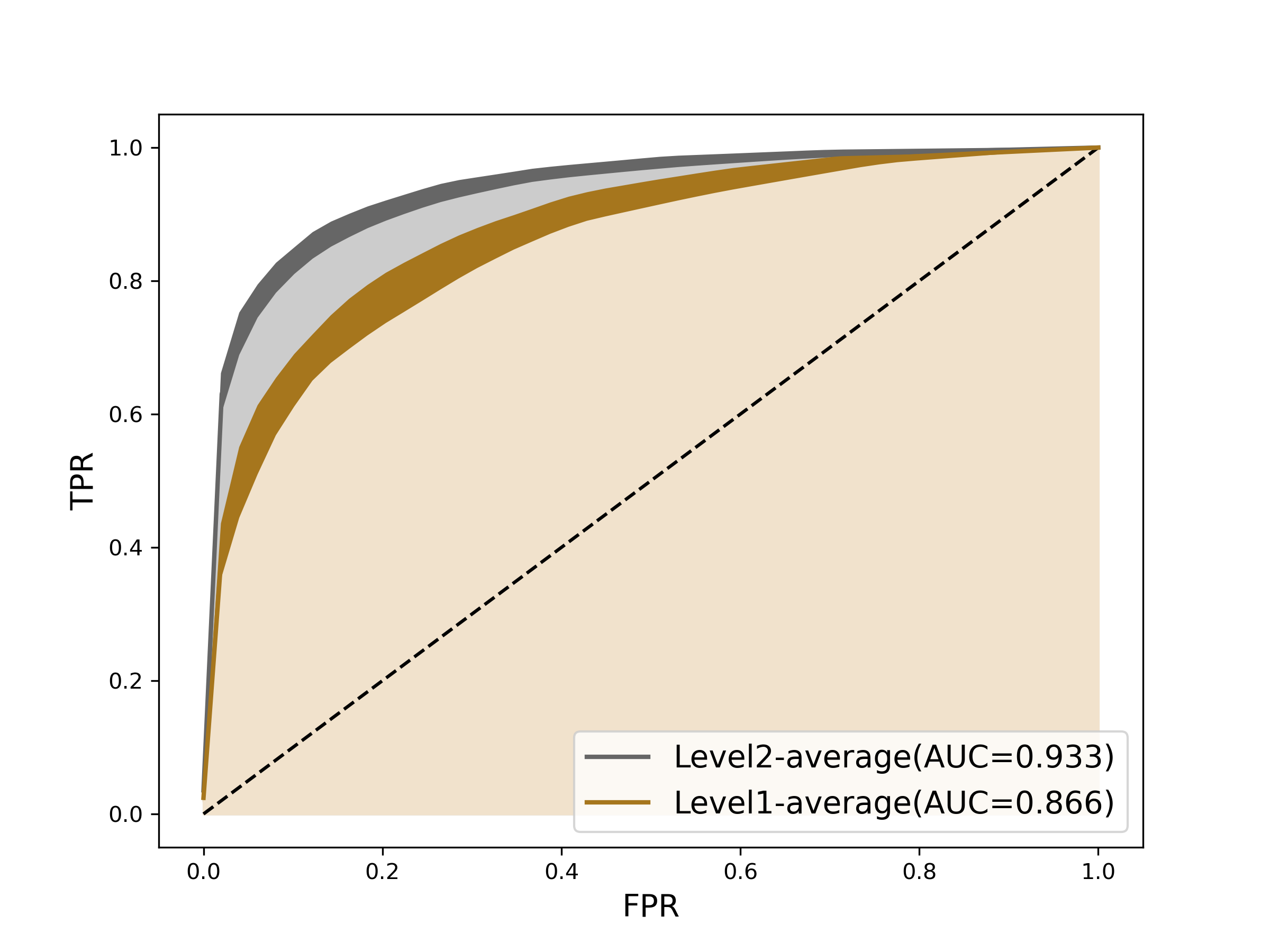}}}\\
    \makebox[\textwidth][c]{\subfloat[Residual Block 2]{%
          \includegraphics[width=2in]{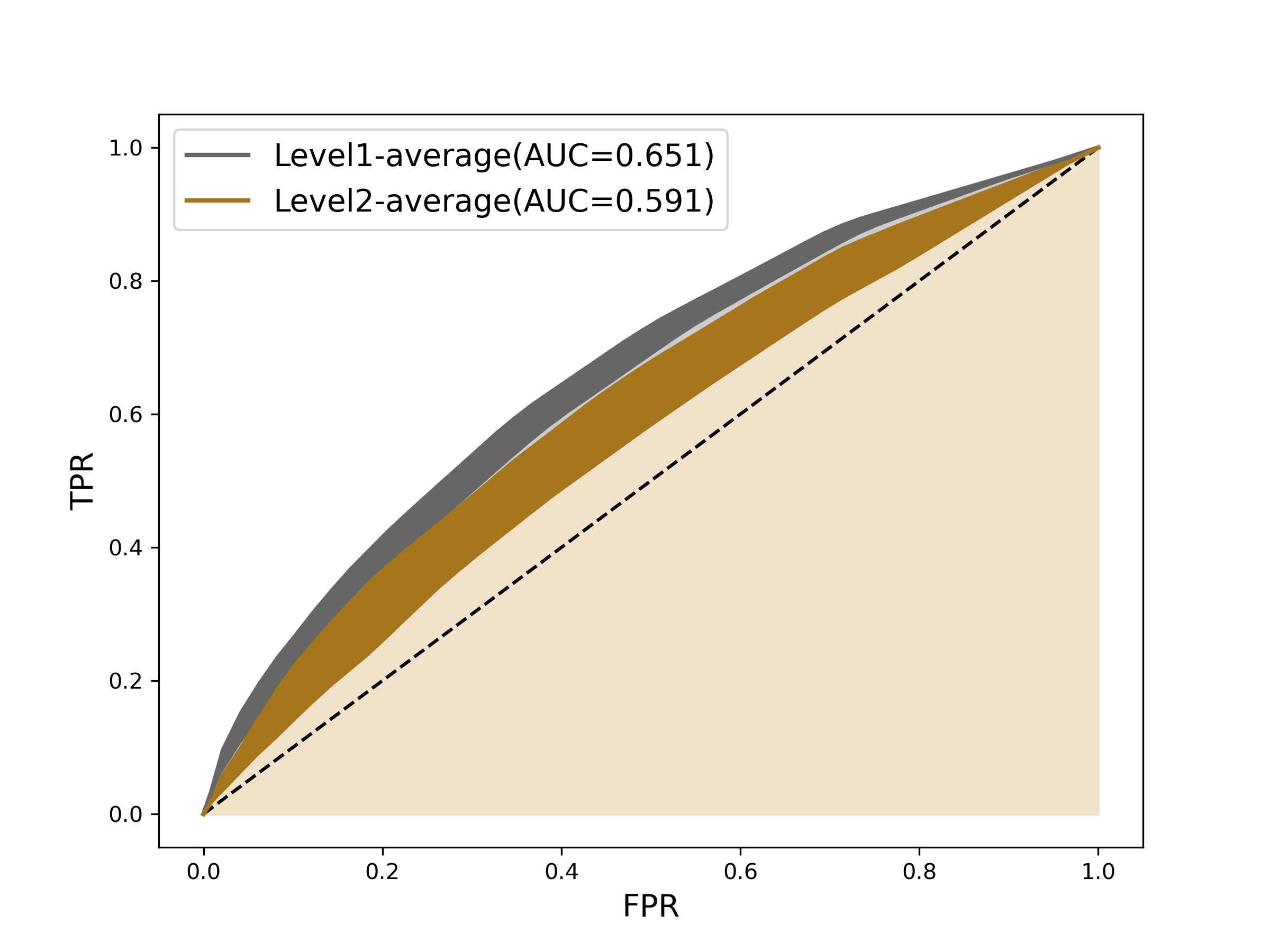}
          \hspace{-0.25in}% 
          \includegraphics[width=2in]{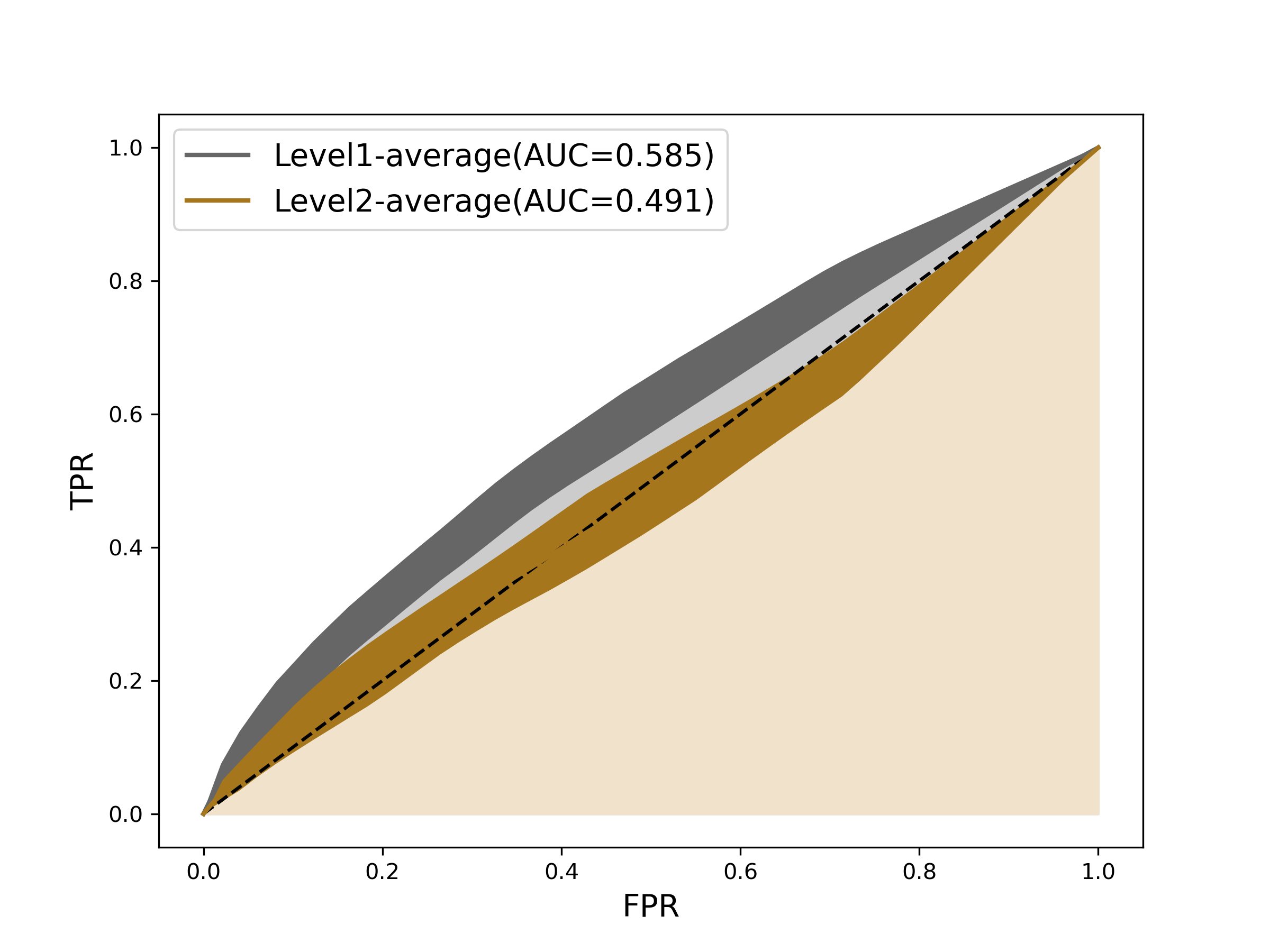}}\hspace{-0.2in}
    \subfloat[Residual Block 3]{%
          \includegraphics[width=2in]{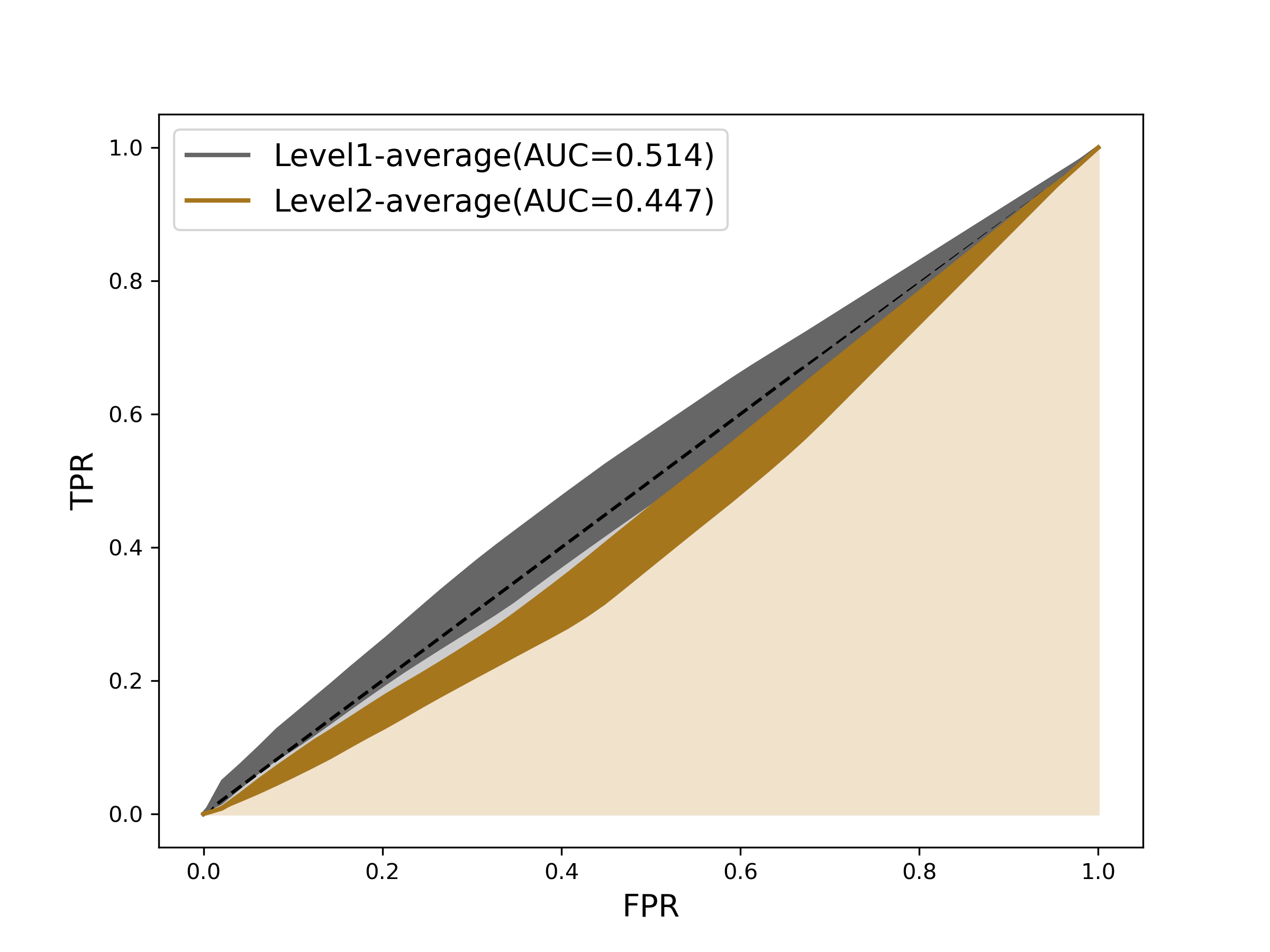} 
          \hspace{-0.25in}% 
          \includegraphics[width=2in]{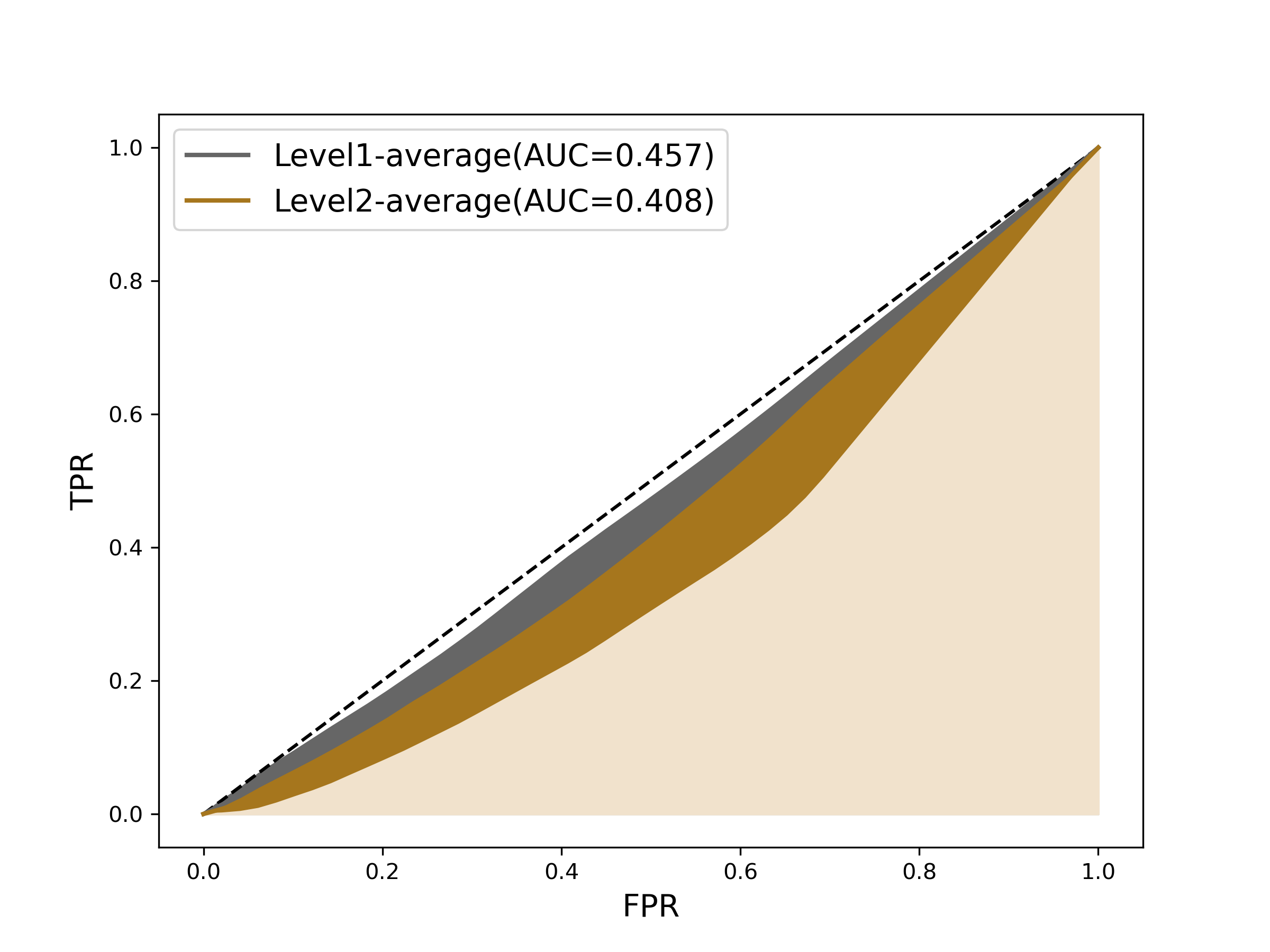}}  }
    \caption{AUROCs of detecting untargeted (left panel) and targeted (right panel) IFGSM attack from RepVGG-A2 at noise levels $\varepsilon=0.03$, via 2DSig-Norm and learned representations from ResNet-20 {(Top row: Normalization \& Residual Block 1; Bottom row: Residual Blocks 2 \& 3 of ResNet-20).}}
    \label{fig:resnet20_layer}  
\end{figure}

The reason for us to choose Residual Block 1 is that, as can be seen from the second row, the model captures and enhances the noise before it begins to learn how to classify objects, i.e., the mean differences for the untargeted and targeted sets are significantly higher than those for the clean set. Additionally, the color variations within the same heat map are subtle. In summary, the distribution difference between clean images and noisy images is much enlarged at this layer while the distribution differences among different classes remain invisible. This observation is important for our purpose: anomaly detection independent of the class of clean images referred to as the training set. Fig. \ref{fig:resnet20_layer} shows the corresponding performance (in term of ROC) of defending against untargeted and targeted attacks using a combination of 
the 2DSig-Norm and different learned representations shown in {the left panel of} Fig. \ref{fig:repvgga2_resnet20}. The AUCs from the second row (Residual Block 1) are significantly higher than the rest choices of layers (Normalisation, Residual Block 2 and 3).

{The right panel of} Fig. \ref{fig:repvgga2_repvgga2} presents the mean differences for different learned representation of RepVGG-A2 while the attack is generated from it too. We observe a similar trend in the first column as in {the left panel of} Fig. \ref{fig:repvgga2_resnet20}. However, we do not see the same patterns as in the bottom row of {the resnet20 part}: with a high untargeted SAR reported in Table \ref{tab:sar}, the model cannot classify correctly the untargeted images, therefore there is no dark diagonal line in the bottom-middle image; with a high targeted SAR reported in Table \ref{tab:sar}, we can see a diagonal line because the model learns from the mislabeled images and identifies them as the current class. Regarding the second row, one can observe a similar pattern as {the left panel of} Fig. \ref{fig:repvgga2_resnet20}. These observations suggest that the earlier layers of a well-trained neural network (designed for classification) focus more on distinguishing polluted images from benign ones rather than on classifying objects.

\begin{figure}
    \centering
    \includegraphics[width=0.6\linewidth]{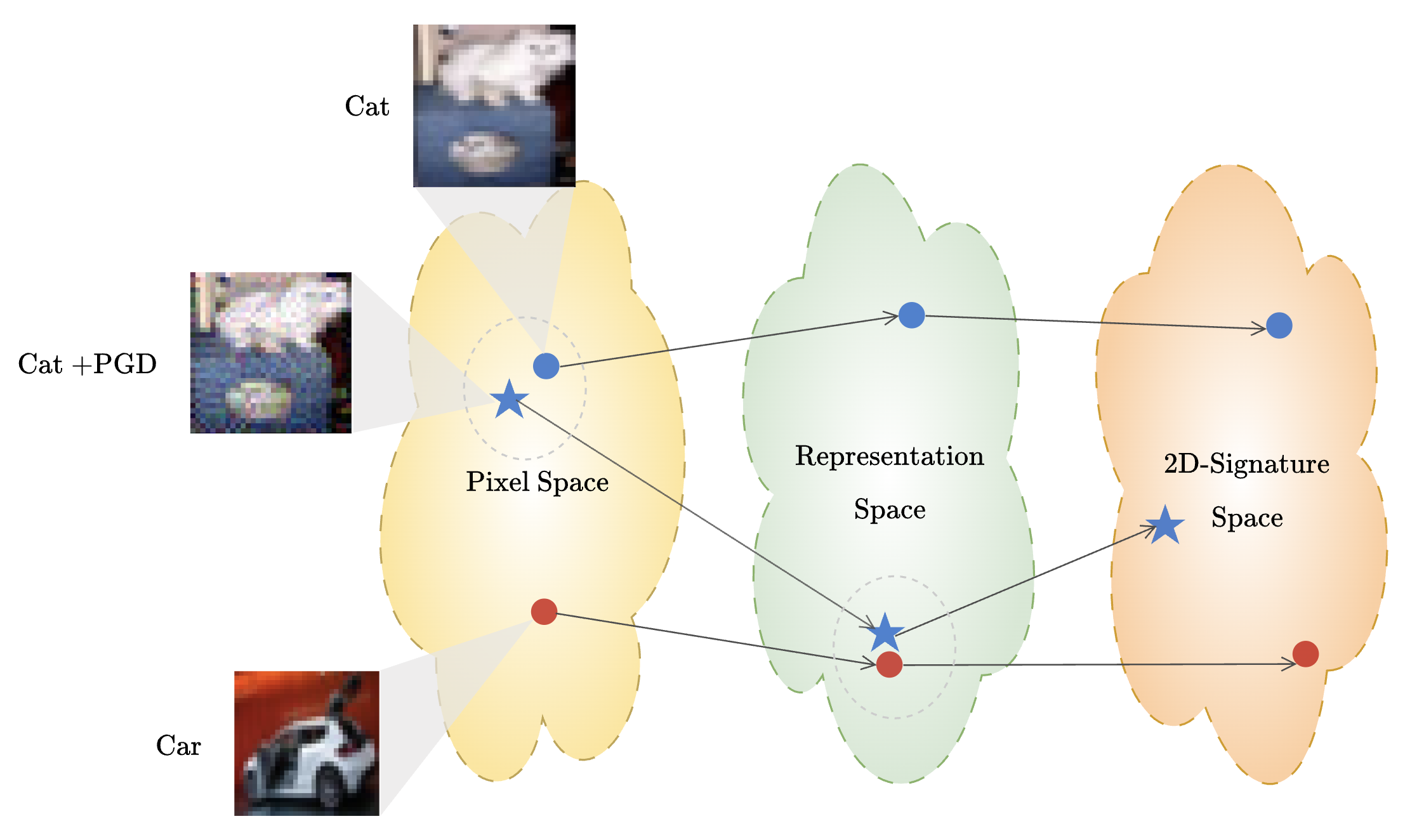}
    \caption{{Schematic representation of why rejecting in the 2D-signature representation space is useful.}}
    \label{fig:novelty}
\end{figure}

{In Figure \ref{fig:novelty}, we illustrate the idea behind why 2D-signatures are so effective at detecting adversarial examples. We see a picture of a cat, and an adversarially perturbed picture of a cat using PGDM. Whilst the adversarial example is similar to the original cat image in the pixel space, it is mapped within the decision boundaries of a car in the representation space. This is what causes the targeted misclassification of an adversarially perturbed cat as a car. It is clear that rejection in the representation space is not enough for computer vision tasks \cite{sabour2016adversarialmanipulationdeeprepresentations}, as the adversarial example is still ``in distribution'' in the representation space due to deliberate feature collision. On the other hand, we demonstrate that rejection in the 2D-signature space has a much higher performance, with adversarial examples returning anomalous distance metrics in this space, using both the covariance norm (Algorithm \ref{alg:covariance}) and the conformance score (Algorithm \ref{alg:conformance}). This is possible because compared to non-signature based methods, computing the 2D-signature allows for a compact description which separates out the cross-channel and self-channel information, capturing the adversarial information effectively.}

\subsubsection{The ``lower-bound" for anomaly detection}
As a sanity check, we test the 2DSig-Norm algorithm with noise level $\epsilon=0$, where the untargeted and targeted test sets are generated by randomly sampling benign test images. In this setting, we would expect the model to behave like a random guess, with a 50\% probability of flagging an image as anomalous. Indeed our experiments show that the average AUCs are 0.50 (level 1) and 0.51 (level 2) for the untargeted test set, and 0.45 (level 1) and 0.47 (level 2) for the targeted test set. These results on clean test data serve as a lower bound for the algorithm's performance, corresponding to random guessing. Our results on perturbed images in Table \ref{tab:ifgsm_f1} and \ref{tab:pgdm_f1} confirm that 2DSig-Norm performs better than the lower bound, with the algorithm's anomaly detection performance increasing with noise level $\epsilon$. 

\subsection{Trainset-level defense}\label{ssec:trainlevel}
There are different types of adversarial attacks in the training phase. This experiment focuses on detecting images with a backdoor within the training set of CIFAR-10. This assumes that the attacker has no access to the model but is able to insert a trigger pattern into a subset of the training data. Whilst the overall accuracy is not much affected after retraining, the attack causes misclassification when prompted with images with the trigger pattern.

\subsubsection{Backdoor attacks}

\begin{figure}[H]
    \centering
    % First row of images
    \subfloat[Airplane - square (top-left)]{
        \begin{minipage}[t]{0.45\textwidth}
            \centering
            \includegraphics[width=0.45\textwidth]{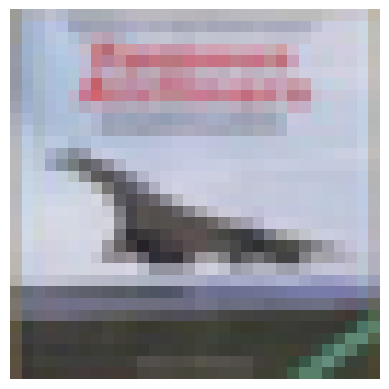}
            \hspace{0.05\textwidth}
            \includegraphics[width=0.45\textwidth]{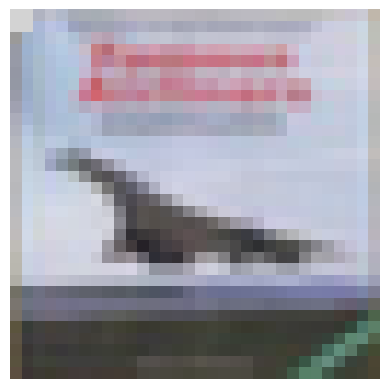}
        \end{minipage}
    }
    \hspace{0.05\textwidth}
    \subfloat[Automobile - dot (bottom-right)]{
        \begin{minipage}[t]{0.45\textwidth}
            \centering
            \includegraphics[width=0.45\textwidth]{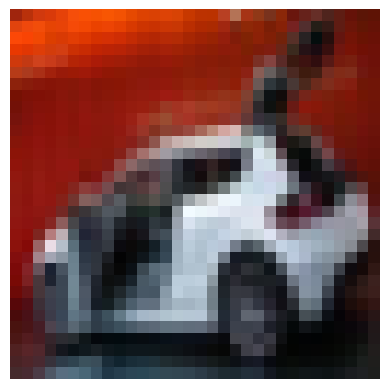}
            \hspace{0.05\textwidth}
            \includegraphics[width=0.45\textwidth]{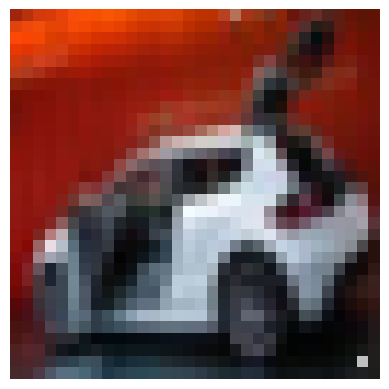}
        \end{minipage}
    }\\
    \vspace{0.05\textwidth}
    % Second row of images
    \subfloat[Cat - rectangle (top-right)]{
        \begin{minipage}[t]{0.45\textwidth}
            \centering
            \includegraphics[width=0.45\textwidth]{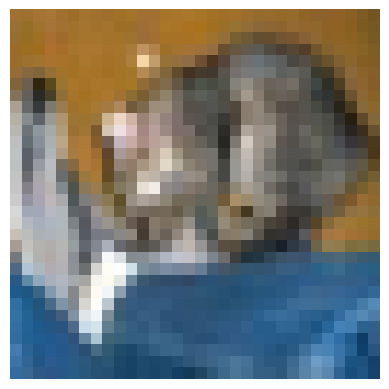}
            \hspace{0.05\textwidth}
            \includegraphics[width=0.45\textwidth]{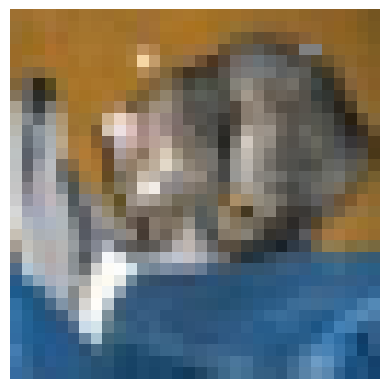}
        \end{minipage}
    }
    \hspace{0.05\textwidth}
    \subfloat[Horse - letter L  (bottom-left) ]{
        \begin{minipage}[t]{0.45\textwidth}
            \centering
            \includegraphics[width=0.45\textwidth]{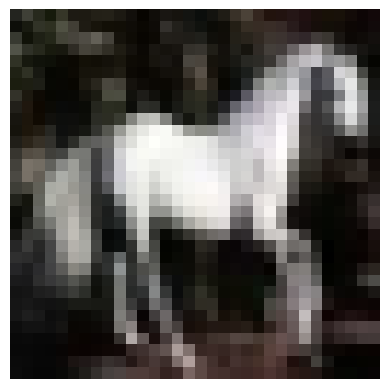}
            \hspace{0.05\textwidth}
            \includegraphics[width=0.45\textwidth]{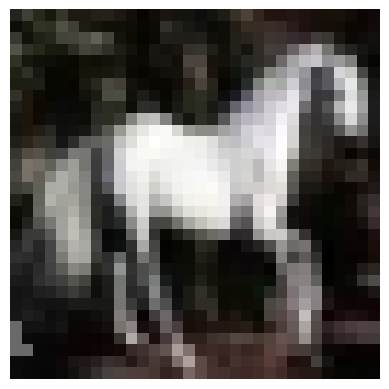}
        \end{minipage}
    }
    \caption{Illustration of Original and Backdoored Images. Each pair of images shows the original image on the left and the image with a backdoor pattern on the right. The patterns are added in the following sequence: square  (top-left), point  (bottom-right),rectangle (top-right) letter L  (bottom-left).}
    \label{fig:backdoor}
    \vspace{0.5em}
\end{figure}

Backdoor attacks are a specific type of data poisoning attack on machine learning models, where an attacker embeds malicious samples with specific triggers within the training data \cite{gu2017badnets, liu2018trojaning}. These triggers cause the model to produce incorrect predictions specified by the attacker when encountered, while performing normally on other clean samples. For our model, following \cite{tran2018spectral}, we choose several specific classes and place distinct backdoor marks (e.g., point, letter L, rectangle, and square) in various locations on the images of each class.  Figure \ref{fig:backdoor} shows examples of these four pairs. Each marked image is then labeled as the target class. The model is then trained with these modified samples and normal samples, ensuring that it misclassifies the backdoor-marked images into the desired target class while maintaining high accuracy on clean images. The models trained are VGG19 \cite{simonyan2014very} and ResNet-18 \cite{he2016deep}, which are commonly used for CIFAR-10 classification task. Their structures are presented in Table \ref{tab:vgg19} in the Appendix.

\subsubsection{The experiment setups}
The goal is to detect backdoored images and restore the integrity of the dataset with minimal loss of clean images. Assuming we have some additional information, such as the maximum number of backdoor images $N_b\in \mathbb{N}$, equivalently, the upperbound of the proportion of images being injected with backdoors\footnote{This can be achieved as normally up to $5\%$ percentage of data is injected with backdoor patterns  \cite{truong2020systematic, saha2022backdoor, zhong2020backdoor}.}, the class that has been targeted and $\beta$\% known clean data of this class ($\beta=5,10$ for example), we can leverage the knowledge of clean data to first extract the high-dimensional learned representations of the distribution of the benign image set. Then, we construct the empirical distribution of these representations through calculating the empirical mean and covariance matrix. 

\begin{algorithm}[!t]
\caption{Backdoor Attack Detection Algorithm (2DSig-Norm version of 2DSig-Detect)\label{alg:backdoor_detection}}
\begin{footnotesize}
\begin{algorithmic}[1]
\STATE \textbf{Input:} $\mathcal{D}_{\text{poison}}$, Poisoned dataset; \\
\STATE \hspace{0.9cm} $\mathcal{N}$, a neural network model providing feature representation $\mathcal{R}_\phi$;\\
\STATE \hspace{0.9cm}  $N_b$, the maximum number of poisoned training examples;\\
\STATE \hspace{0.9cm} $J$ the collection of class labels that have been targeted;\\
\STATE \hspace{0.9cm} $\mathcal{S}$, other transform such as 2D-signature;\\
\STATE \hspace{0.9cm} 
 For each label $j\in J$, let $\mathcal{D}_{j}$ represent the subsets of training examples for that label, and $\mathcal{D}_{j}^{\text{clean}}$ contains $\beta$\% of the clean images of class $j$ which is pre-known.

\STATE \textbf{Output:} Filtered poisoned dataset $\mathcal{D}$.

\STATE \textbf{Initialization:} Randomly initialize the network model $\mathcal{N}$. 
\STATE Train $\mathcal{N}$ on $\mathcal{D}_{\text{poison}}$.
\FOR {each label $j$}
\STATE Split $\mathcal{D}_{j}$ into $\mathcal{D}_{j}^{\text{clean}}$ and $\mathcal{D}^{\text{poison}}_{j}$.
    \STATE Define $n_c$ and $n_p$ as the number of instances in $\mathcal{D}_{j}^{\text{clean}}$ and $\mathcal{D}^{\text{poison}}_{j}$, respectively, and enumerate these instances as $x^{\text{clean}}_1, \ldots, x^{\text{clean}}_{n_c}$ and $x^{\text{poison}}_1, \ldots, x^{\text{poison}}_{n_p}$.
    \STATE Let $X_{\text{clean}} = \left[\mathcal{S}\circ \mathcal{R}_\phi (x^{\text{clean}}_i)\right]_{i=1}^{n_c}$, an $n_c \times d$ representation matrix and $X_{\text{poison}} = \left[\mathcal{S}\circ \mathcal{R}_\phi (x^{\text{poison}}_i) \right]_{i=1}^{n_p}$ an $n_p \times d$ one.
    \STATE Generate the empirical mean $\mu$ and covariance $A$, using $X_{\text{clean}}$ as input, as an intermediate step in Algorithm \ref{alg:covariance}.
    \STATE Compute the score list $\Gamma = \sqrt{\text{diag}\big((X_{\text{poison}}-\mu)A^{-1}(X_{\text{poison}}-\mu)^T\big)}$, is an $n_p$-dimensional vector.
    \STATE Remove the examples with the top $N_b$ scores in $\Gamma$ from $\mathcal{D}_{j}^{\text{poison}}$.
    \STATE $\mathcal{D}_{j} \leftarrow \mathcal{D}_{j}^{\text{poison}}\cup \mathcal{D}_{j}^{\text{clean}}$.
\ENDFOR
\STATE Return $\mathcal{D}$.
\end{algorithmic}
\end{footnotesize}
\end{algorithm}

The learned representation of VGG19 is the output of \emph{VGG Block 4} defined in Table \ref{tab:vgg19}, and for ResNet-18 is the output of \emph{Residual Block 3} defined in the same table. This choice may help us generate high-level features \cite{tran2018spectral} and is supported by evidence in Fig. \ref{fig:layeroutput_vgg19}, which will be deliberated later. Extracting the representations is crucial because it amplifies the backdoor attack signals, making them easier to detect. Using the 2DSig-Norm, we compute the scores of these representations to identify and filter out the attacked samples. The detailed algorithm is shown in Algorithm \ref{alg:backdoor_detection}.

As the number of channels increases from $3$ to either $512$ (for VGG19) or $256$ (for ResNet-18), the dimensionality of the 2D-signature features, especially the second level, increases in a polynomial way. As each channel in the later layers of a trained model aims to capture distinct high-level features \cite{zeiler2014visualizing}, we limit our 2D-signature features to the terms that involves $\mathrm{d} x^i\mathrm{d} x^i$, $\mathrm{d} x^i\hat{ \mathrm{d}} x^i$, $\hat{ \mathrm{d}} x^i\mathrm{d} x^i$ and $\hat{ \mathrm{d}} x^i\hat{ \mathrm{d}} x^i$ only, $i\in [d]$.  For the naive representation, we take the channel mean, which is the collection of the average pixel value per channel. Therefore the size of the naive representation is $512$ (for VGG19) and $256$ (for ResNet-18).

\hspace{0pt}\\\textbf{Experiment 1.}
In this experiment, backdoor attacks are introduced to specific pairs of source and target classes. The following pairs were chosen following \cite{tran2018spectral}: "airplane" to ``bird", ``automobile" to ``cat", ``cat" to ``dog", and ``horse" to ``deer". Four different shapes were added as backdoor triggers: square, point, rectangle, and the ``L" shape. These shapes were embedded in different locations and had various shades of gray, ranging from light gray to dark gray, to ensure the backdoor signals were distinct and detectable. We set $N_b = 500$ for each attacked class, and from these classes, we select $\beta$\% ($\beta=10,20,40$) of the clean data as our known clean dataset $\mathcal{D}^{\text{clean}}_j$.
{We evaluated the models on the test set to measure their accuracy on clean data and their attack success rate, which refers to the proportion of images containing the backdoor pattern that were successfully classified into the target class. The results for the VGG19 and ResNet-18 models trained on $\mathcal{D}_{\text{poison}}$ are presented in Table \ref{tab:backdoor_acc_4pairs}.}

\begin{table}[]
   \caption{Clean accuracy (Acc) on the clean images and Attack Success Rate (ASR) after backdoor attacks of VGG19 and ResNet-18 models under Experiment 1 setting.}
    \label{tab:backdoor_acc_4pairs}
    \centering
\begin{footnotesize}
\begin{tabular}{c|c|c|c|c|c|c|c|c|c}
\hline\hline
\multirow{2}{*}{}  & \multicolumn{2}{c|}{\bf Airplane} & \multicolumn{2}{c|}{\bf Automobile} & \multicolumn{2}{c|}{\bf Cat} & \multicolumn{2}{c|}{\bf Horse} & \multirow{2}{*}{\bf Overall Acc} \\ \cline{2-9} 
                   &  Acc &  ASR &  Acc &  ASR & Acc & ASR & Acc &  ASR & \\ \hline\hline
VGG19              & 90.80\% & 86.90\%  & 94.70\% & 91.90\%  & 91.90\% & 81.70\%  & 91.30\% & 77.50\%  & 88.95\%      \\ \hline
ResNet-18          & 87.70\% & 90.10\%  & 92.00\% & 86.00\%  & 72.90\% & 78.40\%  & 88.00\% & 67.70\%  & 86.00\%   \\ \hline\hline
\end{tabular}
\end{footnotesize}
\end{table}

\hspace{0pt}\\\textbf{Experiment 2.} In the second experiment, we aim to test the scenario where multiple source classes are transformed into a single target class using backdoor triggers. We reuse the same backdoor trigger patterns from Experiment 1. This time, however, the images from four different source classes (``airplane", ``automobile", ``cat", and ``horse") were embedded with backdoor patterns and labeled as ``deer". We produce $125$ images with the backdoor attack from each of the four classes; for the "deer" class, we select $\beta$\% ($\beta=10$) of the clean data as our known clean dataset $\mathcal{D}^{\text{clean}}_j$.
{The accuracy on the clean test set and the attack success rate (ASR) for test set images containing the backdoor pattern, under the Experiment 2 setting, for the VGG19 and ResNet-18 models trained on $\mathcal{D}_{\text{poison}}$, are shown in Table \ref{tab:backdoor_acc_4to1}. }

\begin{table}[]
   \caption{Clean accuracy (Acc) on the clean images and Attack Success Rate (ASR) after backdoor attacks of VGG19 and ResNet-18 models under Experiment 2 setting}
    \label{tab:backdoor_acc_4to1}
    \centering
\begin{footnotesize}
\begin{tabular}{c|c|c|c|c|c|c|c|c|c}
\hline\hline
\multirow{2}{*}{}  & \multicolumn{2}{c|}{\bf Airplane} & \multicolumn{2}{c|}{\bf Automobile} & \multicolumn{2}{c|}{\bf Cat} & \multicolumn{2}{c|}{\bf Horse} & \multirow{2}{*}{\bf Overall Acc} \\ \cline{2-9} 
                   &  Acc & ASR &  Acc & ASR &  Acc &  ASR &  Acc &  ASR & \\ \hline\hline
VGG19              & 91.50\% & 75.20\%  & 95.00\% & 84.70\%  & 78.00\% & 67.20\%  & 92.60\% & 19.70\%  & 89.56\%      \\ \hline
ResNet-18          & 87.80\% & 77.30\%  & 92.00\% & 67.20\%  & 74.50\% & 55.40\%  & 88.80\% & 33.50\%  & 86.12\%   \\ \hline\hline
\end{tabular}
\end{footnotesize}
\end{table}

\hspace{0pt}\\\textbf{Metric.} From Algorithm \ref{alg:backdoor_detection}, TPR is chosen as the metric for both experiments, calculated by the the number of anomalies in the top $N_b$ scores divided by $N_b$. In our case $N_b=500$. 
\subsubsection{Results}
\textbf{Experiment 1.} The main experiment is conducted under the condition that  $500$ many, equivalently $10$\% clean images have been identified per polluted class and used as the training set. We compare the performance of the 2DSig-Norm with respect to the performance of 2DSig-Conf, naive-norm and naive-conf, with accuracy reported in Table \ref{tab:4in4_eachclass_acc_vgg19} for VGG19 and Table \ref{tab:4in4_eachclass_acc_resnet18} for ResNet-18. The performance of the 2DSig-Norm and the 2DSig-Conf is similar at the same level of 2D-signature features, and consistently outperforms the naive ones. With a fixed base framework, the model with level 1 gains better TPR than level 2 across three backdoor patterns (except for the third backdoor
pattern).

\begin{table}[!t]
    \caption{TPR of detecting poisoned images via learned representation of VGG19 (red color marks the highest one for each backdoor pattern).}
    \label{tab:4in4_eachclass_acc_vgg19}
    \centering
    \begin{footnotesize}
    \begin{tabular}{l|cc|cc|c|c}
        \hline\hline
        \multirow{2}{*}{} & \multicolumn{2}{c|}{\bf 2DSig-Norm} & \multicolumn{2}{c|}{\bf 2DSig-Conf} & \multirow{2}{*}{naive-norm} & \multirow{2}{*}{naive-conf} \\ 
        \cline{2-5} 
        & \multicolumn{1}{c|}{Level1} & \multicolumn{1}{c|}{Level2} & \multicolumn{1}{c|}{Level1} & Level2 & & \\ 
        \hline\hline
        (airplane, square) $\rightarrow$ bird & \multicolumn{1}{c|}{\color{red}0.942} & 0.932 & \multicolumn{1}{c|}{0.938} & 0.936 & 0.808 & 0.908 \\ 
        \hline
        (automobile, dot) $\rightarrow$ cat & \multicolumn{1}{c|}{\color{red}0.994} & \color{red}0.994 & \multicolumn{1}{c|}{\color{red}0.994} & \color{red}0.994 & 0.980 & 0.990 \\ 
        \hline
        (cat, rectangle) $\rightarrow$ dog & \multicolumn{1}{c|}{0.762} & 0.730 & \multicolumn{1}{c|}{\color{red}0.766} & 0.732 & 0.668 & 0.686 \\ 
        \hline
        (horse, L) $\rightarrow$ deer & \multicolumn{1}{c|}{\color{red}0.980} & 0.976 & \multicolumn{1}{c|}{\color{red}0.980} & 0.978 & 0.858 & 0.858 \\ 
        \hline\hline
    \end{tabular}
    \end{footnotesize}
\end{table}
\begin{table}[]
   \caption{Accuracy of detecting poisoned images via learned representation of ResNet-18 (red color marks the highest one for each backdoor pattern).}
    \label{tab:4in4_eachclass_acc_resnet18}
       \centering
       \begin{footnotesize}
    \begin{tabular}{l|cc|cc|c|c}
        \hline\hline
        \multirow{2}{*}{} & \multicolumn{2}{c|}{\bf 2DSig-Norm} & \multicolumn{2}{c|}{ 2DSig-Conf} & \multirow{2}{*}{naive-norm} & \multirow{2}{*}{naive-conf} \\ 
        \cline{2-5} 
        & \multicolumn{1}{c|}{Level1} & Level2 & \multicolumn{1}{c|}{Level1} & Level2 & & \\ 
        \hline\hline
        (airplane, square) $\rightarrow$ bird & \multicolumn{1}{c|}{0.912} & 0.802 & \multicolumn{1}{c|}{\color{red}0.914} & 0.802 & 0.814 & 0.798 \\ 
        \hline
        (automobile, dot) $\rightarrow$ cat & \multicolumn{1}{c|}{\color{red}0.980} & 0.978 & \multicolumn{1}{c|}{\color{red}0.980} & 0.978 & 0.978 & \color{red}0.980 \\ 
        \hline
        (cat, rectangle) $\rightarrow$ dog & \multicolumn{1}{c|}{0.808} & \color{red}0.820 & \multicolumn{1}{c|}{0.808} & 0.816 & 0.656 & 0.656 \\ 
        \hline
        (horse, L) $\rightarrow$ deer & \multicolumn{1}{c|}{\color{red}0.962} & 0.960 & \multicolumn{1}{c|}{0.958} & 0.960 & 0.944 & 0.944 \\ 
        \hline\hline
    \end{tabular}
    \end{footnotesize}
\end{table}

\begin{figure}[H]
     \centering
     \includegraphics[width=3in]{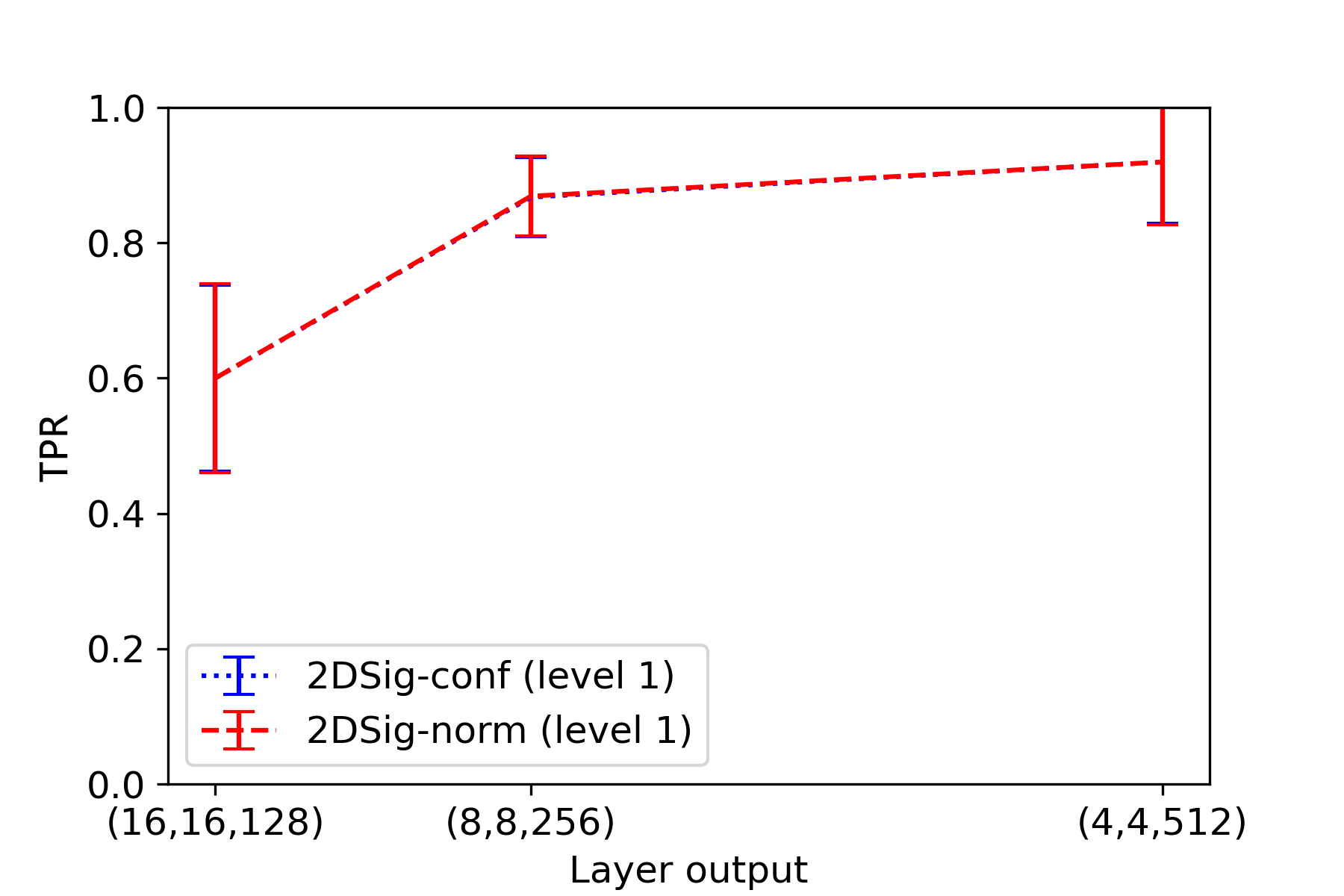}
          \includegraphics[width=3in]{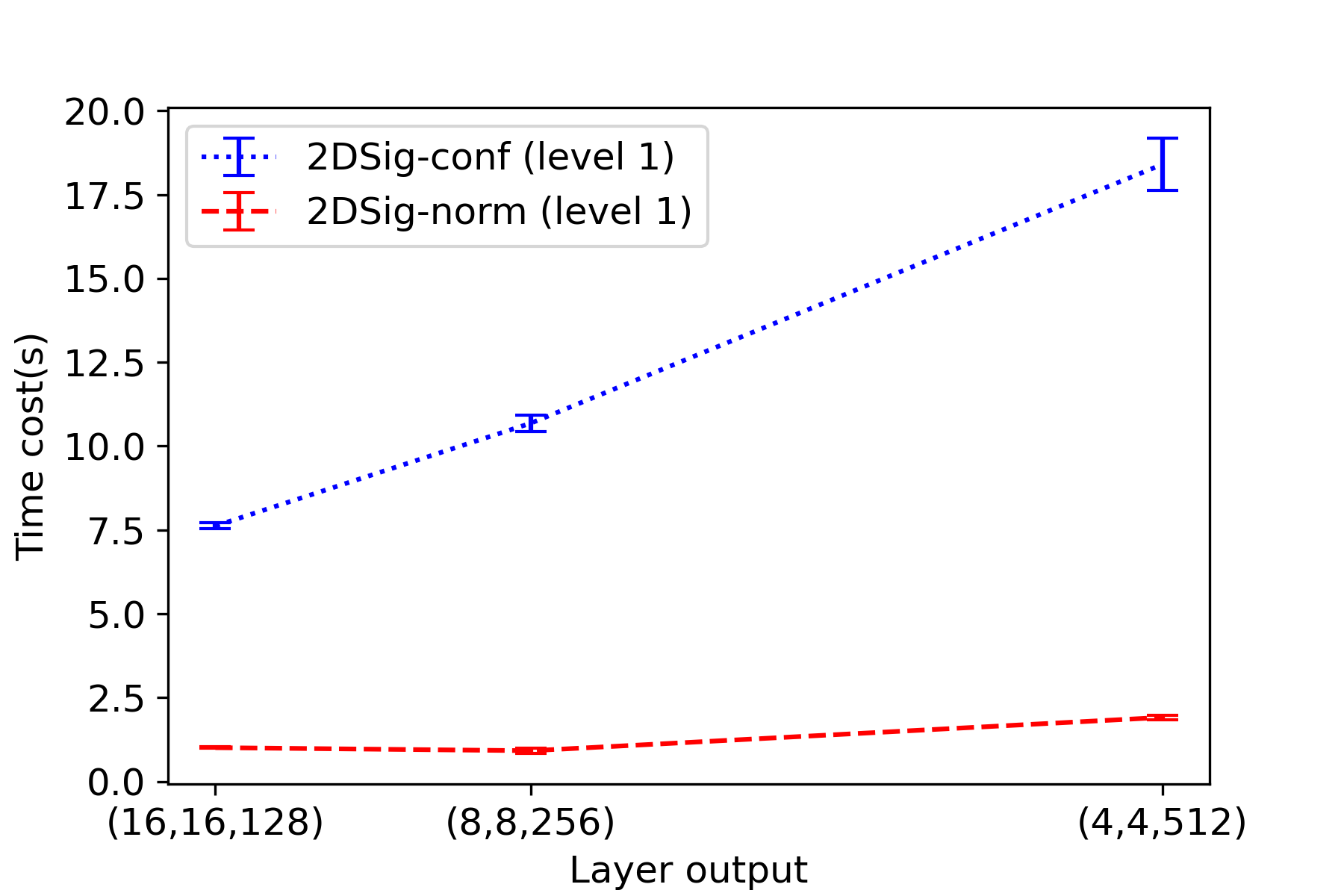}\\
    \caption{TPR and time cost against number of channels (VGG19) for two frameworks.}
    \label{fig:layeroutput_vgg19}  
\end{figure}
\begin{table}[]
   \caption{Performance comparison of models across different proportions of pre-known clean images in the poisoned dataset.}
    \label{tab:4in4_acc_time}
       \centering
       \begin{footnotesize}
\begin{tabular}{llllc}
\hline                                                                                              \hline 
\multirow{2}{*}{}           & \multicolumn{2}{|c}{\bf 2DSig-Norm}                                               & \multicolumn{2}{|c}{2DSig-Conf}                          \\ \cline{2-5} 
                            & \multicolumn{1}{|c|}{TPR}               & \multicolumn{1}{c|}{time (s)}       & \multicolumn{1}{c|}{TPR}               & time (s)        \\ \hline
\hline   
\multicolumn{1}{l|}{$10$\% or 500}  & \multicolumn{1}{l|}{$0.916 \pm 0.067$} & \multicolumn{1}{l|}{\color{red}$5.22\pm 0.10$} & \multicolumn{1}{l|}{$0.915 \pm 0.066$} & $15.19\pm 0.61$ \\ \hline
\multicolumn{1}{l|}{$20$\% or 1000} & \multicolumn{1}{l|}{$0.912 \pm 0.070$} & \multicolumn{1}{l|}{$5.38\pm 0.21$} & \multicolumn{1}{l|}{$0.908 \pm 0.071$} & $76.85\pm 2.21$ \\ \hline\multicolumn{1}{l|}{$40$\% or 2000} & \multicolumn{1}{l|}{\color{red}$0.922 \pm 0.063$} & \multicolumn{1}{l|}{$5.49\pm 0.26$} & \multicolumn{1}{l|}{$0.915 \pm 0.068$} & $146.25\pm 2.60$ \\ \hline \hline                                              
\end{tabular}
\end{footnotesize}
\end{table}

\hspace{0pt}\\\textbf{Learned representation.} To examine the performance of different learned representation of a trained model, we test the TPR and time cost of both the 2DSig-Norm and the 2DSig-Conf (level 1) with learned representations of VGG19 from  \emph{VGG Block 2}, \emph{VGG Block 3} and \emph{VGG Block 4} (defined in Table \ref{tab:vgg19}) respectively. The layer output sizes are $(16,16,128)$, $(8,8,256)$ and $(4,4,512)$. The performance is presented in Fig. \ref{fig:layeroutput_vgg19}. Both frameworks achieve their best TPR with learned features from \emph{VGG Block 4}, while the 2DSig-Conf suffers from an increase in time cost because of the growth of feature dimensionality. One can observe similar pattern for ResNet-18, with learned features from \emph{Residual Block 1}, \emph{Residual Block 2} and \emph{Residual Block 3} respectively (defined in Table \ref{tab:resnet18}).

\hspace{0pt}\\\textbf{Complexity analysis.} To further compare the efficiency of both frameworks, we test their performance in terms of TPR and time cost across different choice of $\beta$, ie, the proportion of benign images that have been identified per polluted class. The results are collected in Table \ref{tab:4in4_acc_time}. The TPR stays in the same level for all $\beta$ and for both methods. The time cost for 2DSig-Norm slightly increases while the one for 2DSig-Conf increases dramatically with $\beta$. In total, 2DSig-Norm uses much shorter time to achieve the same level of TPR. 

\begin{figure}[!t]
    \centering
        \includegraphics[width=0.6\linewidth]{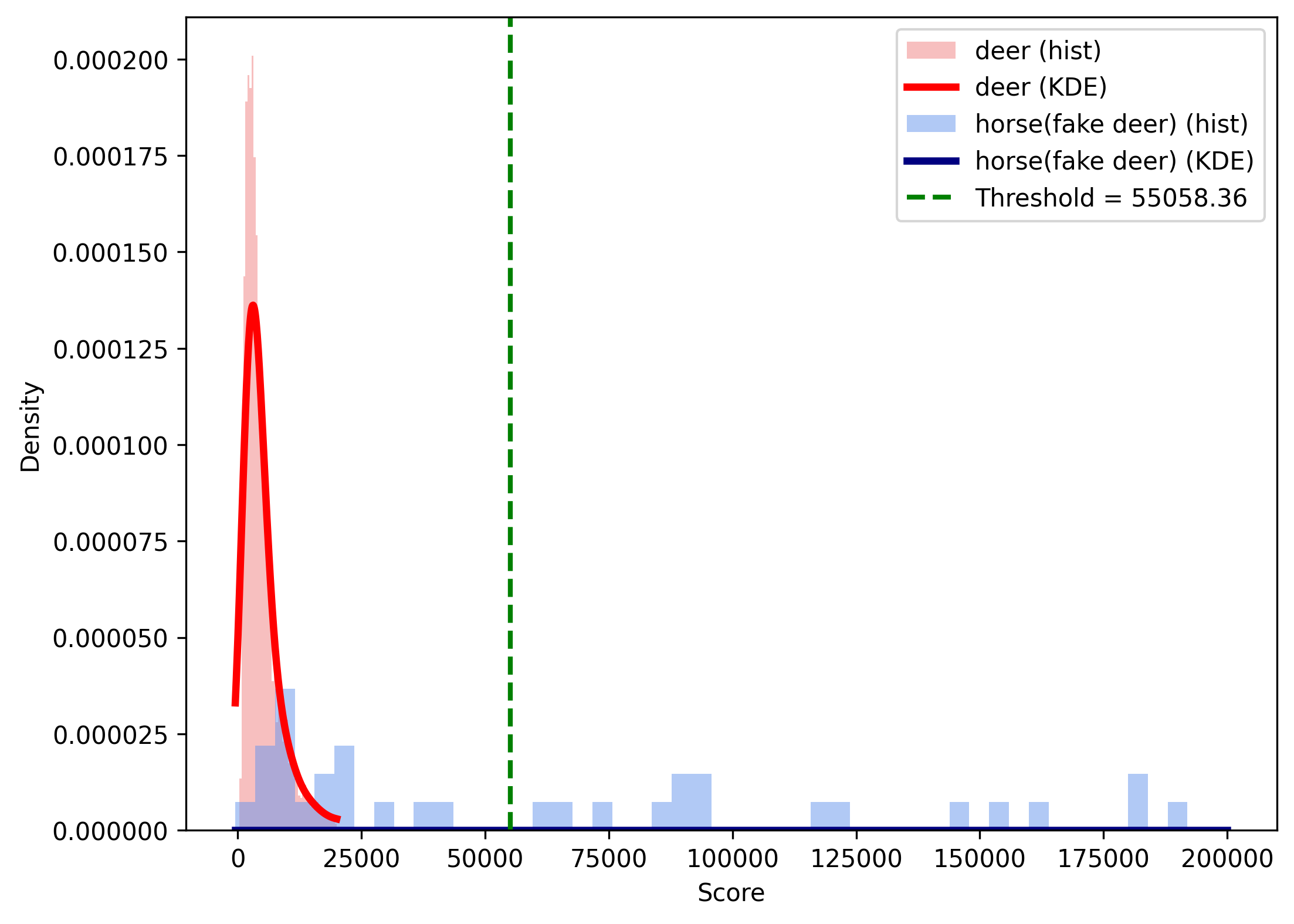}
        \caption{{Distribution of signature values for the (horse, L) $\to$ deer scenario. In this experiment, there were 5000 training examples correctly labelled (red) and 500 poisoned examples incorrectly labelled (blue). The values were sorted in an descending order and the threshold was taken at the top 500th score.}}
    \label{fig:deer}
\end{figure}

\hspace{0pt}\\\textbf{Experiment 2.} In Experiment 2, conditioning on that 10\% of clean images have been identified for the polluted "deer" class and used as the training set. We also compare the performance of 2DSig-Norm with that of 2DSig-Conf, naive-norm, and naive-conf, with accuracy reported in Table \ref{tab:4to1_eachclass_acc_vgg19} for VGG19 and Table \ref{tab:4to1_eachclass_acc_resnet18} for ResNet-18. The experimental results indicated outcomes similar to those of Experiment 1, demonstrating that our algorithm can effectively handle multiple backdoor attacks within the same class. 
\begin{table}[!t]
   \caption{TPR of detecting poisoned images via learned representation of VGG19 under Experiment 2 setting (red color marks the highest one for each backdoor pattern).}
    \label{tab:4to1_eachclass_acc_vgg19}
    \centering
    \begin{footnotesize}
\begin{tabular}{l|cc|cc|c|c}
\hline\hline
                   & \multicolumn{2}{c|}{\bf 2DSig-Norm}                                 & \multicolumn{2}{c|}{ 2DSig-Conf}                                 &                               &                               \\ \cline{2-5}
\multirow{-2}{*}{} & \multicolumn{1}{c|}{Level1}                        & Level2                        & \multicolumn{1}{c|}{Level1}                        & Level2                        & \multirow{-2}{*}{naive-norm}  & \multirow{-2}{*}{naive-conf}  \\ \hline\hline
airplane           & \multicolumn{1}{c|}{{\color{red} 0.9440}} & 0.9360                        & \multicolumn{1}{c|}{{\color{red} 0.9440}} & 0.9360                        & {\color{red} 0.9440} & {\color{red} 0.9440} \\ \hline
automobile         & \multicolumn{1}{c|}{{\color{red} 0.9760}} & {\color{red} 0.9760} & \multicolumn{1}{c|}{{\color{red} 0.9760}} & {\color{red} 0.9760} & {\color{red} 0.9760} & {\color{red} 0.9760} \\ \hline
cat                & \multicolumn{1}{c|}{0.7280}                        & 0.7280                        & \multicolumn{1}{c|}{0.7280}                        & {\color{red} 0.7360} & 0.7200                        & 0.7120                        \\ \hline
horse              & \multicolumn{1}{c|}{{\color{red} 0.9360}} & {\color{red} 0.9360} & \multicolumn{1}{c|}{{\color{red} 0.9360}} & {\color{red} 0.9360} & 0.9200                        & 0.9200                        \\ \hline\hline
\end{tabular}%
\end{footnotesize}
\end{table}

\begin{table}[!t]
   \caption{Accuracy of detecting poisoned images via learned representation of ResNet-18 under Experiment 2 setting (red color marks the highest one for each backdoor pattern).}
    \label{tab:4to1_eachclass_acc_resnet18}
    \centering
    \begin{footnotesize}
\begin{tabular}{l|cc|cc|c|c}
\hline \hline
                   & \multicolumn{2}{c|}{\bf 2DSig-Norm}                                 & \multicolumn{2}{c|}{2DSig-Conf}                                 &                               &                         \\ \cline{2-5}
\multirow{-2}{*}{} & \multicolumn{1}{c|}{Level1}                        & Level2                        & \multicolumn{1}{c|}{Level1}                        & Level2                        & \multirow{-2}{*}{naive-norm}  & \multirow{-2}{*}{naive-conf}  \\ \hline  \hline    
airplane           & \multicolumn{1}{c|}{{\color{black} 0.9440}} & {\color{black} 0.9280} & \multicolumn{1}{c|}{{\color{black} 0.9440}} & {\color{black} 0.9280} & {\color{red} 0.9520} & {\color{red} 0.9520} \\ \hline
automobile         & \multicolumn{1}{c|}{{\color{red} 1.0000}} & {\color{black} 0.9920} & \multicolumn{1}{c|}{{\color{red} 1.0000}} & {\color{black} 0.9920} & {\color{red} 1.0000} & {\color{red} 1.0000} \\ \hline
cat                & \multicolumn{1}{c|}{{\color{red} 0.8400}} & {\color{black} 0.8320} & \multicolumn{1}{c|}{{\color{red} 0.8400}} & {\color{black} 0.8320} & {\color{black} 0.8320} & {\color{black} 0.8320} \\ \hline
horse              & \multicolumn{1}{c|}{{\color{red} 0.9440}} & {\color{red} 0.9440} & \multicolumn{1}{c|}{{\color{red} 0.9440}} & {\color{red} 0.9440} & {\color{black} 0.9200} & {\color{black} 0.9200} \\ \hline  \hline    
\end{tabular}%
\end{footnotesize}
\end{table}
\section{Discussion}\label{sec:discussion}
%[Varun]
In this paper we proposed a mathematically-principled semi-supervised framework for anomaly detection in image data, called 2DSig-Detect. This framework is based on using 2D-signature embedded features of learned representations to achieve better discrimination between distributions. {It is worth noting that common poisoning and adversarial attacks can be considered as specific, highly targeted anomalies, designed to deceive standard detection methods while remaining invisible to the human eye. Therefore, our 2DSig-Detect framework enhances the sensitivity of anomaly detection systems by utilizing advanced feature representation techniques, thereby improving the detection capabilities against these meticulously designed attacks.} We presented examples using different types of attacks to demonstrate the effectiveness of our approach in detecting security threats using 2DSig-Detect (specifically using the 2DSig-Norm variant). We benchmarked against GMM, a naive-norm approach, and the 2DSig-Conf variant of 2DSig-Detect. Our algorithm performs effectively under both variants, however our preferred variant, 2DSig-Norm operates at up to 1/10th of the cost of 2DSig-Conf.

%[Varun] 
In Section \ref{ssec:twoframeworks} we outlined key factors for the effective performance of 2DSig-Detect: the feature dimensionality should be small, and the distributions must be sufficiently distinct, as indicated by the mean difference \eqref{eqn:meandiff}. We addressed the former requirement by using truncated 2D-signatures to provide faithful low-dimensional feature representations for images \cite{diehl2024signature}. The second condition is challenging to fulfill, as by their nature a adversarial attack aims to only modify the image by a small magnitude to mitigate the risk of detection. Hence it is difficult for a human observer to detect adversarial image attacks. In order to increase distribution separation, in Section \ref{ssec:role_lr} we demonstrated the importance of using appropriate learned representations (from a well-trained neural network) to emphasize the distinction between distributions of benign and poisoned images.

The distance between the empirical means of two distributions, standardised under the covariance norm as defined in \eqref{eqn:meandiff}, which coincides with the Kullback-Leibler divergence under certain conditions, can serve as an informal indicator for identifying/validating which layer(s) of the trained model provides the suitable learned representation for anomaly detection. By leveraging the 2D-signature transform, which captures both the aggregated linear and nonlinear effects of pixel values per channel of the image, and combining it with robust learned representations from neural networks, 2DSig-Detect enhances the detection and mitigation of adversarial manipulations of image data. This makes it a promising approach for anomaly detection in various applications. Moreover, this framework, by its nature, shows robustness for different kinds of attacks as well as for different image resolutions. 

\subsection{Future Work}
%[Varun] 
As it is only recently that multiple variants of the 2D-signature were proposed in the literature, there is little theoretical work rigorously investigating the properties of these objects. As an natural extension of (1D-) path signatures, we would expect the 2D-signature to inherit most of the properties of the one dimensional version. For example, by definition the 2D-signature should be invariant to stretching (see \cite[Definition 4.22]{diehl2024signature}): the 2D-signatures of one continuous image in $\mathcal{Z}_d$ over $[0,1]^2$ (see Definition \ref{def:image_data} and Definition \ref{def:2d_signature}) remain the same regardless of the speed going over $[0,1]^2$. However, this property does not hold after discretization, ie, the 2D-signatures differ when the same continuous image is discretized at different resolutions $(N,M,\cdot)$, $N,M\in\mathbb{N}$. As a result, we observe a much larger performance gap between 2DSig-Norm and the naive-norm in Section \ref{ssec:testlevel} than in Section \ref{ssec:trainlevel}. One possible explanation for this is that the learned representation size in the latter case is only $(4,4,512)$; in this case, the 2D-signature cannot capture much more information than the flattened image representation. In the future we could investigate how the 2D-signature evolves in response to varying degrees of information loss. Additionally, unlike (1D-) signatures \cite{shao2023dimensionless}, it is not the case that performance increases with level of the 2D-signature. In fact, we only saw in the evasion attack experiment (Section \ref{ssec:testlevel}) that using 2DSig-Norm with Level-2 features improved performance compared with only using Level-1. This highlights the importance of studying the performance of models under different levels of 2D-signature features.

As mentioned in Section \ref{ssec:2D-signature}, there are different types of 2D-signature defined in literature, including the 2D-id-signature \cite{diehl2024signature} and 2D-signature derived from the Jacobian minor \cite{giusti2022topological}.
In \cite{diehl2024signature}, the authors show that the 2D-id-signature, defined using $\mathrm{d} x^i$ terms only, lacks a shuffle property i.e.  the iterated integrals in \eqref{eqn:2dsig_full} do not form an algebra, thus they are not stable under multiplication. They define a revised version of the 2D-id-signature, named the symmetrized 2D-signature,  by summing over all the permutation possibilities of the second parameter, which does have the shuffle property. An interesting direction of research would be to investigate the performance of the symmetrized 2D-signature, though there are numerical challenges to address prior to this, and compare with the existing 2D-signature features used in this work.

{In order to build from our proof-of-concept application, we would need to perform a more comprehensive assessment of how our signature-based method compares to other mitigation techniques, particularly other anomaly detection techniques~\cite{granz2024weiperooddetectionusing, sotgiu2020deepneuralrejectionadversarial}. In addition, we would need to conduct an assessment on the limitations of our detection method for threat models with increasing adversary capabilities and perturbation budgets. Such as assessment will support a rigorous defence evaluation to validate any robustness claims~\cite{blueteamplaybook,carlini2019evaluatingadversarialrobustness}.}
\section{Data Availability}\label{sec:data_availability}
{
The data and example code related to this project will be made available on GitHub. Please follow the link below for updates: \url{https://github.com/xiexinheng/2DsignatureAnomalyDetection}.}
\subsection*{Acknowledgments}
We extend our gratitude to Dr. Sam Morley for his proofreading and to the computational support of Mathematical Institute, University of Oxford, via the EPSRC under the program grant EP/S026347/1.
\subsection*{Conflict of interest}
None of the authors have a conflict of interest to disclose.

\bibliographystyle{elsarticle-num}
\bibliography{references.bib}
\appendix
% \appendix
% \begin{appendices}
\section{Proofs}\label{sec:app_proofs}
\begin{proof}[The proof of Proposition \ref{prop: covnorm_eq}]
Because $V=\mathbb{R}^d$,  it is not hard to verify that $\text{cov}(f,f)=\langle f,  Af\rangle=\langle A^{1/2}f,  A^{1/2}f\rangle=|A^{1/2}f|^2$, where the inner product $\langle \cdot,  \cdot\rangle$ is defined on the Euclidean space $\mathbb{R}^d$.   Then we can rewrite Eqn.  \eqref{eqn:covnorm} as follows:
\begin{align*}
&\|x\|_\mathcal{L}^2=\sup_{\text{cov}(f,f)\leq 1}f(x)^2=\sup_{f\in \mathbb{R}^d/\{\mathbf{0}\}}\frac{\langle f,  x\rangle^2}{\text{cov}(f,f)}=\sup_{f\in \mathbb{R}^d/\{\mathbf{0}\}}\frac{\langle A^{1/2}f,  A^{-1/2}x\rangle^2}{\langle A^{1/2}f,  A^{1/2}f \rangle}\\
&\quad \leq \sup_{f\in \mathbb{R}^d/\{\mathbf{0}\}}\frac{\langle A^{1/2}f,  A^{1/2}f\rangle \langle A^{-1/2}x,  A^{-1/2}x\rangle}{\langle A^{1/2}f,  A^{1/2}f \rangle}=\langle A^{-1/2}x,  A^{-1/2}x\rangle=\langle x,  A^{-1}x\rangle .
\end{align*}
On the other hand,  for simplicity,  assume that the first element of $x$ is nonzero, i.e., $x^{1}\neq 0$. Then we can take $f=(|A^{-1/2}x|/x^{1},0,\ldots,0)^T$ and get $f(x)=|A^{-1/2}x|$. Therefore $f(x)^2\leq \|x\|_\mathcal{L}^2$. The assertion is confirmed. 
\end{proof}
\begin{proof}[Proposition \ref{prop: covnorm}]
By the multivariate Chebshev inequality, it is easy to deduce that
$$   \mathbb{P}\left(\big|A^{-\frac{1}{2}}(y-\mu)\big| > \delta \Big| y \sim \mathcal{L}_c\right) \leq \frac{d}{\delta^2}. $$
Assume that $\mu_u\neq \mu$. Take $\delta \in \big(\sqrt{d},|A^{-\frac{1}{2}}(\mu_u-\mu)|\big)$, then it suffices to show the inequality \eqref{eqn:prop_covnorm_abnormal}.
Indeed:
\begin{footnotesize}
    \begin{align*}
    &\mathbb{P}\left(\big| A^{-\frac{1}{2}}(y-\mu)\big| < \delta\, \big|\, y \sim \mathcal{L}_u\right) = \mathbb{P}\left(\big|A^{-\frac{1}{2}}(y-\mu_u) + A^{-\frac{1}{2}}(\mu_{a}-\mu)\big| < \delta\, \big| \,y \sim \mathcal{L}_{u}\right) \\
    \leq & \mathbb{P}\left(\big|A^{-\frac{1}{2}}(y-\mu_u)\big| > \big|A^{-\frac{1}{2}}(\mu_u - \mu)\big| - \delta \,\big|\, y \sim \mathcal{L}_{u}\right) 
    \leq  \frac{d\big\|A^{-\frac{1}{2}} A_{u}^{\frac{1}{2}}\big\|^2}{(|A^{-\frac{1}{2}}(\mu_u-\mu)|-\delta)^2}.
\end{align*}
\end{footnotesize}
Taking $\delta = \frac{\big|A^{-\frac{1}{2}}(\mu_u - \mu)\big|}{1+\big\|A^{-\frac{1}{2}} A_{\mu}^{\frac{1}{2}}\big\|} $ reduces to \eqref{eqn:prop_covnorm_abnormal2}.
\end{proof}
\begin{proof}[Proposition \ref{prop: conformancescore}]
Note that if $x \in \mathcal{L}_c$, then $\mathbb{E}\left[x+ y_i\right]=0$ and 
$\operatorname{Var}\left(x-y_i\right)=\operatorname{Var}(X)+\operatorname{Var}(y_i)=2A $. Therefore,
\begin{footnotesize}
\begin{align*}
  & \mathbb{P}\left(\text{dist}(x ; \mathcal{L}_c)>\delta \mid x \sim \mathcal{L}_c\right)  =\mathbb{P}\left(\min _{i \in[n]}\| x-y_i\|_{\mathcal{L}_c}>\delta \mid x \sim \mathcal{L}_c\right) \\
& =\prod_{i \in[n]} \mathbb{P}\left(\|x-y_i\|_{\mathcal{L}_c} >\delta\mid x \sim \mathcal{L}_c\right)  \leqslant \prod_{i \in[n]} \frac{2 d}{\delta^2}=\left(\frac{2 d}{\delta^2}\right)^n.
\end{align*}
\end{footnotesize}
We shall choose $\delta> \sqrt{2d}$ so the right hand side is smaller than one.

If $x$ has mean $\mu_u$ and covariance $A_u$, then
\begin{footnotesize}
\begin{align*}
    & \operatorname{Var}\left(A^{-\frac{1}{2}}\left(\left(\mu_u-x\right)-\left(\mu-y_i\right)\right)\right)  =\operatorname{Var}\left(A^{-\frac{1}{2}}\left(\mu_{u}-x\right)\right)+\operatorname{Var}\left(A^{-\frac{1}{2}}\left(\mu-y_i\right)\right) \\
&\quad =A^{-\frac{1}{2}} A_u A^{-\frac{1}{2}}+A^{-\frac{1}{2}} A A^{-\frac{1}{2}}  =A^{-\frac{1}{2}} A_u A^{-\frac{1}{2}}+I.
\end{align*}
\end{footnotesize}

Now taking $\Delta_\mu:=\left|A^{-\frac{1}{2}}\left(\mu_u-\mu\right)\right|$ and $\Delta_A:=A^{-\frac{1}{2}} A_u A^{-\frac{1}{2}}+I$ yields
\begin{footnotesize}
\begin{align*}
    &\mathbb{P}\left(\left\|x-y_i\right\|_{\mathcal{L}_c  }<\delta \mid x \nsim \mathcal{L}_c\right)   \\
     &\leq \mathbb{P}\left(\left|A^{-\frac{1}{2}}\left(\left(\mu_u-x\right)-\left(\mu-y_i\right)\right)\right|>\Delta_\mu-\delta \mid x \nsim \mathcal{L}_c\right) \leq \frac{d\big\|\Delta_A^{\frac{1}{2}}\big\|^2}{\left(\Delta_\mu-\delta\right)^2},
\end{align*}
\end{footnotesize}
where we shall choose $\delta<\Delta_\mu$ and  
$\frac{d\big\|\Delta_A^{\frac{1}{2}}\big\|^2}{\left(\Delta_\mu-\delta\right)^2}<1$.
The estimate above can be used to deduce the following:
\begin{footnotesize}
\begin{align*}
    & \mathbb{P}\left(\operatorname{dist}\left(x ; \mathcal{L}_c\right)<\delta \mid x \nsim \mathcal{L}_c\right) = 1-\mathbb{P}\left(\min _{i \in[n]}\left\|x-y_i\right\|_{\mathcal{L}_c  }\geq \delta \mid x \nsim \mathcal{L}_c\right) \\ 
    &= 1-\prod_{i\in [n]}\mathbb{P}\left(\left\|x-y_i\right\|_{\mathcal{L}_c  }\geq \delta \mid x \nsim \mathcal{L}_c\right)  =1-\prod_{i\in [n]}\Big(1-\mathbb{P}\left(\left\|x-y_i\right\|_{\mathcal{L}_c  }<\delta \mid x \nsim \mathcal{L}_c\right) \Big) \\ 
    & \leqslant 1-\left(1-\frac{ d\big\|\Delta_A^{\frac{1}{2}}\big\|^2}{\left(\Delta_\mu-\delta\right)^2}\right)^n.
\end{align*}
\end{footnotesize}
\end{proof}

\section{The discretization for the level 2 of 2D-signature}\label{sec:app_dis}

\textbf{The 2nd level} of $S$ contains 4$d^2$ elements $$\{ \mathrm{d} x^i \mathrm{d} x^j,\mathrm{d} x^i \hat{\mathrm{d}} x^j, \hat{\mathrm{d}} x^i \mathrm{d} x^j , \hat{\mathrm{d}} x^i \hat{\mathrm{d}} x^j, i,j\in [d]\}$$ of four types,  
where the first two reinforce the effects of the same type and the last two capture interaction between different effects:
\begin{footnotesize}
\begin{align*}
&   \int\int_{(0,0)}^{(1,1)} \Big(\int\int_{(0,0)}^{(s_1,t_1)} \frac{\partial^2 x^{i}(s_2,t_2)}{\partial s_2 \partial t_2}\mathrm{d}s_2\mathrm{d}t_2\Big)\frac{\partial^2 x^{i}(s_1,t_1)}{\partial s_1 \partial t_1}\mathrm{d}s_1\mathrm{d}t_1\\
   &\approx\sum^{N-1}_{k_1 = 1} \sum^{M-1}_{k_2 = 1}\Big[\Big(\sum^{k_1-1}_{k_3 = 1} \sum^{k_2-1}_{k_4 = 1} \left(\mathbf{x}^{i}_{k_3 + 1, k_4 + 1} - \mathbf{x}^{i}_{k_3, k_4+1} - \mathbf{x}^{i}_{k_3+1, k_4} + \mathbf{x}^{i}_{k_3, k_4 }\right)\Big)\\
&\qquad \times \left(\mathbf{x}^{j}_{k_1 + 1, k_2 + 1} - \mathbf{x}^{j}_{k_1, k_2+1} - \mathbf{x}^{j}_{k_1+1, k_2} + \mathbf{x}^{j}_{k_1, k_2 }\right)
\Big]\\
&=\sum^{N-1}_{k_1 = 1} \sum^{M-1}_{k_2 = 1}\Big[\left(\mathbf{x}^{i}_{k_1 , k_2 } - \mathbf{x}^{i}_{k_1, 1} - \mathbf{x}^{i}_{1, k_2} + \mathbf{x}^{i}_{1, 1 }\right)\\
&\qquad \times \left(\mathbf{x}^{j}_{k_1 + 1, k_2 + 1} - \mathbf{x}^{j}_{k_1, k_2+1} - \mathbf{x}^{j}_{k_1+1, k_2} + \mathbf{x}^{j}_{k_1, k_2 }\right)\Big];
\end{align*}
\end{footnotesize}
and 
\begin{footnotesize}
\begin{align*}
&     \int\int_{(0,0)}^{(1,1)} \Big(\int\int_{(0,0)}^{(s_1,t_1)}\frac{\partial x^{i}(s_2,t_2)}{\partial s_2 }\frac{\partial x^{i}(s_2,t_2)}{\partial t_2}\mathrm{d}s_2\mathrm{d}t_2\Big)\frac{\partial x^{i}(s_1,t_1)}{\partial s_1 }\frac{\partial x^{i}(s_1,t_1)}{\partial t_1 }\mathrm{d}s_1\mathrm{d}t_1\\
   & \approx  \sum^{N-1}_{k_1 = 1} \sum^{M-1}_{k_2 = 1}\Big[\Big(\sum^{k_1-1}_{k_3 = 1} \sum^{k_2-1}_{k_4 = 1} \left(\mathbf{x}^{i}_{k_3 + 1, k_4} - \mathbf{x}^{i}_{k_3, k_4}\right) \times \left(\mathbf{x}^{i}_{k_3, k_4+1} - \mathbf{x}^{i}_{k_3, k_4 }\right)\Big)\\
&\qquad \times \left(\mathbf{x}^{j}_{k_1 + 1, k_2} - \mathbf{x}^{j}_{k_1, k_2}\right) \times \left(\mathbf{x}^{j}_{k_1, k_2+1} - \mathbf{x}^{j}_{k_1, k_2 }\right)
\Big].
\end{align*}
\end{footnotesize}
Similarly, for the latter two we shall have the following approximation
\begin{footnotesize}
\begin{align*}
    \sum^{N-1}_{k_1 = 1} \sum^{N-1}_{k_2 = 1}\Big[&\Big(\sum^{k_1-1}_{k_3 = 1} \sum^{k_2-1}_{k_4 = 1} \left(\mathbf{x}^{i}_{k_3 + 1, k_4} - \mathbf{x}^{i}_{k_3, k_4}\right)\times \left(\mathbf{x}^{i}_{k_3, k_4+1} - x^{i}_{k_3, k_4 }\right)\Big)\\
&\qquad\times \left(\mathbf{x}^{j}_{k_1 + 1, k_2 + 1} - \mathbf{x}^{j}_{k_1, k_2+1} - \mathbf{x}^{j}_{k_1+1, k_2} + \mathbf{x}^{j}_{k_1, k_2 }\right)\Big];
\end{align*}
\end{footnotesize}
and
\begin{footnotesize}
    \begin{align*}
   & \sum^{N-1}_{k_1 = 1} \sum^{M-1}_{k_2 = 1}\Big[\Big(\sum^{k_1-1}_{k_3 = 1} \sum^{k_2-1}_{k_4 = 1} \left(\mathbf{x}^{i}_{k_3 + 1, k_4 + 1} - \mathbf{x}^{i}_{k_3, k_4+1} - \mathbf{x}^{i}_{k_3+1, k_4} + \mathbf{x}^{i}_{k_3, k_4 }\right)\Big) \left(\mathbf{x}^{j}_{k_1 + 1, k_2} - \mathbf{x}^{j}_{k_1, k_2}\right) \left(\mathbf{x}^{j}_{k_1, k_2+1} - \mathbf{x}^{j}_{k_1, k_2 }\right)
\Big]\\
&= \sum^{N-1}_{k_1 = 1} \sum^{M-1}_{k_2 = 1}\Big[\left(\mathbf{x}^{i}_{k_1 , k_2 } - \mathbf{x}^{i}_{k_1, 1} - \mathbf{x}^{i}_{1, k_2} + \mathbf{x}^{i}_{k_1, k_1 }\right)  \left(\mathbf{x}^{j}_{k_1 + 1, k_2} - \mathbf{x}^{j}_{k_1, k_2}\right) \left(\mathbf{x}^{j}_{k_1, k_2+1} - \mathbf{x}^{j}_{k_1, k_2 }\right)
\Big].
\end{align*}
\end{footnotesize}

\section{Tables and plots}\label{sec:app_plots}
\setlength{\tabcolsep}{3pt} 
\begin{table}[!t]
    \centering
       \caption{ Overview of Model Architectures. From left to right: ResNet-18, ResNet-20, RepVGG-A2, and VGG-19. Layers highlighted in red indicate those used for representation extraction in this study.}
    \begin{adjustwidth}{-1cm}{-1cm} % 可调整左右边距
    \begin{minipage}{0.24\textwidth} % 每个表格宽度
    \scriptsize
        \centering
        \begin{tabular}{|c|c|}
            \hline
            Layer Type                        & Output Shape                 \\ \hline
            Input                             & (3, 32, 32)                  \\ \hline
            \color{red} Residual Block 1      & (64, 16, 16)                 \\ \hline
            \color{red} Residual Block 2      & (128, 8, 8)                  \\ \hline
            \color{red} Residual Block 3      & (256, 4, 4)                  \\ \hline
            Residual Block 4                  & (512, 2, 2)                  \\ \hline
            Pooling Layer                      & (512, 1, 1)                  \\ \hline
            Fully Connected                   & (10)                         \\ \hline
        \end{tabular}
        \caption*{ResNet-18}
        \label{tab:resnet18}
        
    \end{minipage}
    \hfill
    \begin{minipage}{0.24\textwidth}
    \scriptsize
        \centering
        \begin{tabular}{|c|c|}
            \hline
            Layer Type                  & Output Shape \\ \hline
            Input                       & (3, 32, 32)  \\ \hline
            \color{red}Residual Block 1 & (16, 32, 32) \\ \hline
            \color{red}Residual Block 2 & (32, 16, 16) \\ \hline
            \color{red}Residual Block 3 & (64, 8, 8)   \\ \hline
            Pooling Layer               & (64, 1, 1)   \\ \hline
            Fully Connected             & (10)         \\ \hline
        \end{tabular}
        \caption*{ResNet-20}
        \label{tab:resnet20}
        
    \end{minipage}
    \hfill
    \begin{minipage}{0.24\textwidth}
    \scriptsize
        \centering
        \begin{tabular}{|c|c|}
            \hline
            Layer Type                  & Output Shape  \\ \hline
            Input                       & (3, 32, 32)   \\ \hline
            \color{red}RepVGG Block 1   & (96, 32, 32)  \\ \hline
            \color{red}RepVGG Block 2   & (192, 16, 16) \\ \hline
            \color{red}RepVGG Block 3   & (384, 8, 8)   \\ \hline
            \color{red}RepVGG Block 4   & (1408, 4, 4)  \\ \hline
            Pooling Layer               & (1408, 1, 1)  \\ \hline
            Fully Connected             & (10)          \\ \hline
        \end{tabular}
         \caption*{RepVGG-A2}
        \label{tab:repvgg_a2}
       
    \end{minipage}
    \hfill
    \begin{minipage}{0.24\textwidth}
    \scriptsize
        \centering
        \begin{tabular}{|c|c|}
            \hline
            Layer Type             & Output Shape  \\ \hline
            Input                  & (3, 32, 32)   \\ \hline
            VGG Block 1            & (64, 32, 32)  \\ \hline
            \color{red}VGG Block 2 & (128, 16, 16) \\ \hline
            \color{red}VGG Block 3 & (256, 8, 8)  \\ \hline
            \color{red}VGG Block 4 & (512, 4, 4)   \\ \hline
            VGG Block 5            & (512, 2, 2)   \\ \hline
            Pooling layer          & (512, 1, 1)   \\ \hline
            Fully Connected        & (10)          \\ \hline
        \end{tabular}
         \caption*{VGG-19}
        \label{tab:vgg19}
       
    \end{minipage}
    \end{adjustwidth}

\end{table}

% \end{appendices}
\begin{figure}[!t]
    \centering
    \makebox[\textwidth][c]{%
          \includegraphics[width=2in]{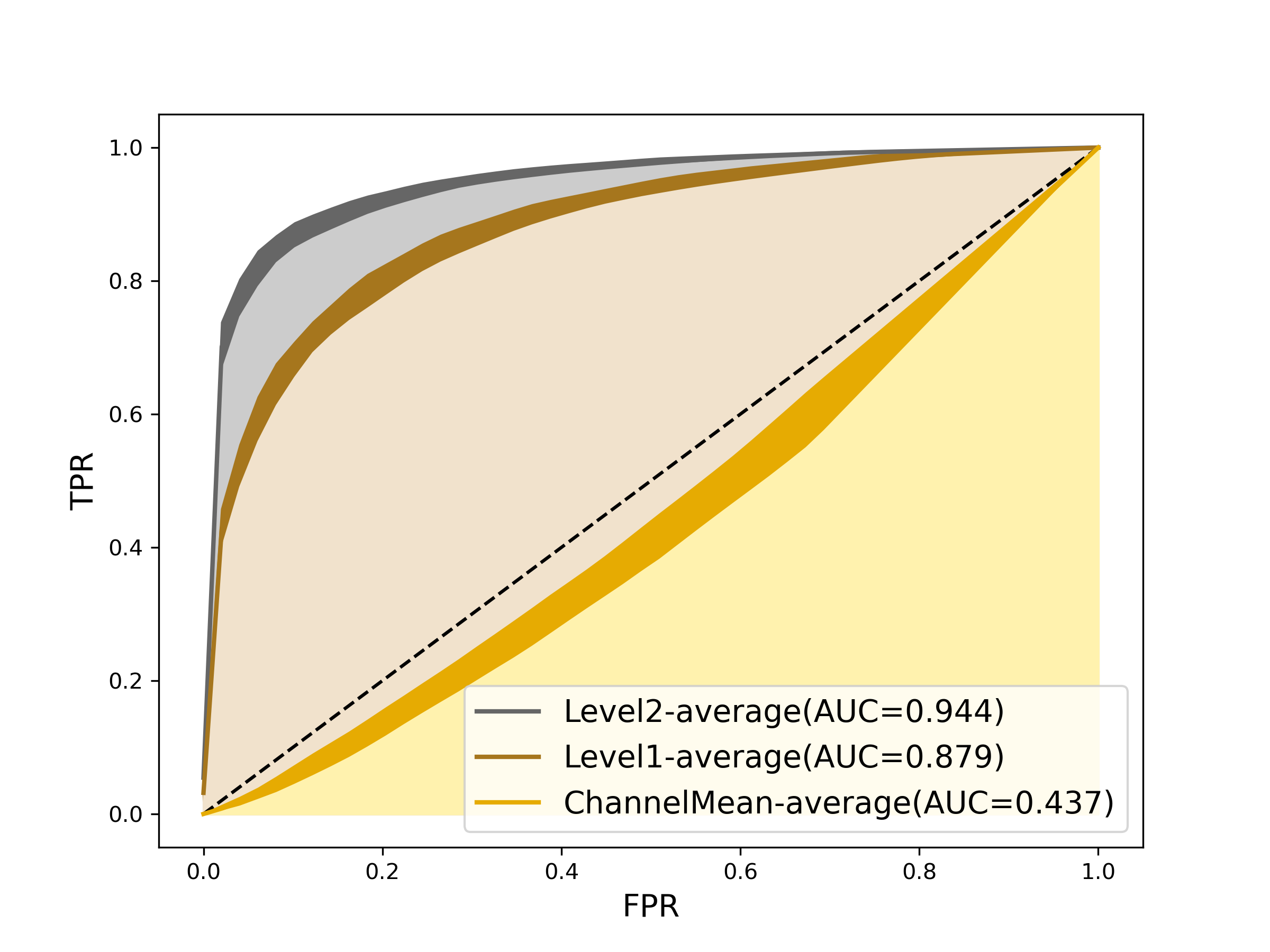}    
          \hspace{-0.2in}% 
                    \includegraphics[width=2in]{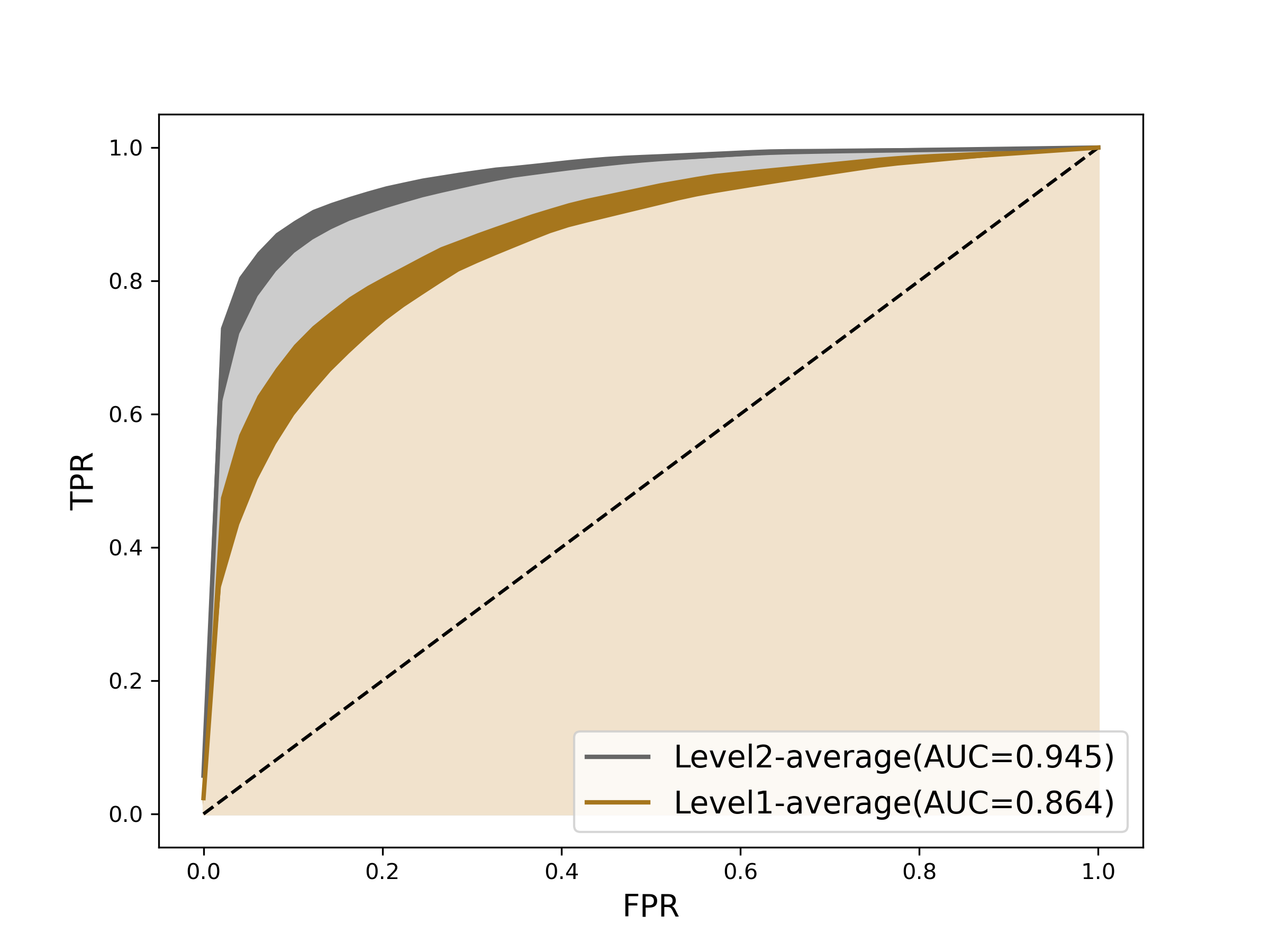} 
                    \hspace{-0.2in}% 
                              \includegraphics[width=2in]{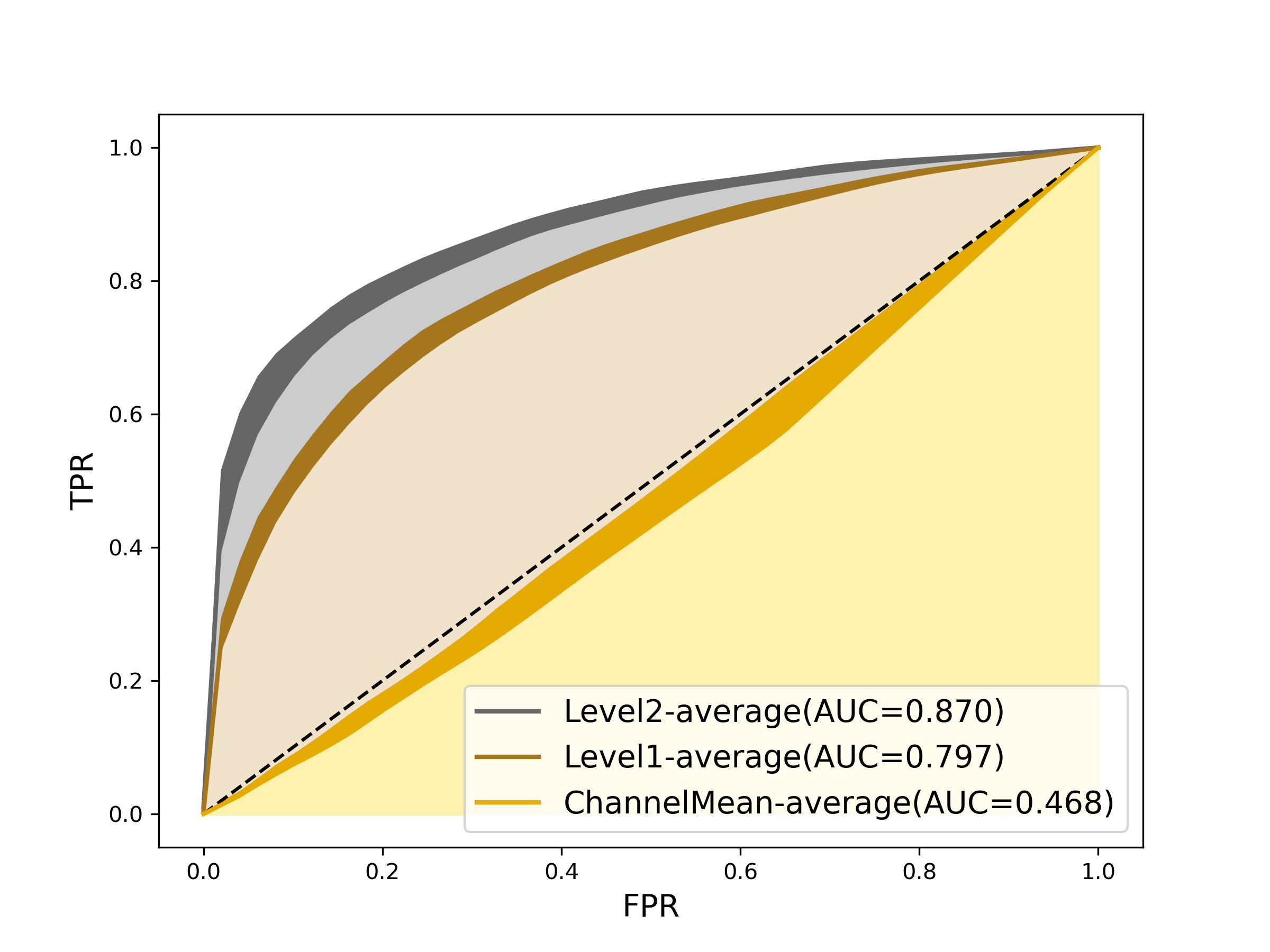}    
                              \hspace{-0.2in}% 
                    \includegraphics[width=2in]{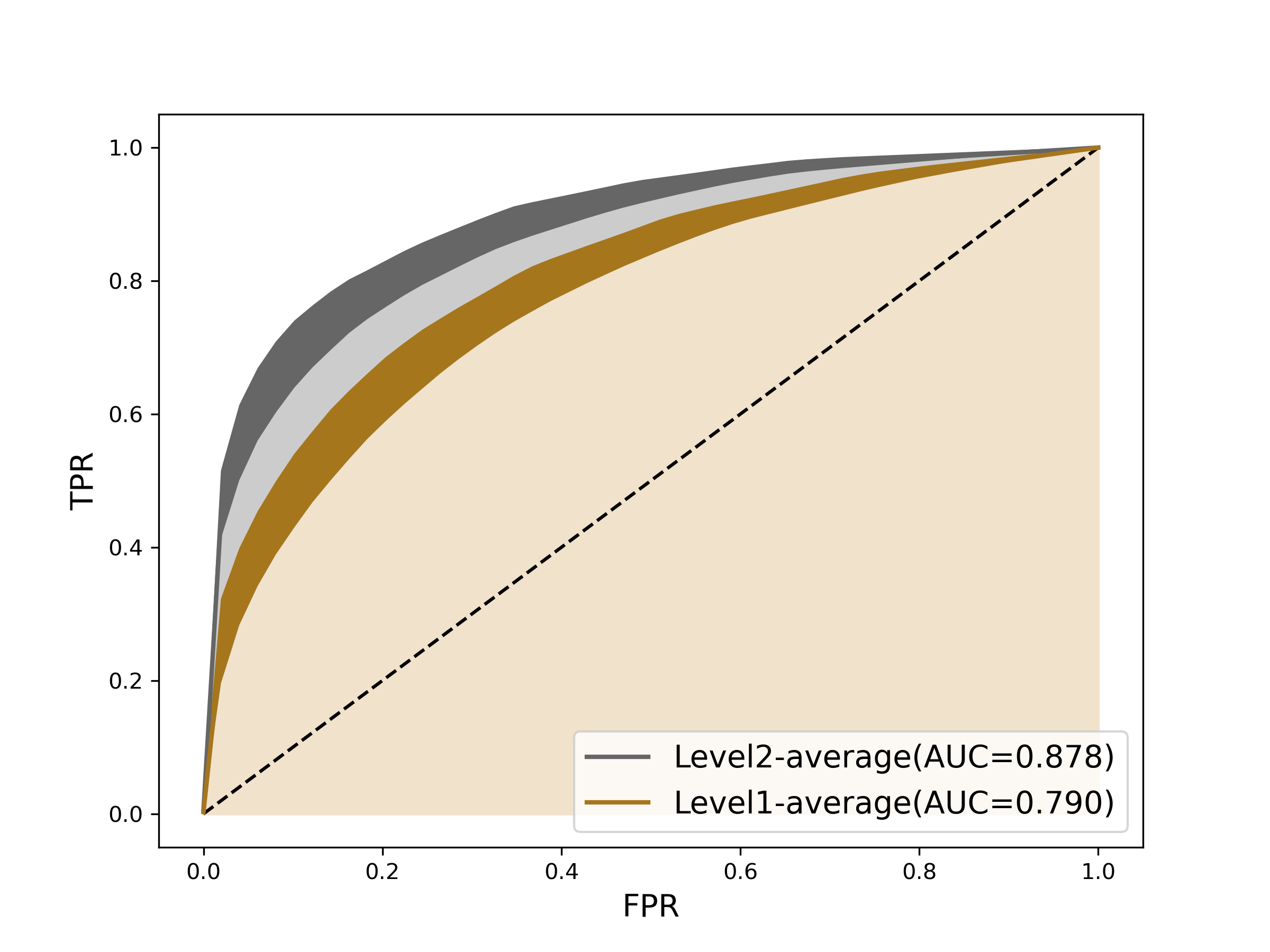} }
    \caption{AUROCs of detecting untargeted IFGSM attack from RepVGG-A2 with noise levels $\varepsilon=0.03$ and $\varepsilon=0.02$. From left to right: \text{2DSig-Norm} ($\varepsilon=0.03$), \text{2DSig-Conf} ($\varepsilon=0.03$), \text{2DSig-Norm} ($\varepsilon=0.02$), \text{2DSig-Conf} ($\varepsilon=0.02$).}
    \label{fig:repvgga2_resnet20_auc_ifgsm_untarget}  
\end{figure}

\begin{figure}[!t]
    \centering
    \makebox[\textwidth][c]{%
        \includegraphics[width=2in]{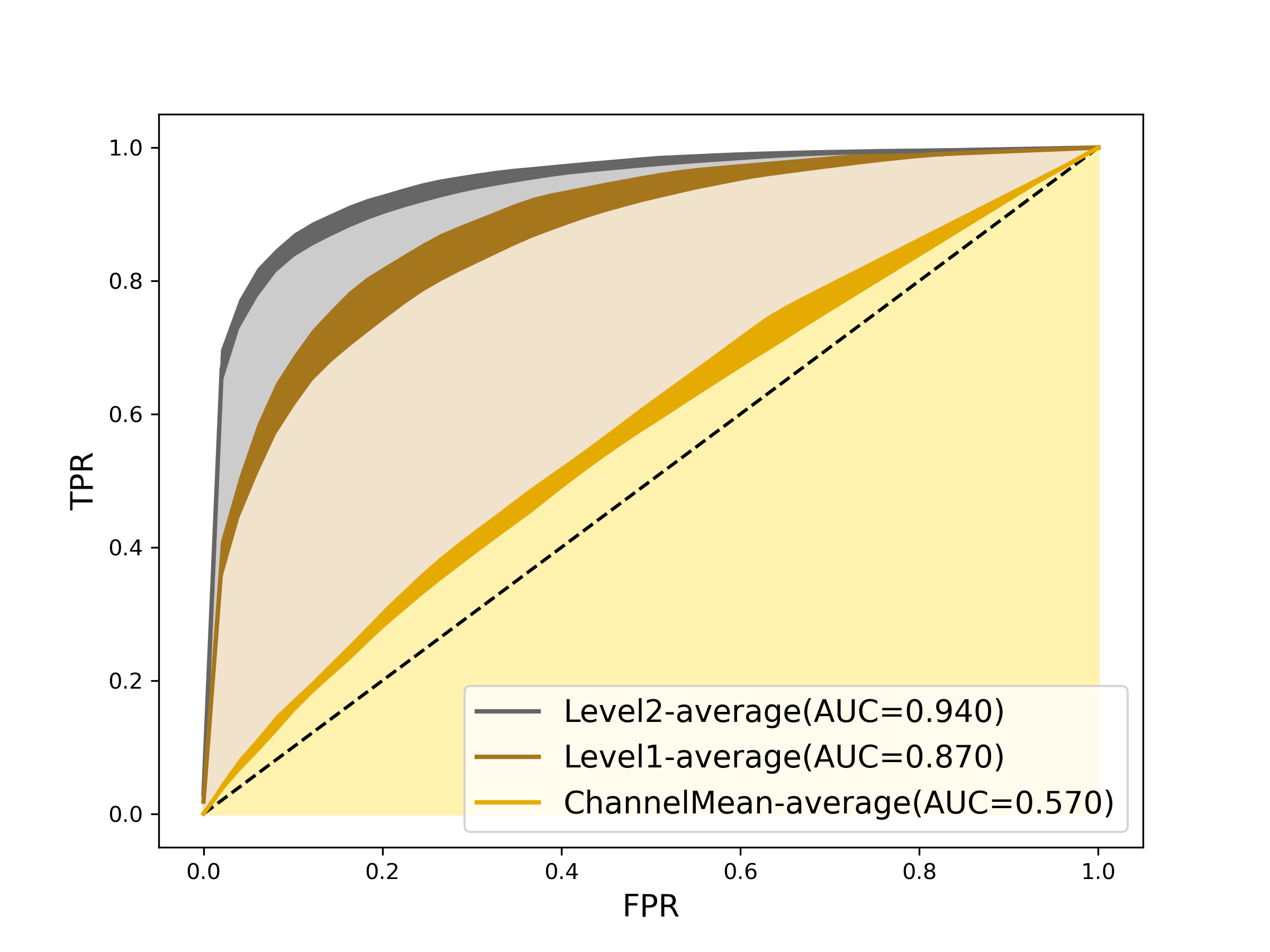}%
        \hspace{-0.2in}% 
        \includegraphics[width=2in]{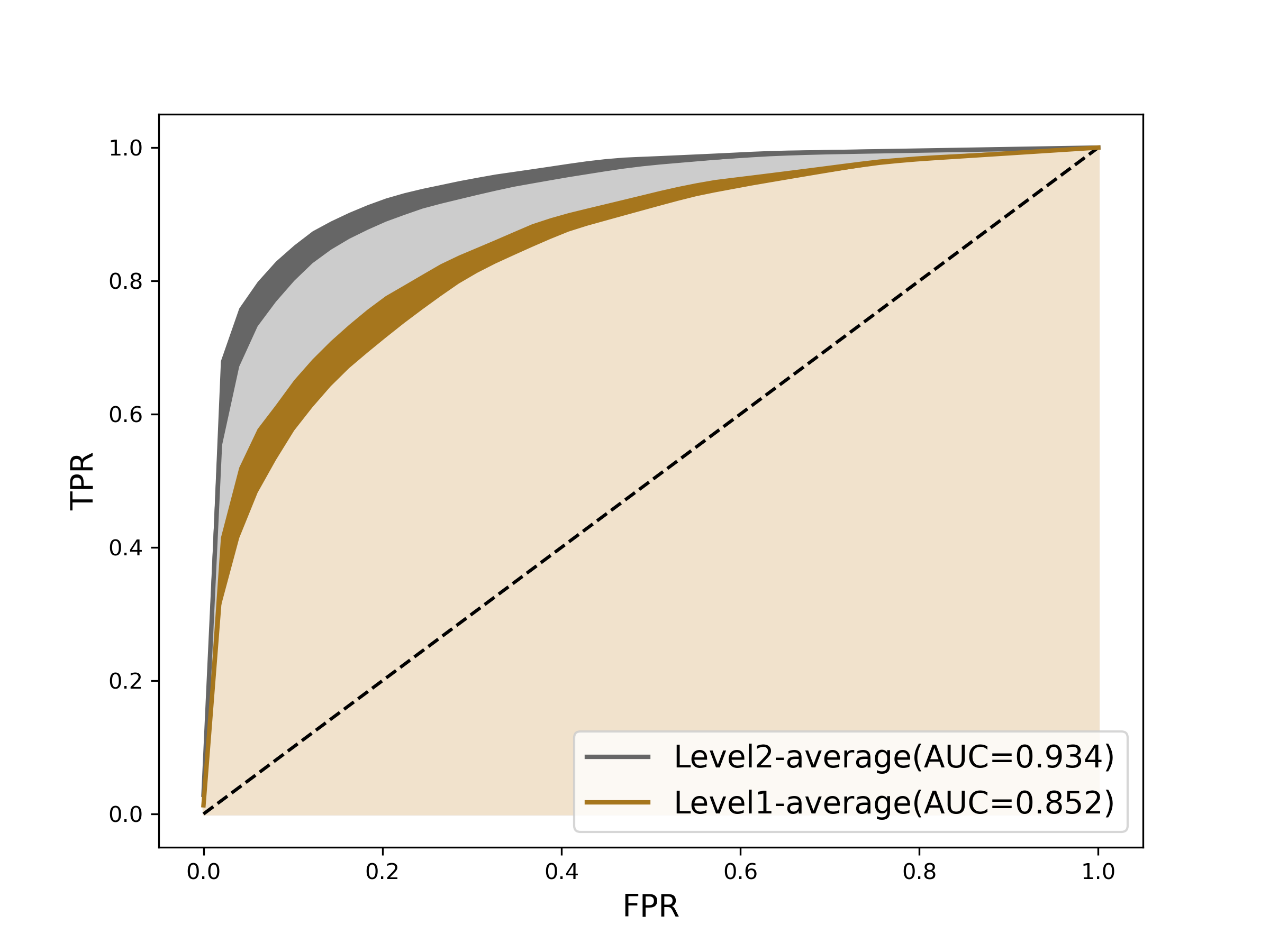}%
        \hspace{-0.2in}% 
        \includegraphics[width=2in]{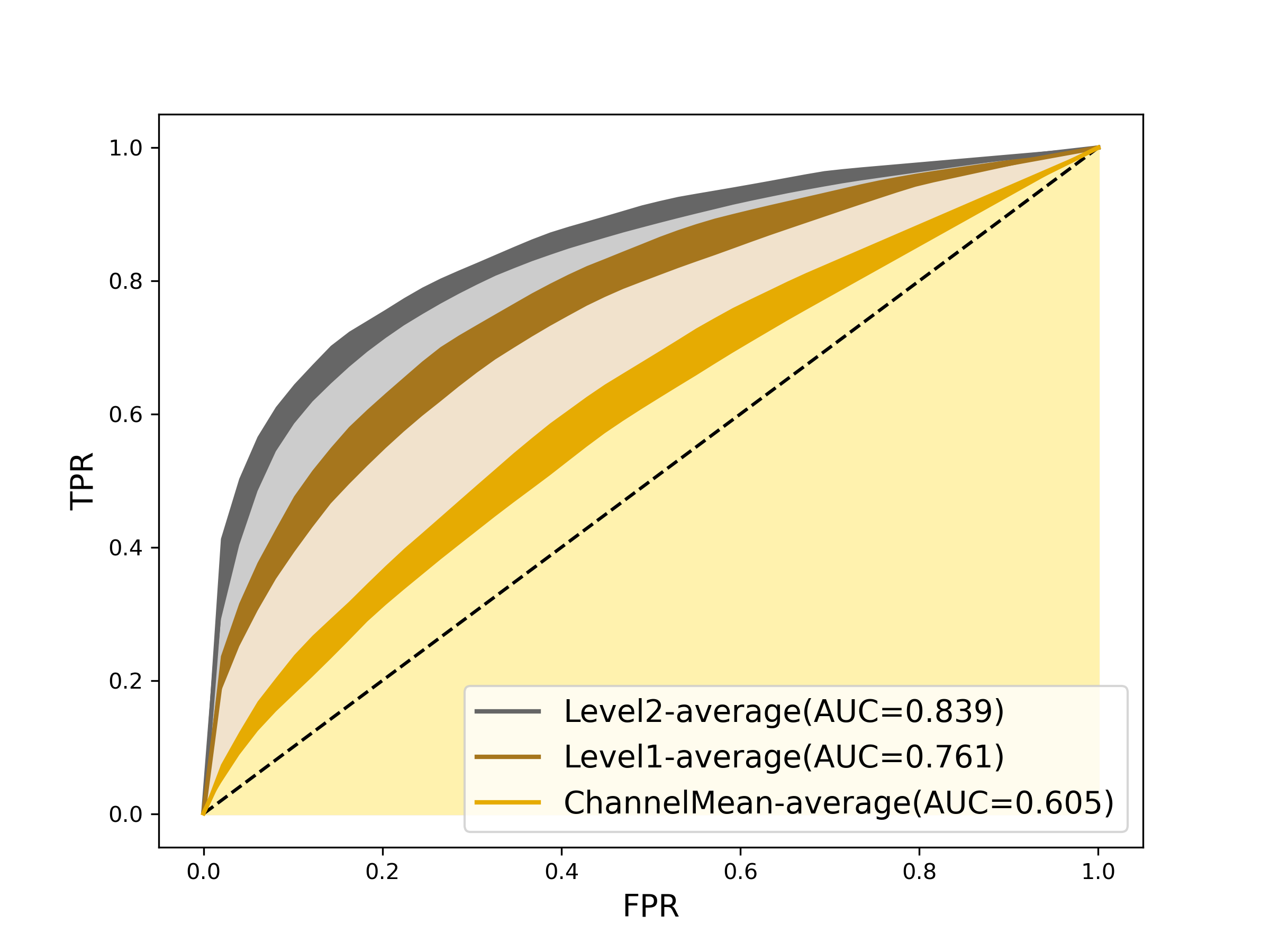}%
        \hspace{-0.2in}% 
        \includegraphics[width=2in]{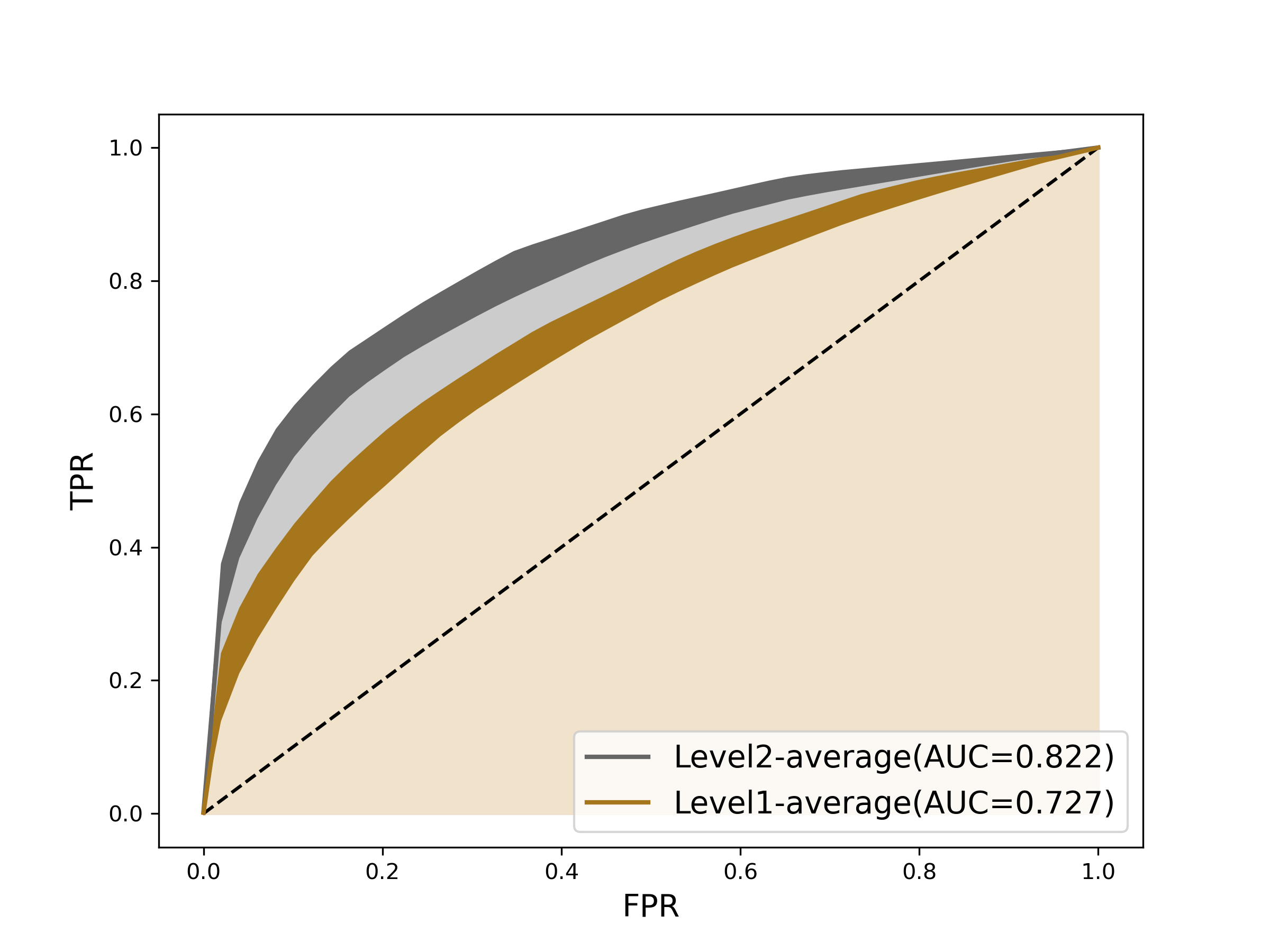}%
    }
    \caption{AUROCs of detecting targeted IFGSM attack from RepVGG-A2 with noise levels $\varepsilon=0.03$ and $\varepsilon=0.02$. From left to right: \text{2DSig-Norm} ($\varepsilon=0.03$), \text{2DSig-Conf} ($\varepsilon=0.03$), \text{2DSig-Norm} ($\varepsilon=0.02$), \text{2DSig-Conf} ($\varepsilon=0.02$).}
    \label{fig:repvgga2_resnet20_auc_ifgsm_target}  
\end{figure}

\begin{figure}[!t]
     \centering
     \makebox[\textwidth][c]{
   \includegraphics[width=2in]{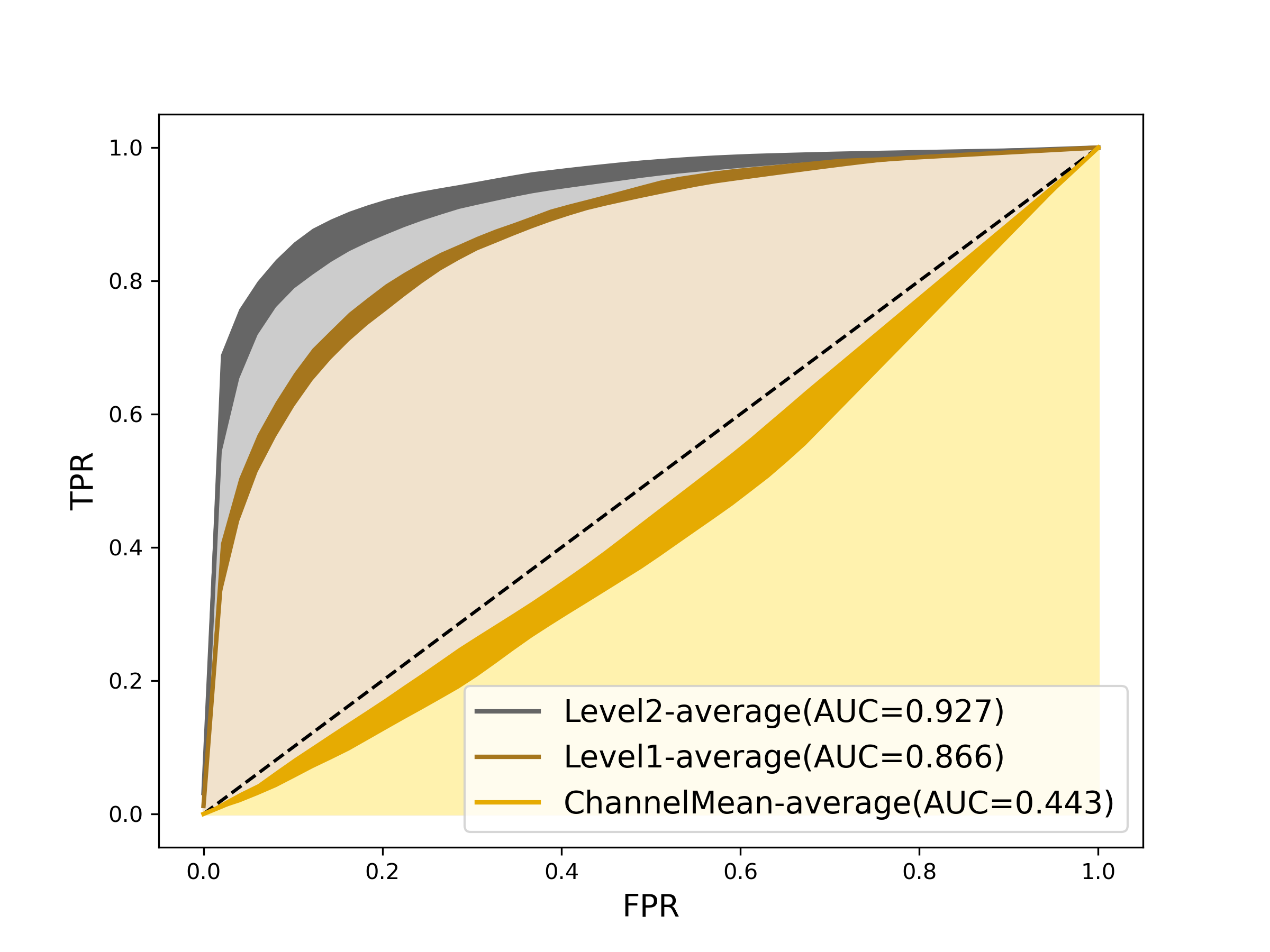} 
   \hspace{-0.2in}% 
  \includegraphics[width=2in]{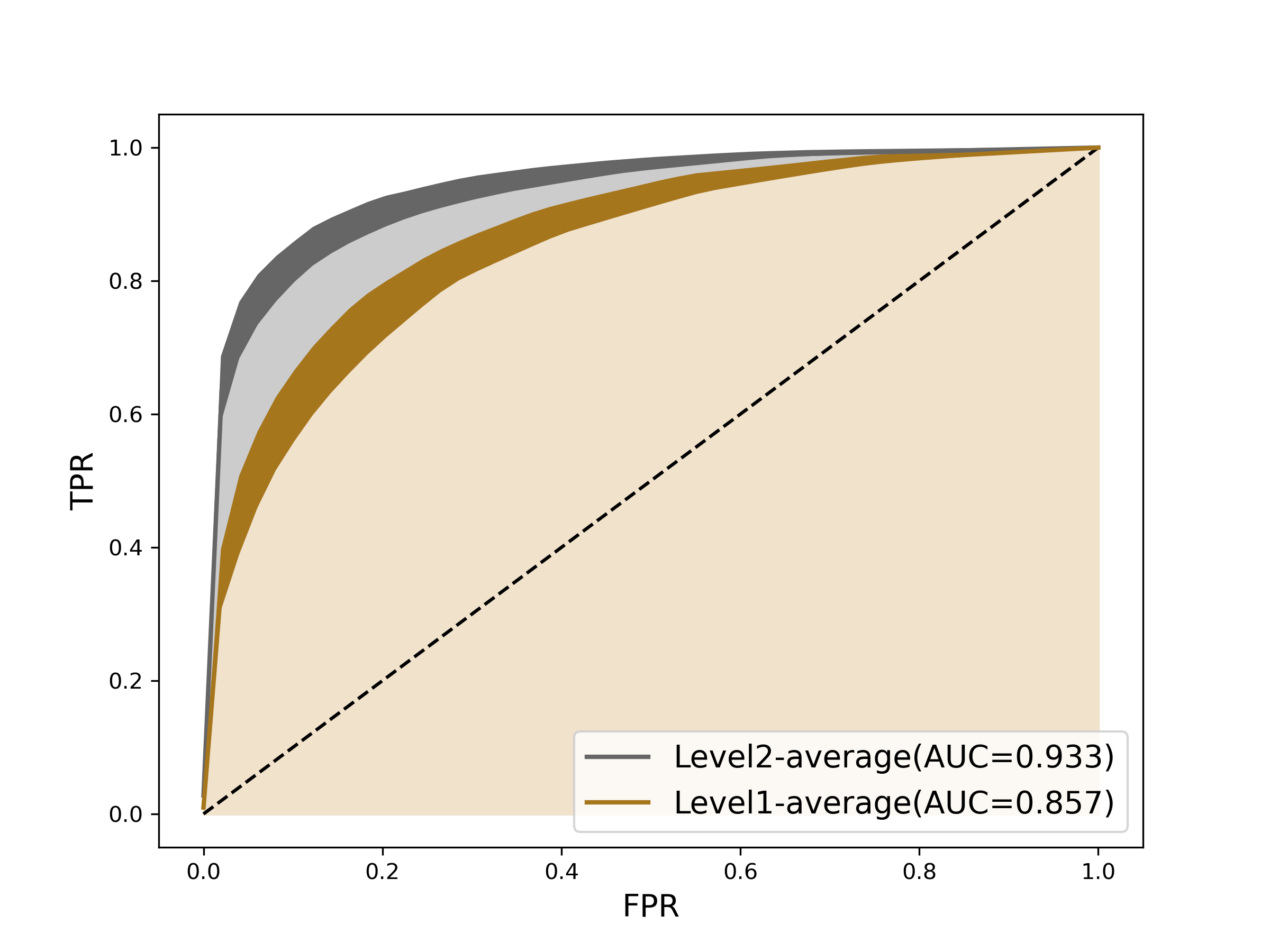}
  \hspace{-0.2in}% 
     \includegraphics[width=2in]{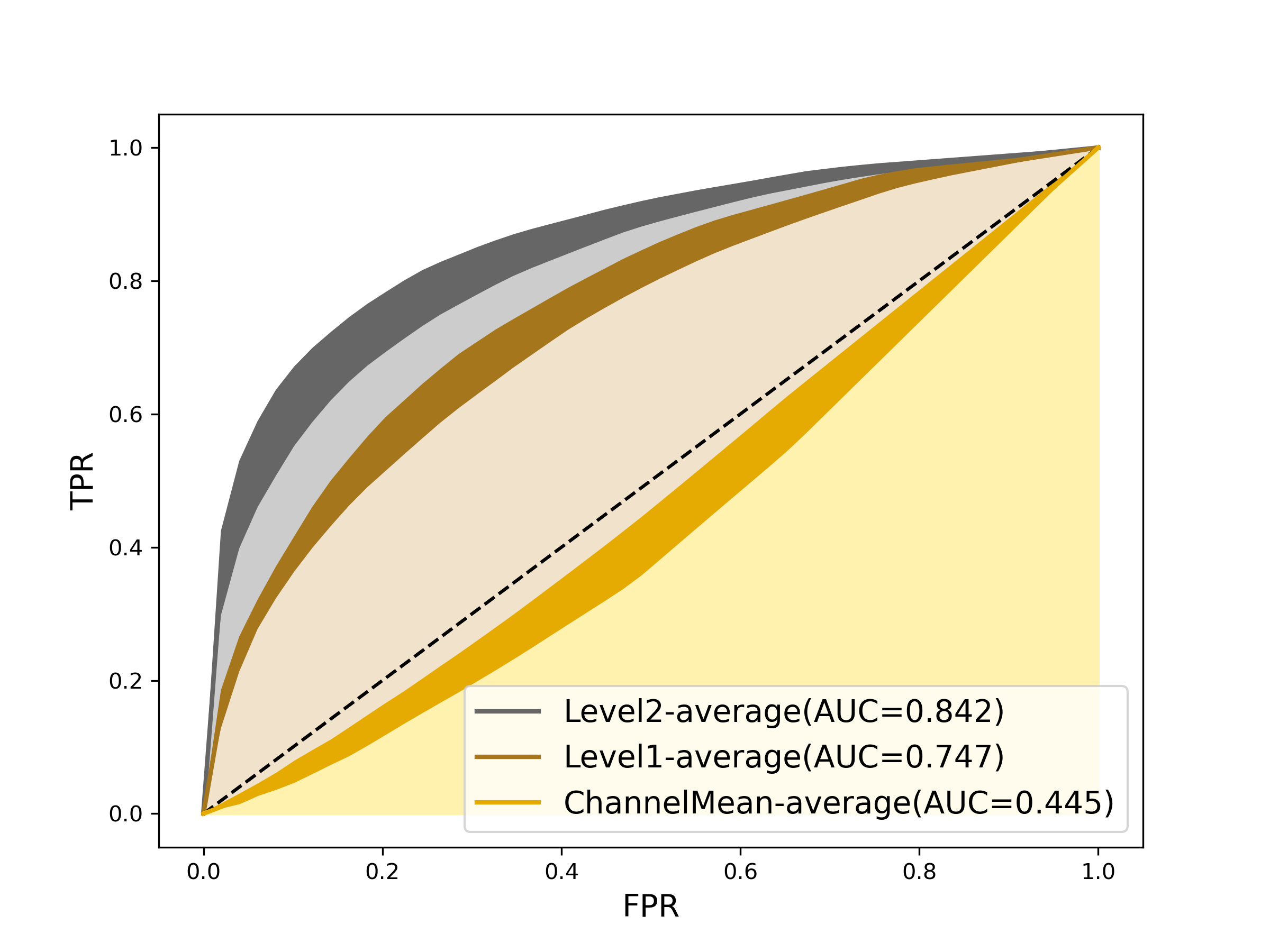} 
     \hspace{-0.2in}% 
  \includegraphics[width=2in]{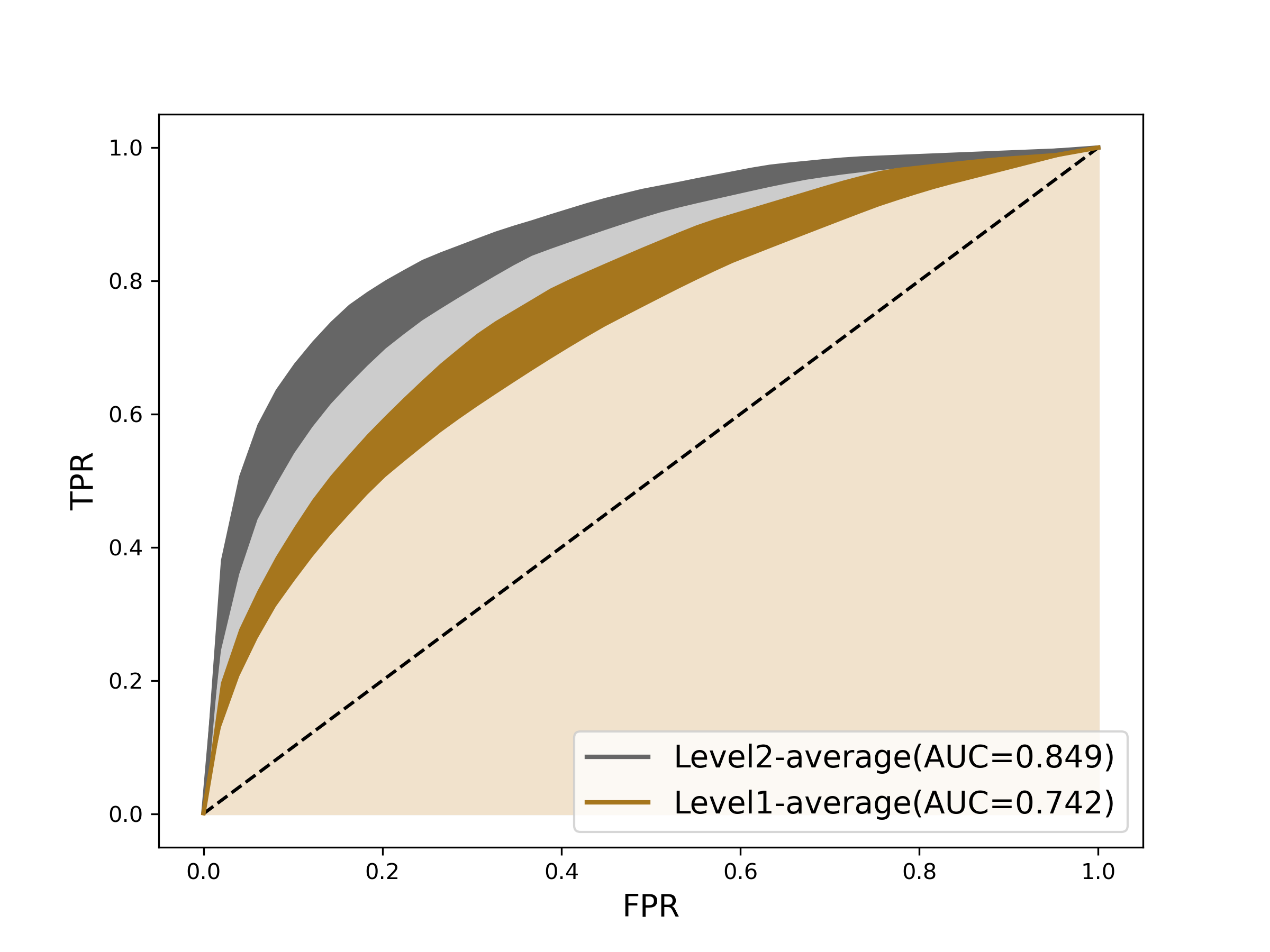}}
    \caption{AUROCs of detecting untargeted IFGSM attack from RepVGG-A2 with noise levels $\varepsilon=0.03$ and $\varepsilon=0.02$. From left to right: \text{2DSig-Norm} ($\varepsilon=0.03$), \text{2DSig-Conf} ($\varepsilon=0.03$), \text{2DSig-Norm} ($\varepsilon=0.02$), \text{2DSig-Conf} ($\varepsilon=0.02$).}
    \label{fig:resnet20_resnet20_auc_ifgsm_untarget}  
\end{figure}

\begin{figure}[!t]
    \centering
    \makebox[\textwidth][c]{%
   \includegraphics[width=2in]{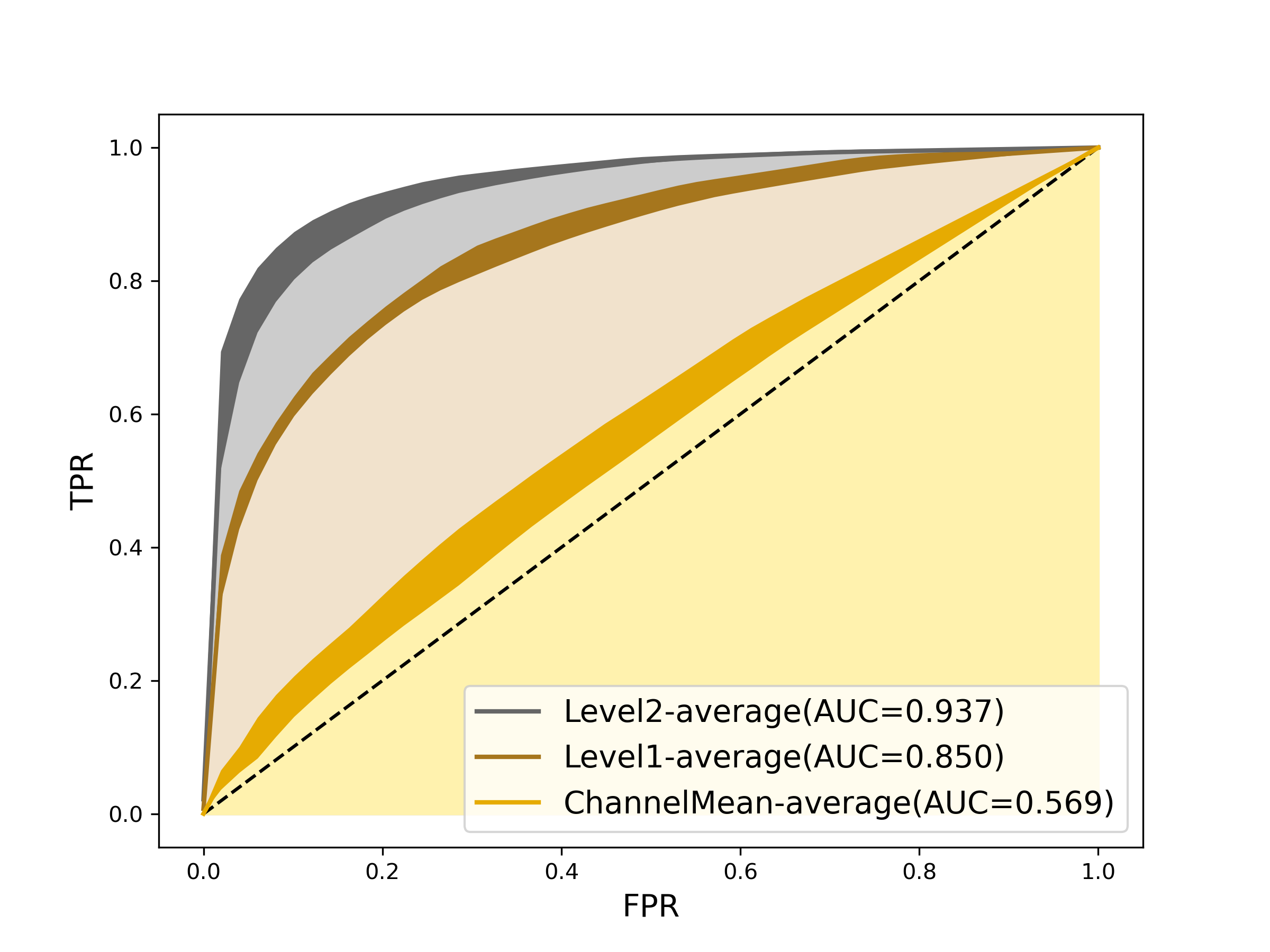} 
   \hspace{-0.2in}% 
  \includegraphics[width=2in]{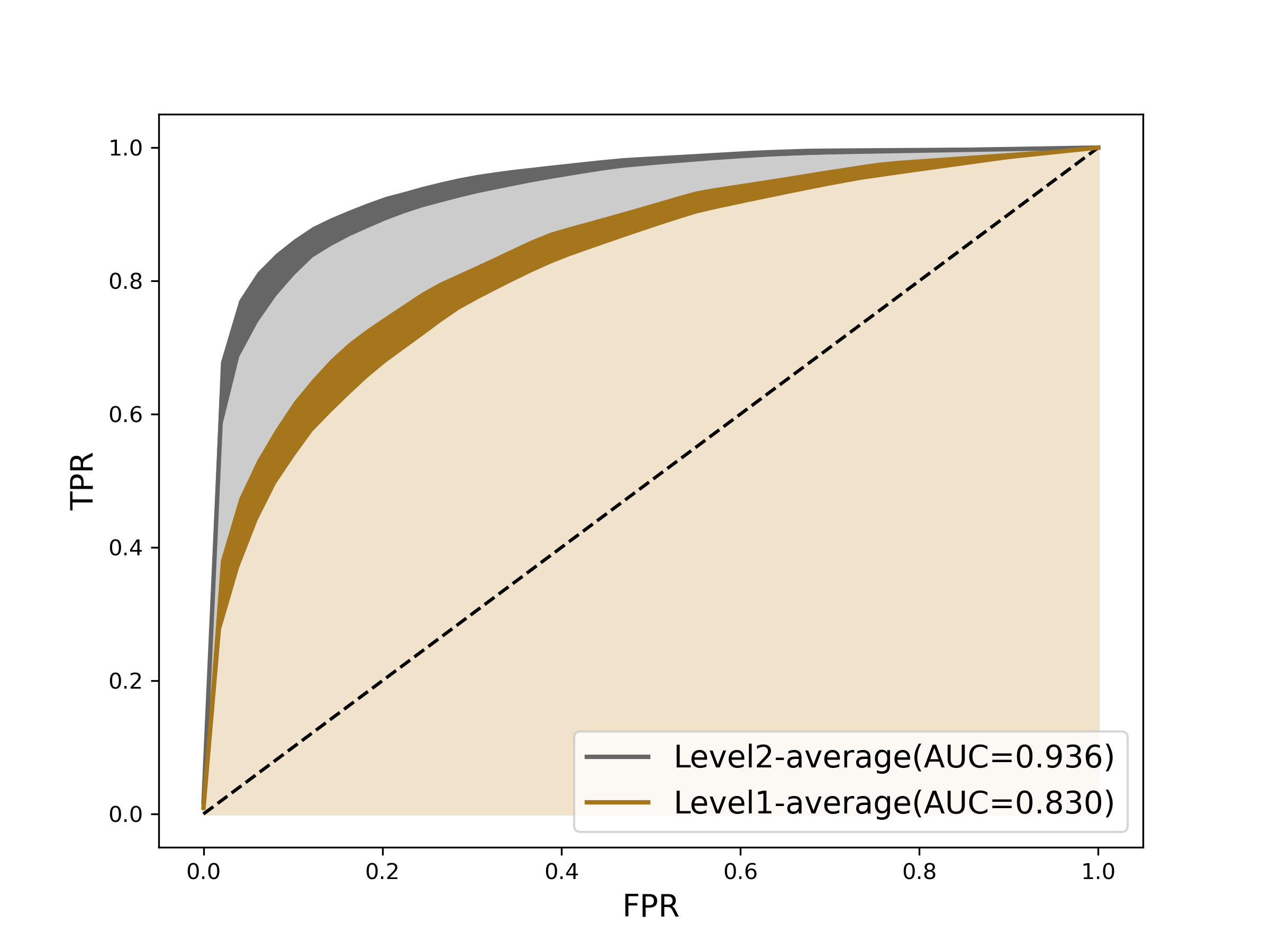} 
  \hspace{-0.2in}% 
     \includegraphics[width=2in]{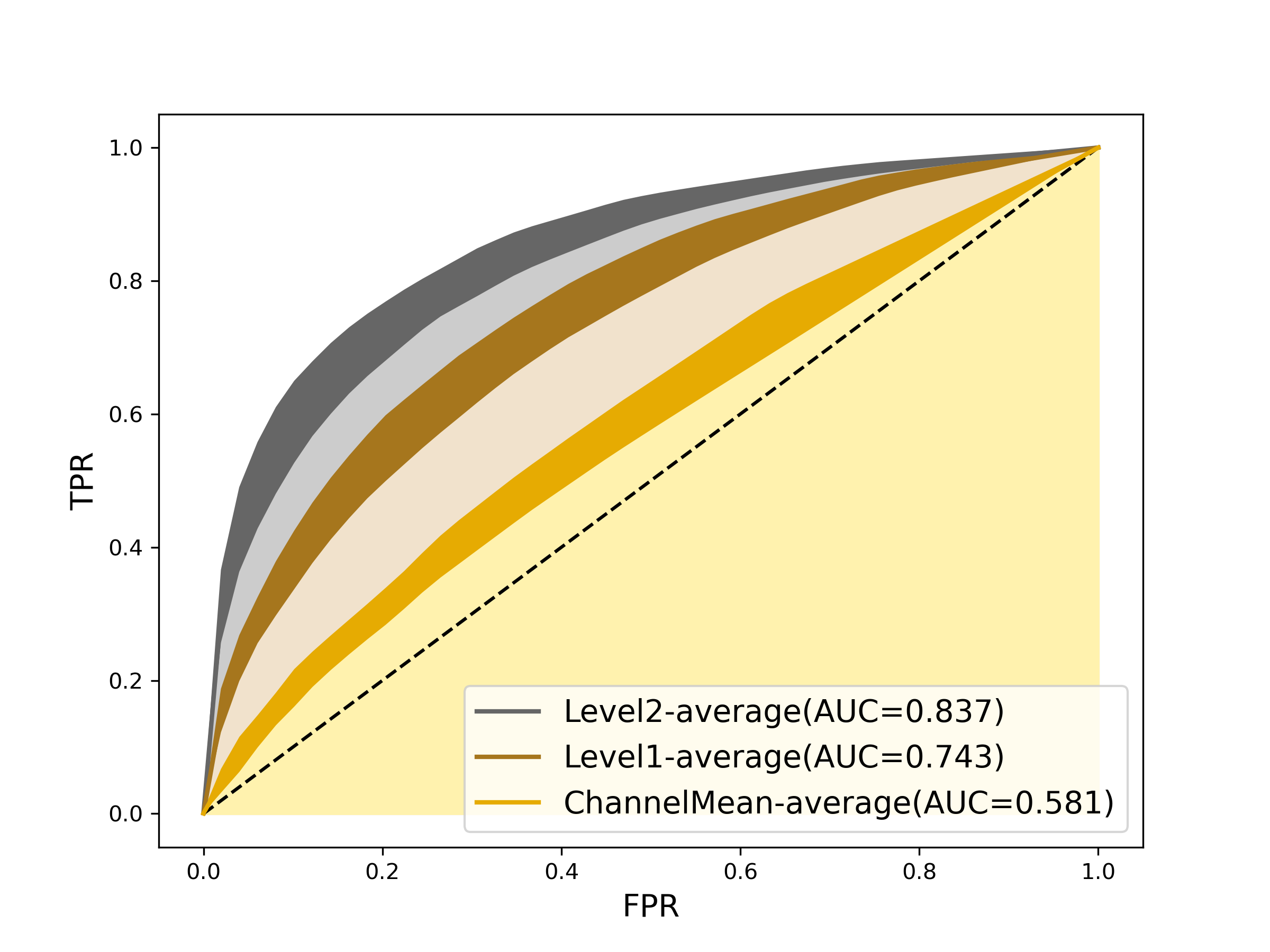} 
     \hspace{-0.2in}% 
  \includegraphics[width=2in]{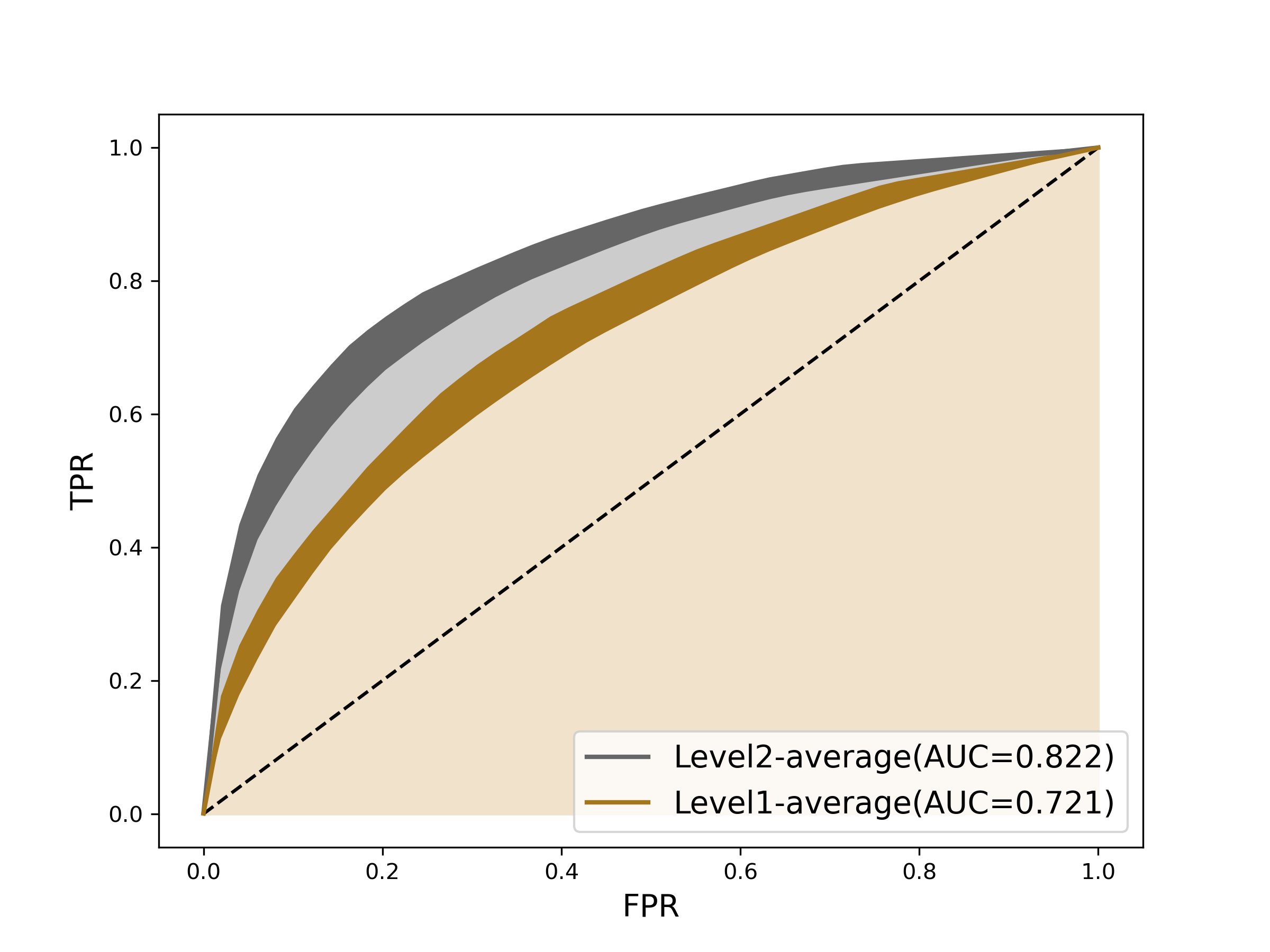}}
    \caption{AUROCs of detecting targeted IFGSM attack from ResNet-20 with noise levels $\varepsilon=0.03$ and $\varepsilon=0.02$. From left to right: \text{2DSig-Norm} ($\varepsilon=0.03$), \text{2DSig-Conf} ($\varepsilon=0.03$), \text{2DSig-Norm} ($\varepsilon=0.02$), \text{2DSig-Conf} ($\varepsilon=0.02$).}
    \label{fig:resnet20_resnet20_auc_ifgsm_target}  
\end{figure}

\begin{figure}[!t]
    \centering
    \makebox[\textwidth][c]{%
          \includegraphics[width=2in]{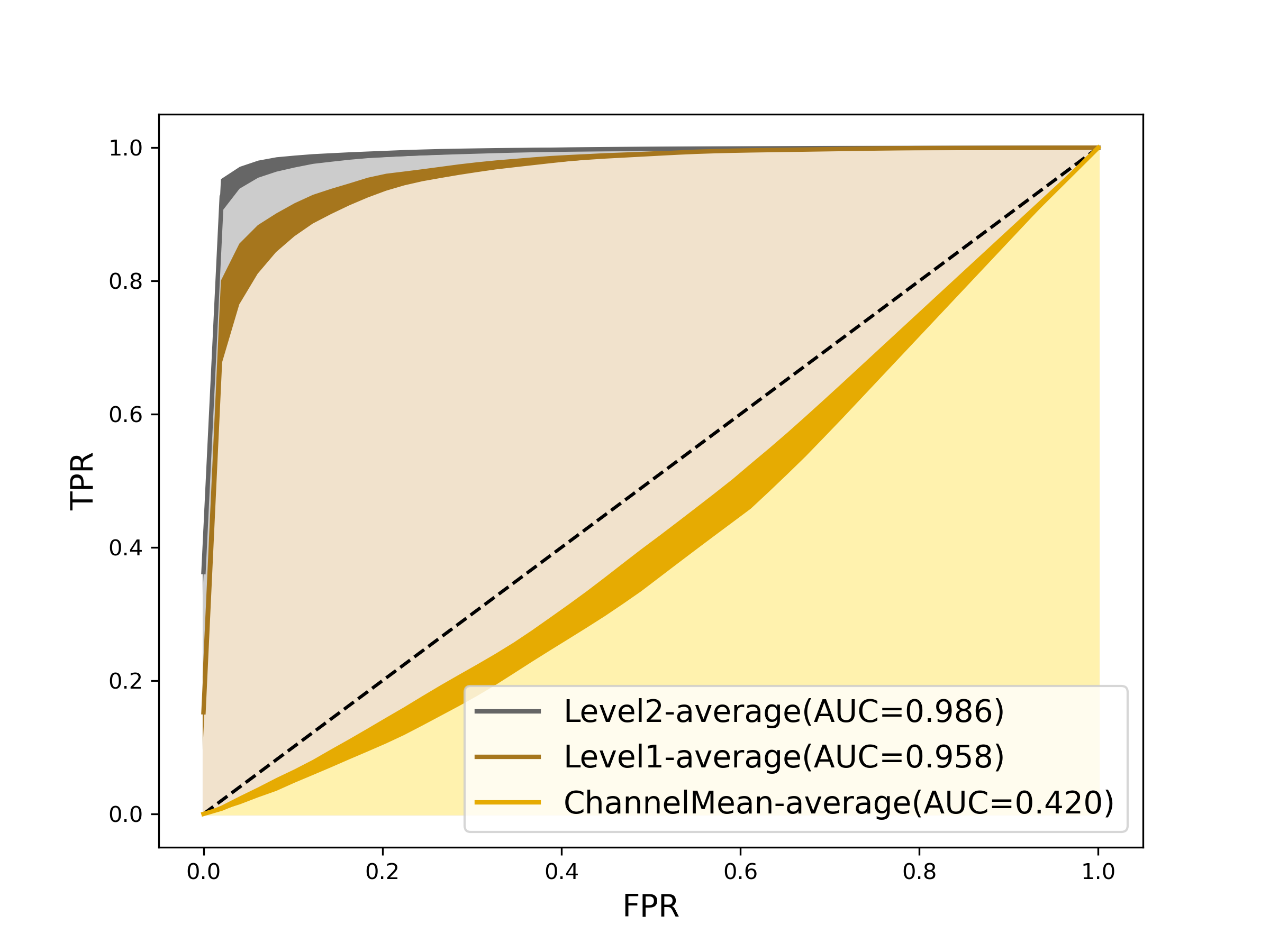}    
          \hspace{-0.2in}% 
                    \includegraphics[width=2in]{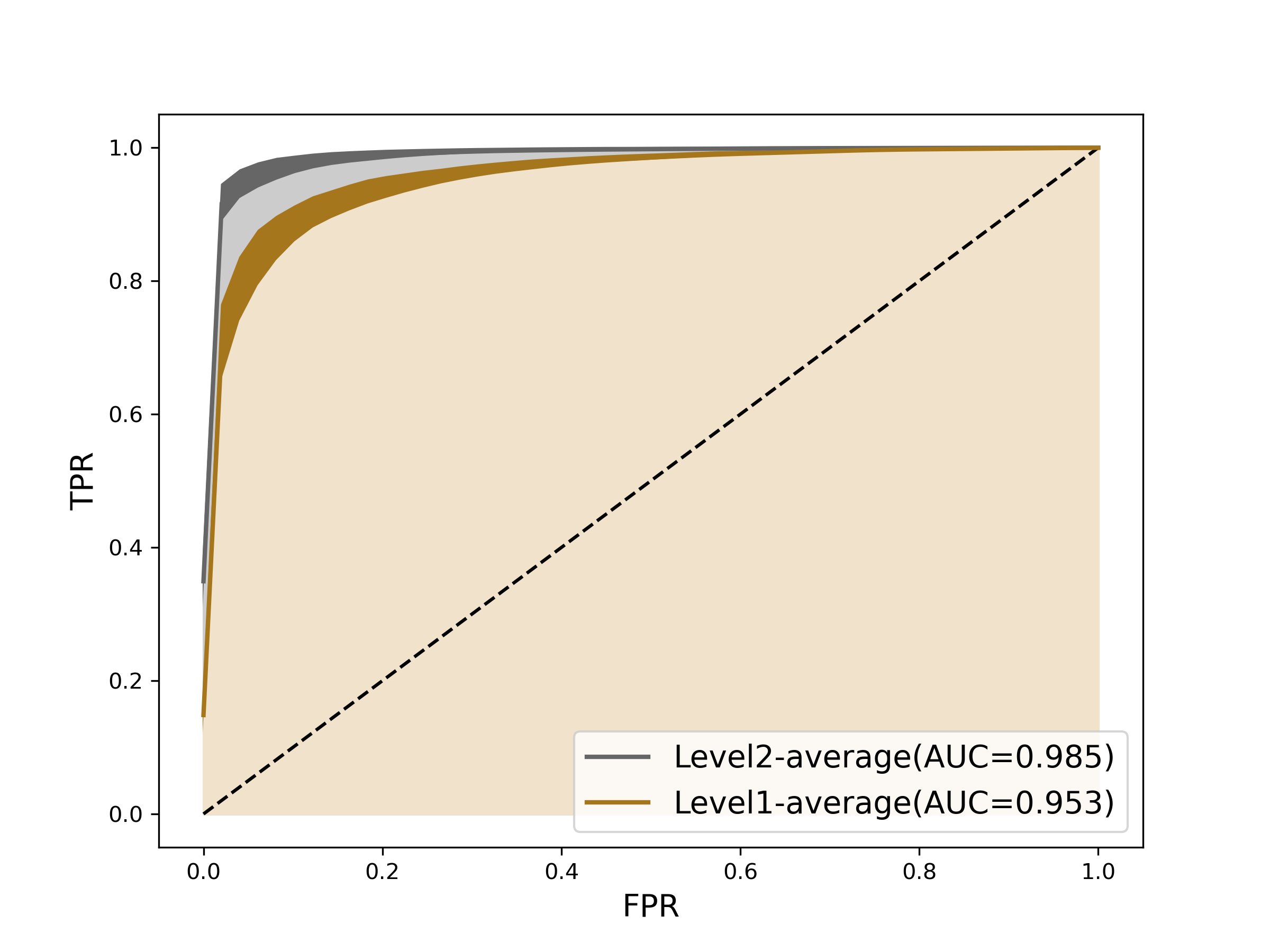}   
                    \hspace{-0.2in}% 
                              \includegraphics[width=2in]{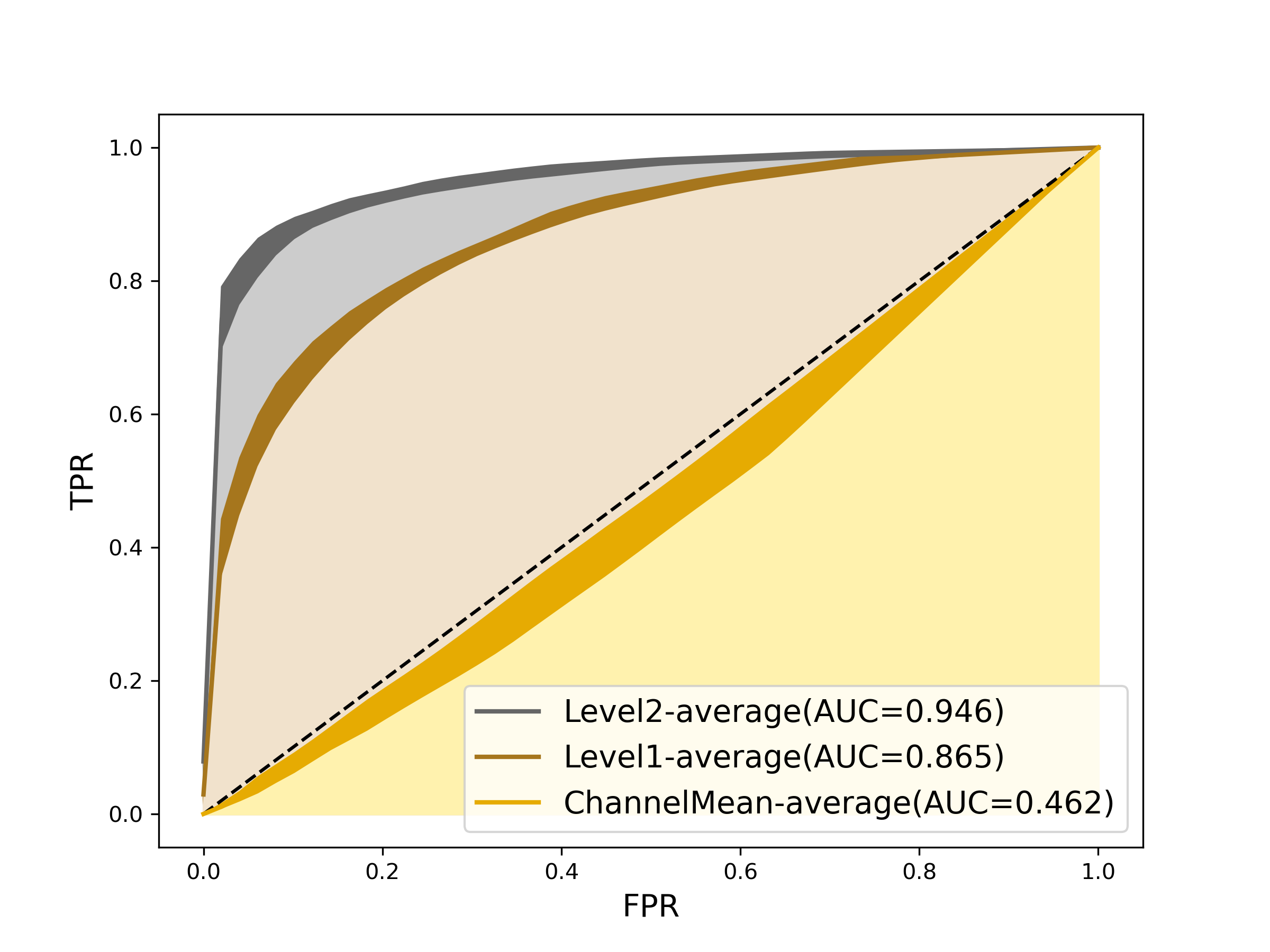}    
                              \hspace{-0.2in}% 
                    \includegraphics[width=2in]{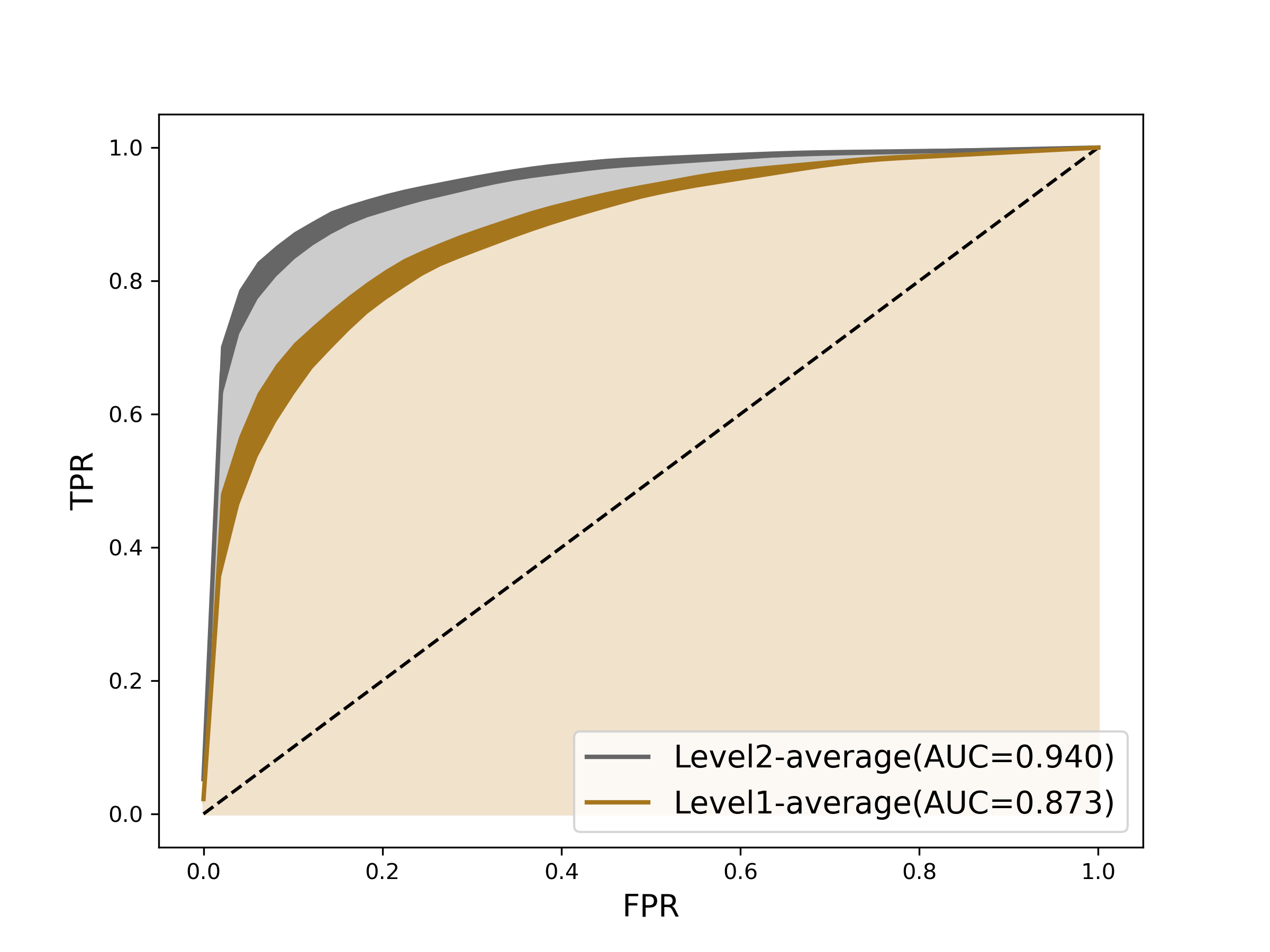}   }
    \caption{AUROCs of detecting untargeted PGDM attack from RepVGG-A2 with noise levels $\varepsilon=0.03$ and $\varepsilon=0.02$. From left to right: \text{2DSig-Norm} ($\varepsilon=0.03$), \text{2DSig-Conf} ($\varepsilon=0.03$), \text{2DSig-Norm} ($\varepsilon=0.02$), \text{2DSig-Conf} ($\varepsilon=0.02$).}
    \label{fig:repvgga2_resnet20_auc_pgdm_untarget}  
\end{figure}

\begin{figure}[!t]
    \centering
    \makebox[\textwidth][c]{%
    \includegraphics[width=2in]{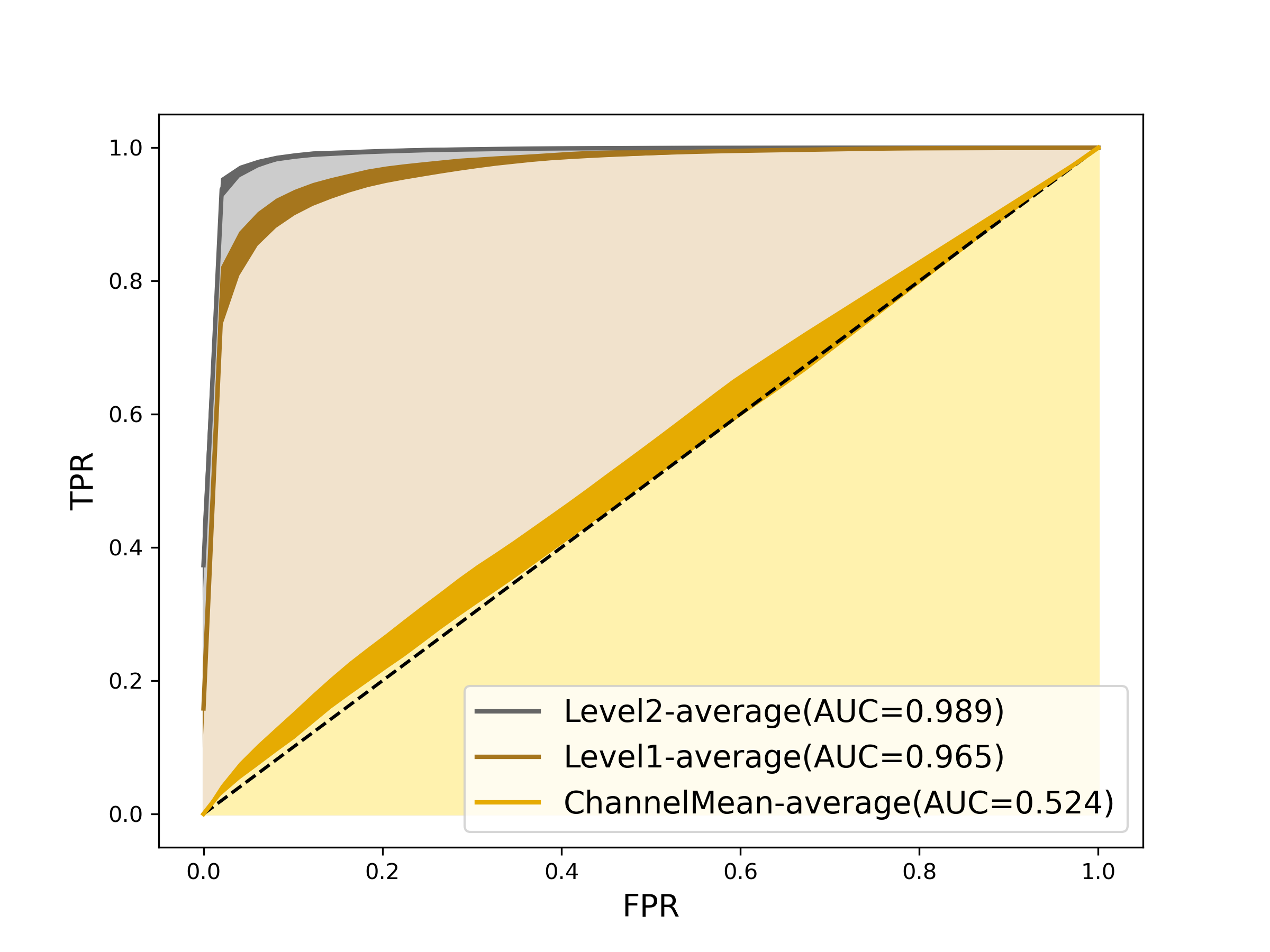}  
    \hspace{-0.2in}% 
    \includegraphics[width=2in]{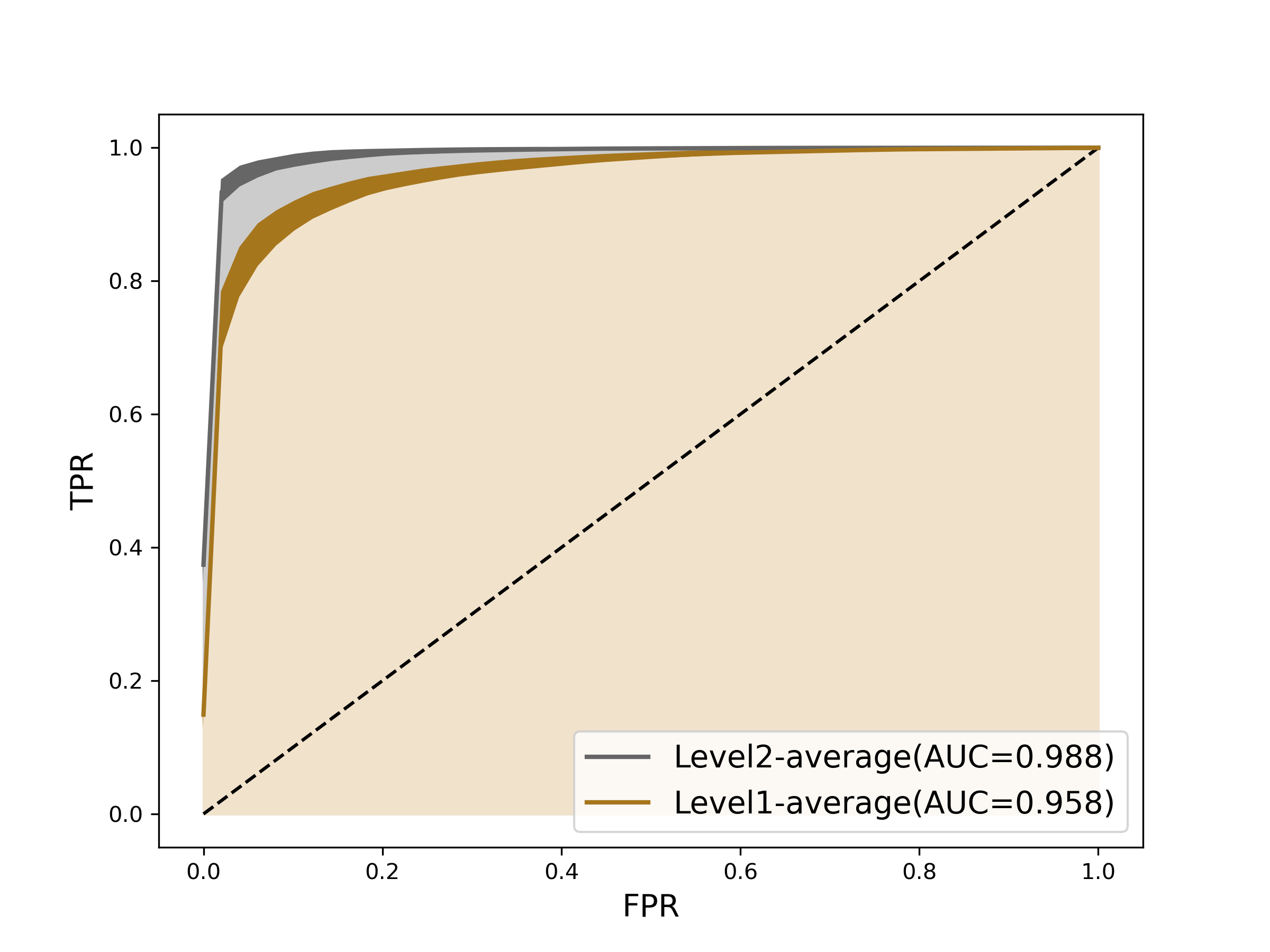}
    \hspace{-0.2in}% 
      \includegraphics[width=2in]{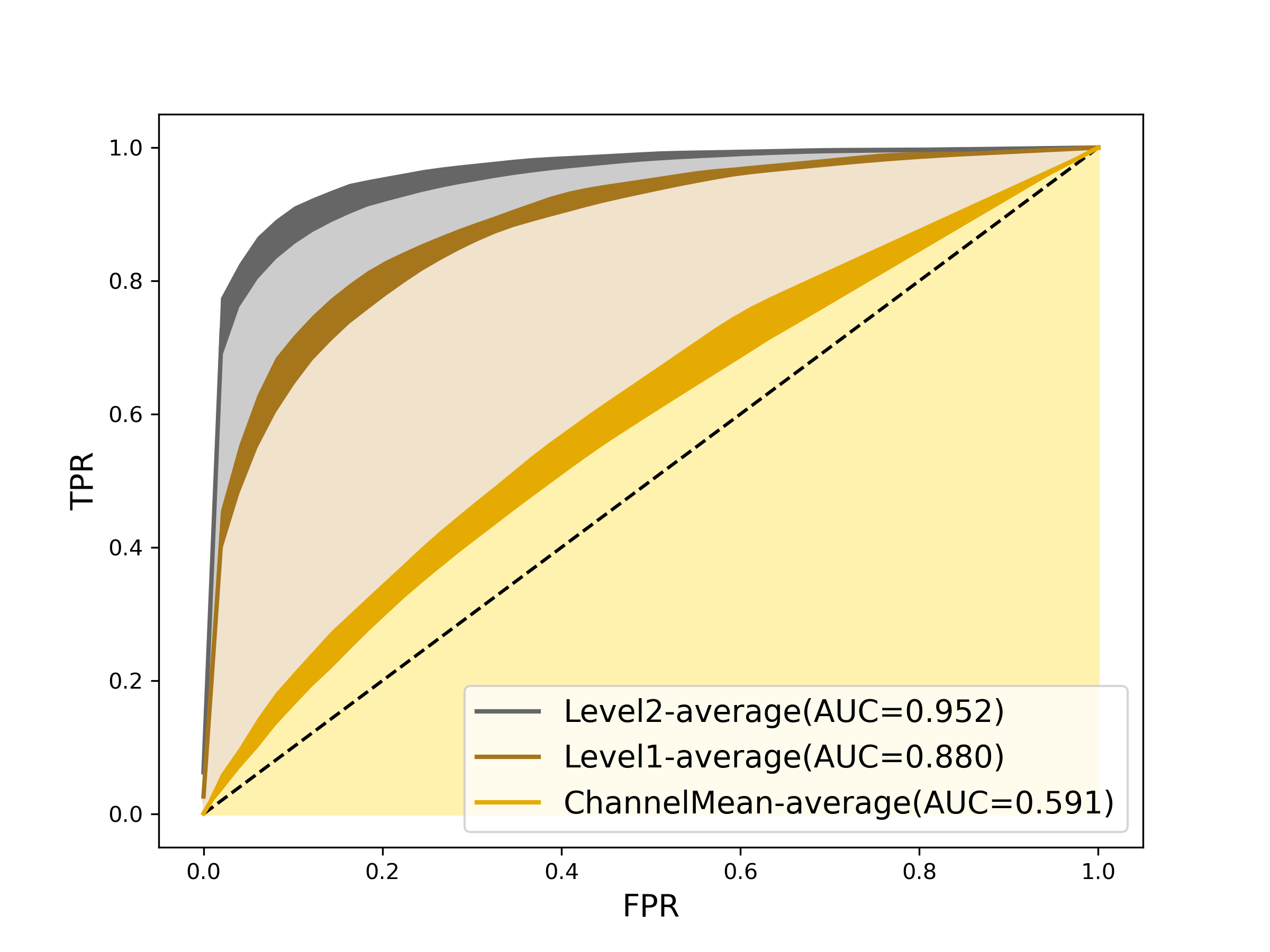} 
      \hspace{-0.2in}% 
      \includegraphics[width=2in]{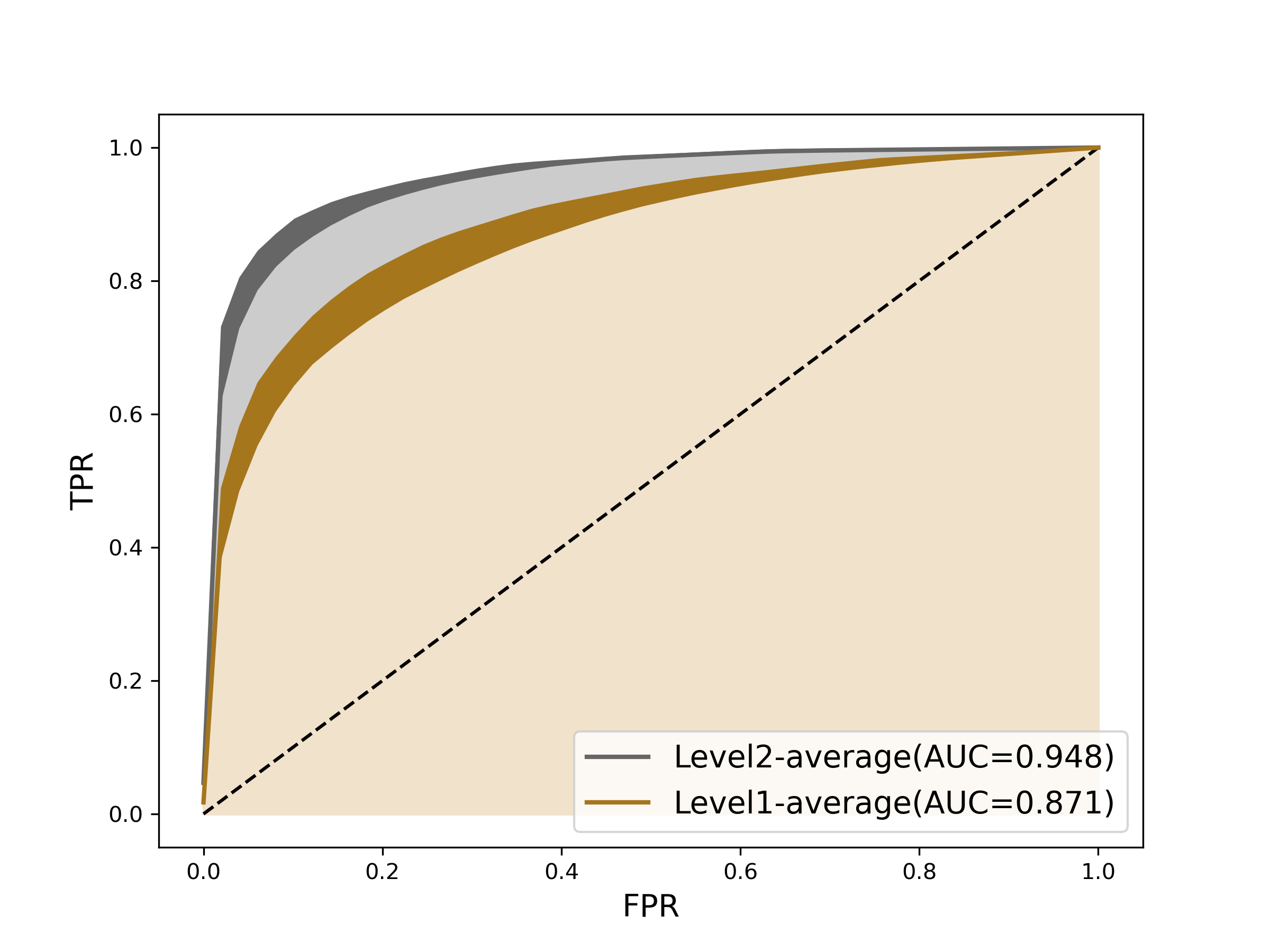} }
    \caption{AUROCs of detecting targeted PGDM attack from RepVGG-A2 with noise levels $\varepsilon=0.03$ and $\varepsilon=0.02$. From left to right: \text{2DSig-Norm} ($\varepsilon=0.03$), \text{2DSig-Conf} ($\varepsilon=0.03$), \text{2DSig-Norm} ($\varepsilon=0.02$), \text{2DSig-Conf} ($\varepsilon=0.02$).}
    \label{fig:repvgga2_resnet20_auc_pgdm_target}  
\end{figure}

\begin{figure}[!t]
    \centering
    \makebox[\textwidth][c]{%
   \includegraphics[width=2in]{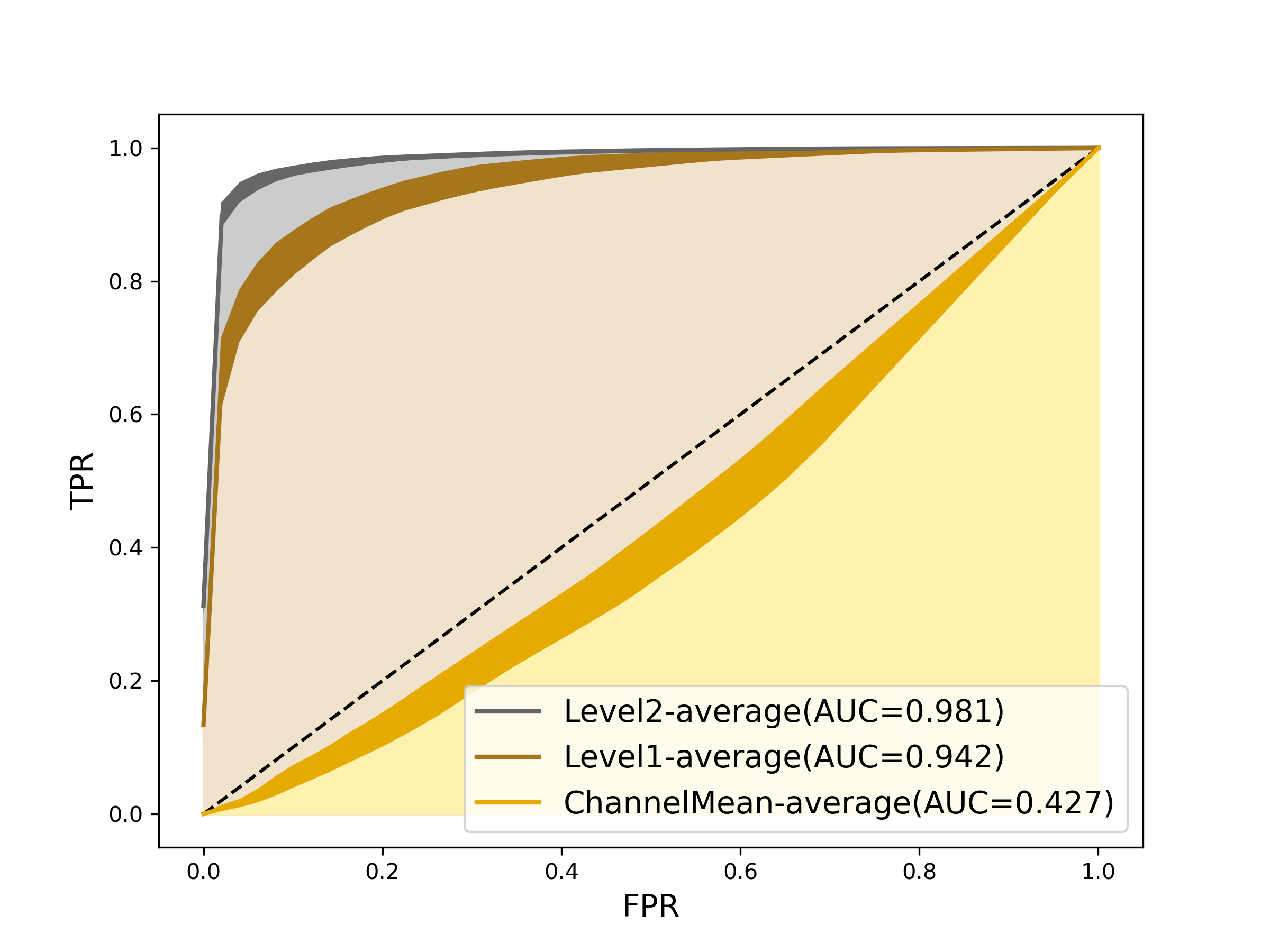}
   \hspace{-0.2in}% 
  \includegraphics[width=2in]{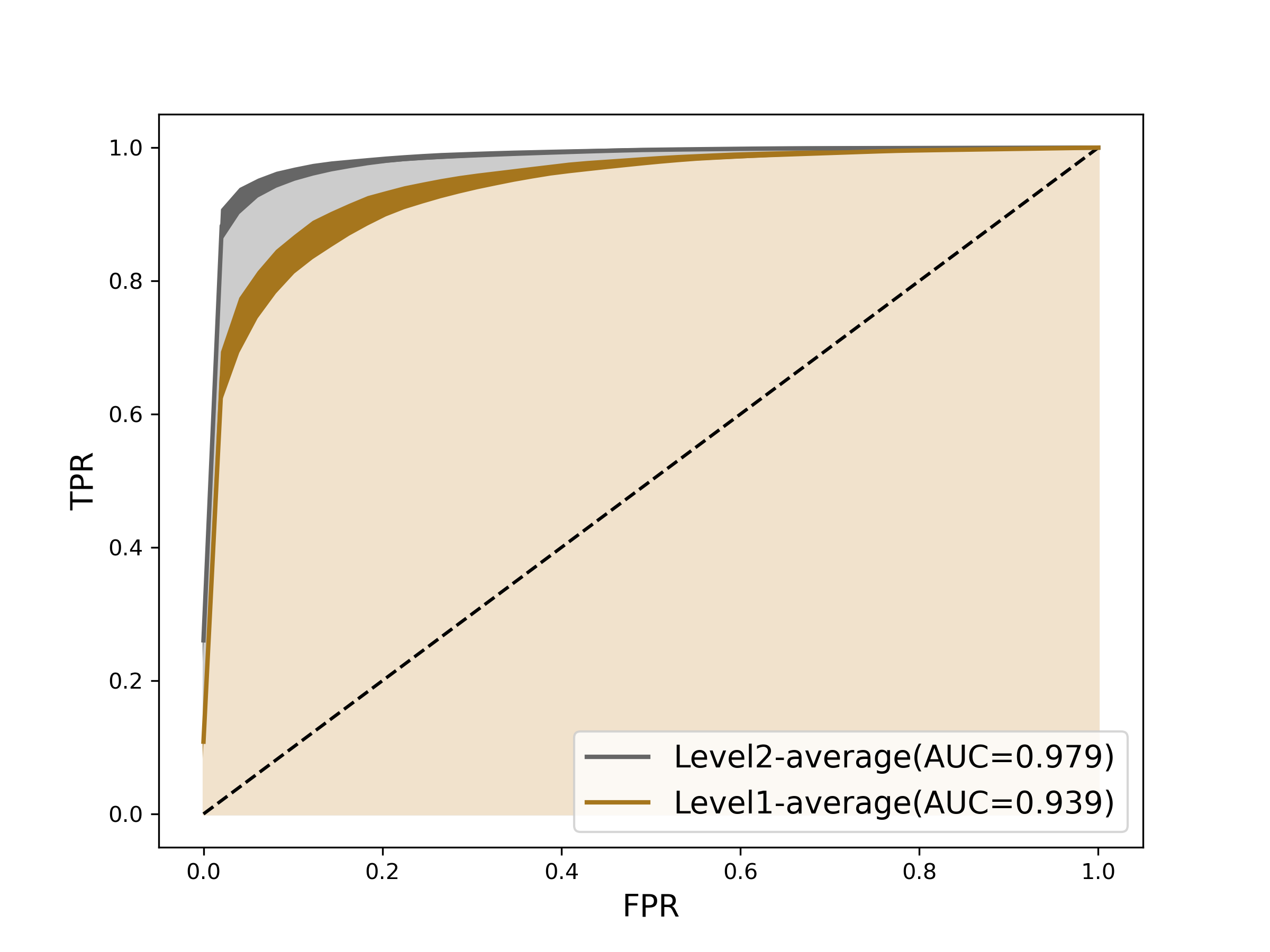} 
  \hspace{-0.2in}% 
     \includegraphics[width=2in]{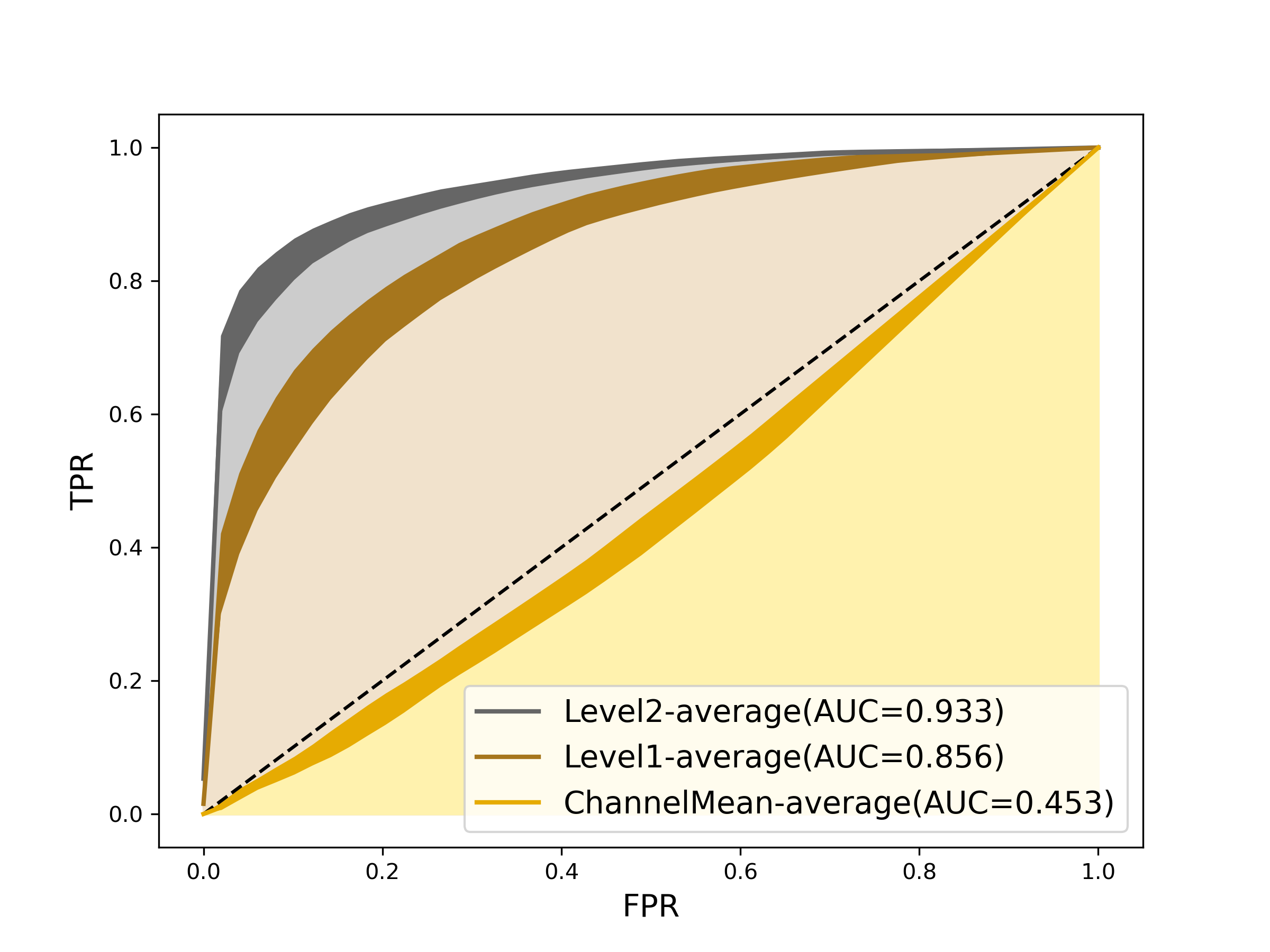} 
     \hspace{-0.2in}% 
  \includegraphics[width=2in]{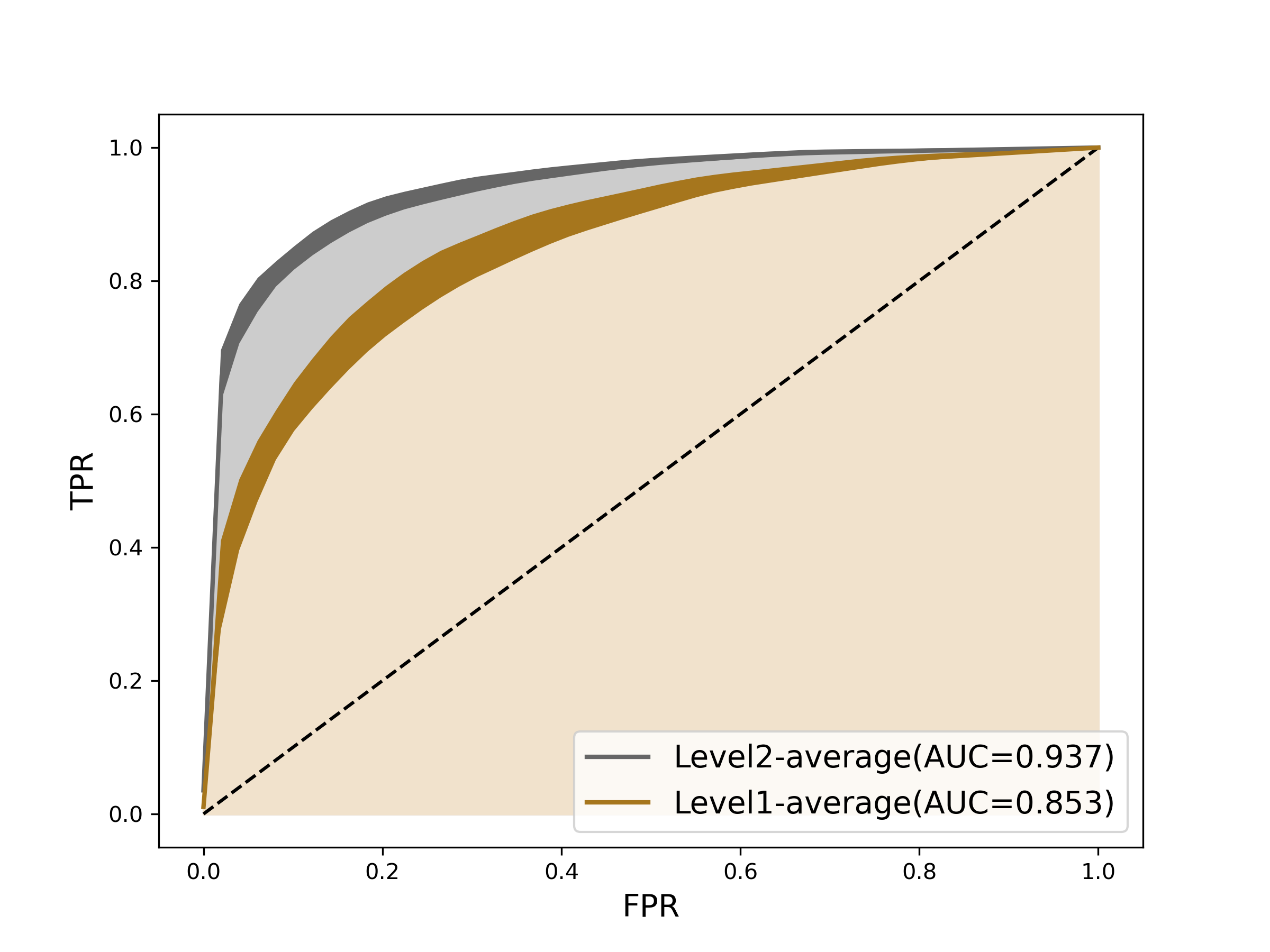} }
    \caption{AUROCs of detecting untargeted PGDM attack from ResNet-20 with noise levels $\varepsilon=0.03$ and $\varepsilon=0.02$. From left to right: \text{2DSig-Norm} ($\varepsilon=0.03$), \text{2DSig-Conf} ($\varepsilon=0.03$), \text{2DSig-Norm} ($\varepsilon=0.02$), \text{2DSig-Conf} ($\varepsilon=0.02$).}
    \label{fig:resnet20_resnet20_auc_pgdm_untarget}  
\end{figure}

\begin{figure}[!t]
    \centering
    \makebox[\textwidth][c]{%
   \includegraphics[width=2in]{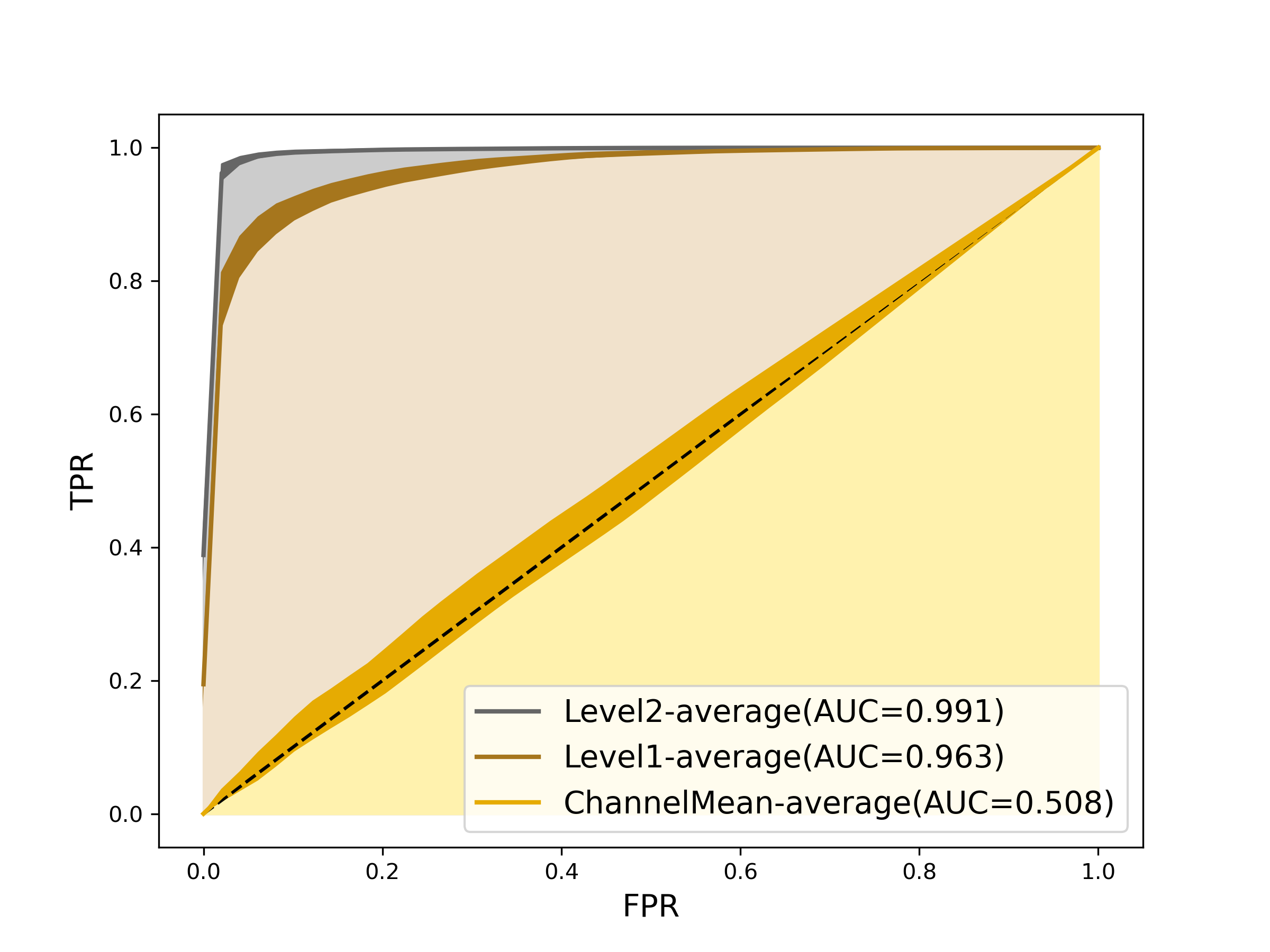} 
   \hspace{-0.2in}% 
  \includegraphics[width=2in]{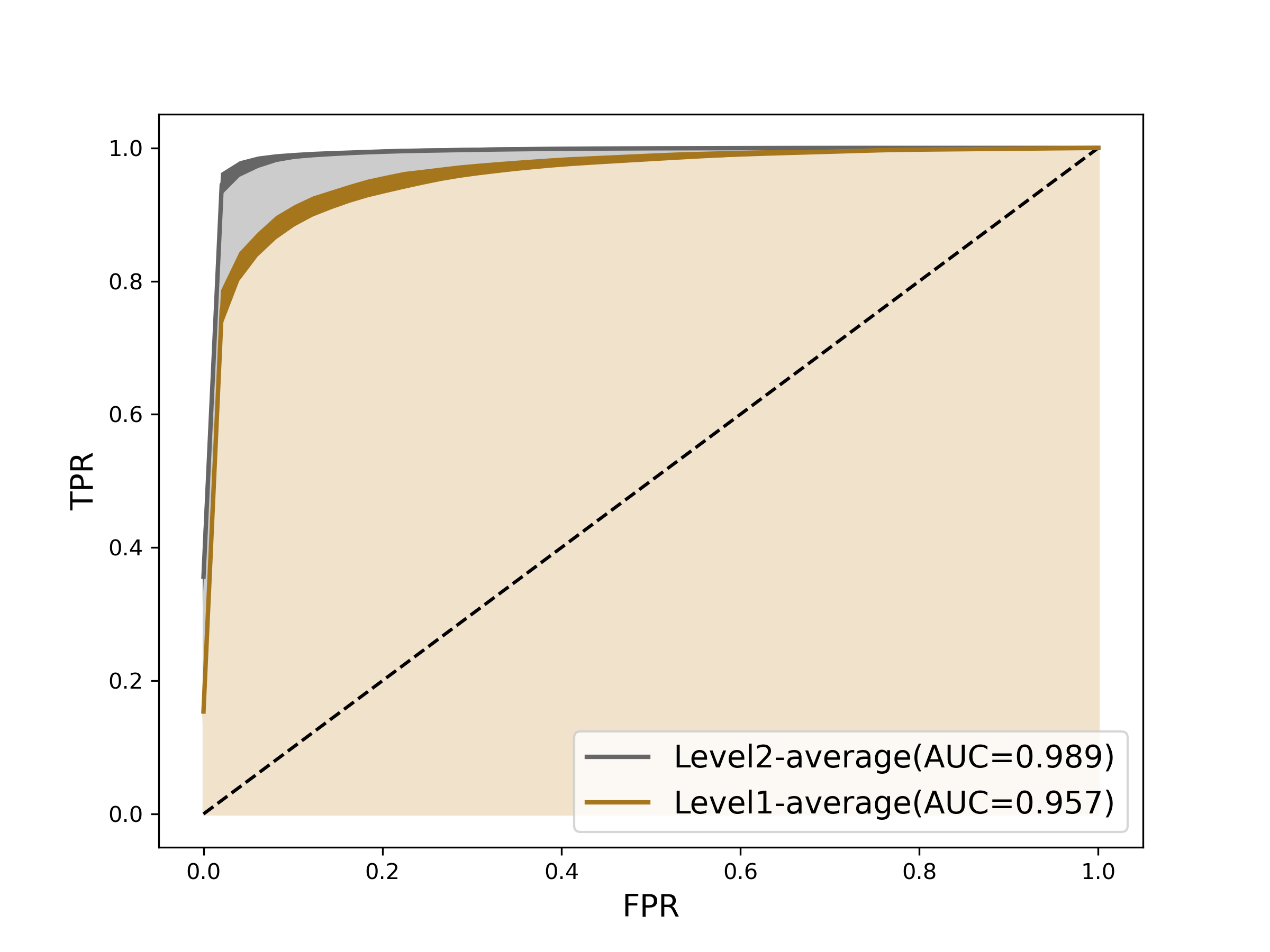} 
  \hspace{-0.2in}% 
     \includegraphics[width=2in]{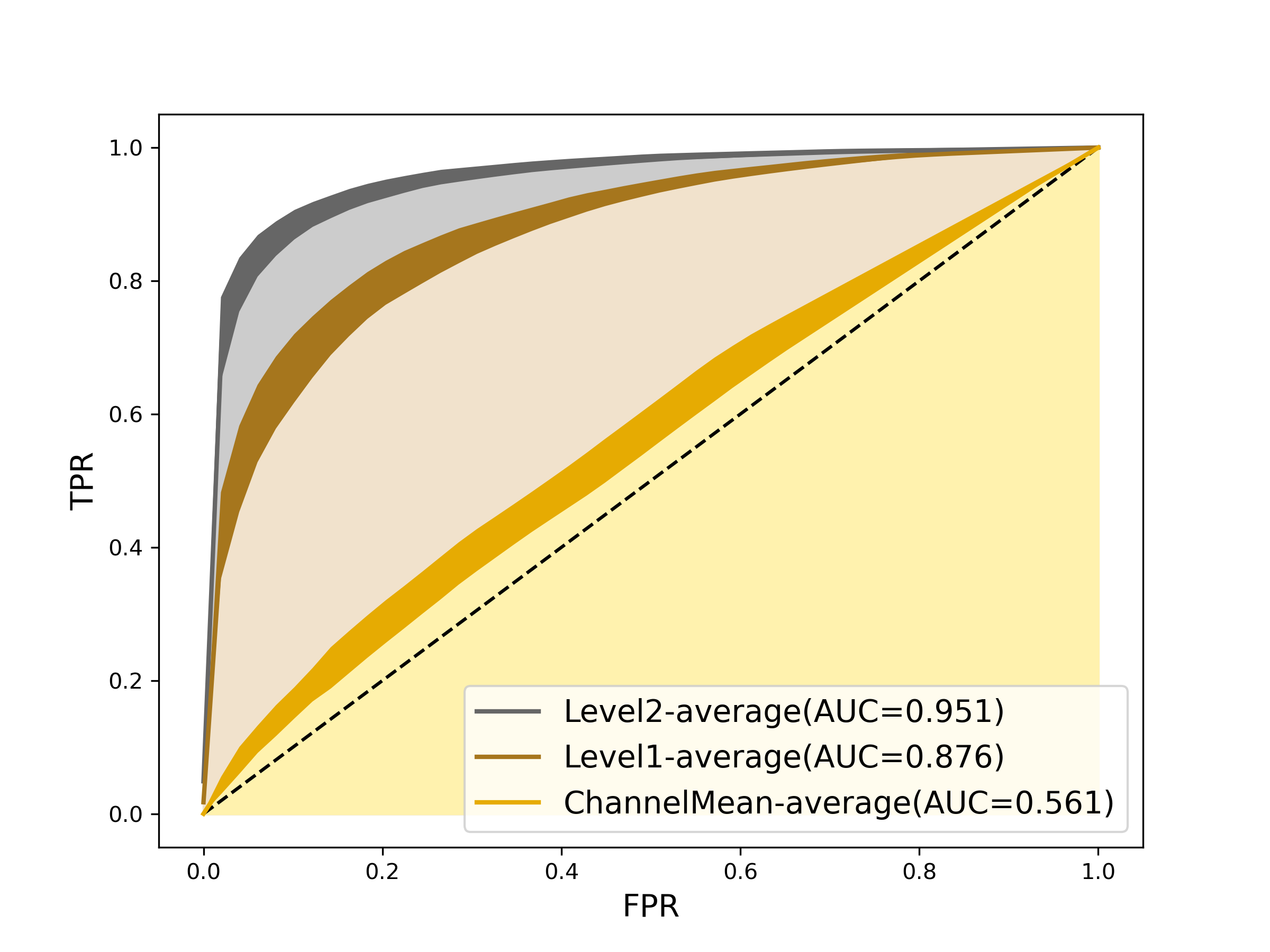} 
     \hspace{-0.2in}% 
  \includegraphics[width=2in]{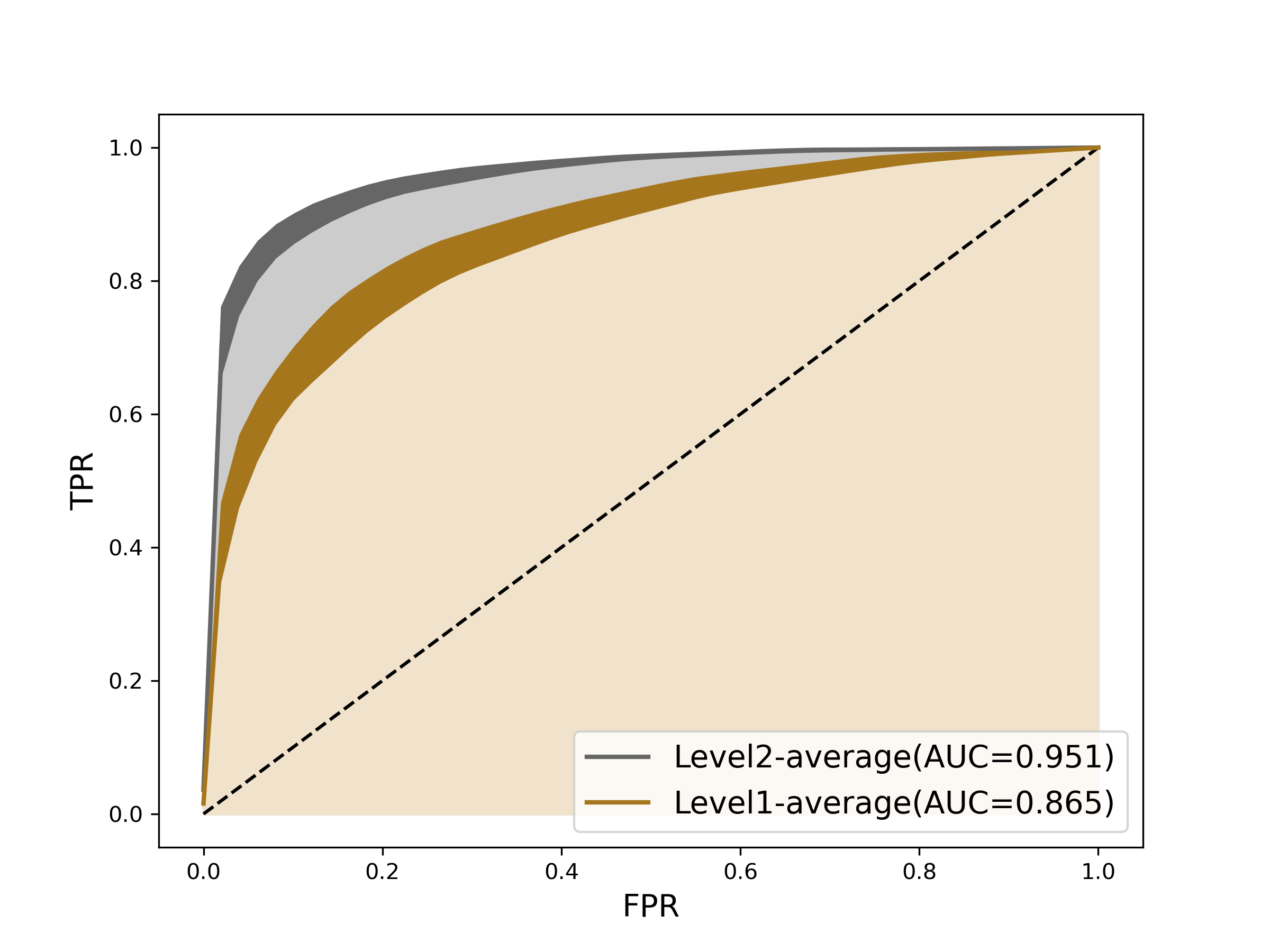} }
    \caption{AUROCs of detecting targeted PGDM attack from ResNet-20 with noise levels $\varepsilon=0.03$ and $\varepsilon=0.02$. From left to right: \text{2DSig-Norm} ($\varepsilon=0.03$), \text{2DSig-Conf} ($\varepsilon=0.03$), \text{2DSig-Norm} ($\varepsilon=0.02$), \text{2DSig-Conf} ($\varepsilon=0.02$).}
    \label{fig:resnet20_resnet20_auc_pgdm_target}  
\end{figure}

\begin{table}[]
\centering
\caption{The F1-score and TPR for the 2DSig-Norm to defend against PGDM attacks with different noise levels $\varepsilon$.}
\label{tab:pgdm_f1}
\begin{footnotesize}
\begin{tabular}{cc|cc|cc}
\hline\hline
\multicolumn{2}{c|}{\multirow{2}{*}{}}                              & \multicolumn{2}{c|}{RepVGG-A2}                         & \multicolumn{2}{c}{ResNet-20}                         \\ \cline{3-6} 
\multicolumn{2}{c|}{}                                               & \multicolumn{1}{c|}{Untarget}        & Target          & \multicolumn{1}{c|}{Untarget}        & Target          \\ \hline\hline
\multicolumn{1}{c|}{\multirow{2}{*}{F1-score}} & $\varepsilon=0.03$ & \multicolumn{1}{c|}{0.88 $\pm$ 0.01} & 0.89 $\pm$ 0.01 & \multicolumn{1}{c|}{0.88 $\pm$ 0.01} & 0.88 $\pm$ 0.02 \\ \cline{2-6} 
\multicolumn{1}{c|}{}                          & $\varepsilon=0.02$ & \multicolumn{1}{c|}{0.86 $\pm$ 0.01} & 0.86 $\pm$ 0.01 & \multicolumn{1}{c|}{0.84 $\pm$ 0.01} & 0.86 $\pm$ 0.01 \\ \hline
\multicolumn{1}{c|}{\multirow{2}{*}{TPR}}      & $\varepsilon=0.03$ & \multicolumn{1}{c|}{0.99 $\pm$ 0.00}  & 1.00 $\pm$ 0.00   & \multicolumn{1}{c|}{0.99 $\pm$ 0.0}  & 1.00 $\pm$ 0.00   \\ \cline{2-6} 
\multicolumn{1}{c|}{}                          & $\varepsilon=0.02$ & \multicolumn{1}{c|}{0.94 $\pm$ 0.01} & 0.95 $\pm$ 0.01 & \multicolumn{1}{c|}{0.92 $\pm$ 0.01} & 0.96 $\pm$ 0.01 \\ \hline\hline
\end{tabular}
\end{footnotesize}
\end{table}

\end{document}